\documentclass[12pt, a4paper]{dalthesis}

\usepackage[english]{babel}
\usepackage[latin1]{inputenc}

\usepackage{graphicx}
\graphicspath{{./img_conclusion/}{./img_introduction/}}

\usepackage{amsmath,amsfonts,amsthm,amssymb}

\usepackage{enumerate}
\usepackage{multicol}
\usepackage{url}
\usepackage{xspace}
\usepackage{algorithm}
\usepackage[noend]{algorithmic}

\newtheorem{theorem}{Theorem}[chapter]
\newtheorem{lemma}[theorem]{Lemma}
\newtheorem{proposition}[theorem]{Proposition}

\theoremstyle{definition}
\newtheorem{definition}[theorem]{Definition}

\newtheorem{example}[theorem]{Example}
\newtheorem{problem}[theorem]{Problem}

\usepackage[usenames]{color} 
\definecolor{Blue}{rgb}{0,0.16,0.90}
\definecolor{Red}{rgb}{0.90,0.16,0}
\definecolor{DarkBlue}{rgb}{0,0.08,0.45}
\definecolor{ChangedColor}{rgb}{0.9,0.08,0}
\definecolor{CommentColor}{rgb}{0.2,0.8,0.2}
\definecolor{ToDoColor}{rgb}{0.1,0.2,1}

\usepackage{moreverb}
\usepackage{bm}
\renewcommand{\vec}[1]{\bm{#1}}
\def\vv{\vec{v}}

\usepackage{mdwlist}

\def \R {\mathbb R}
\def \Z {\mathbb Z}

\def \calL {\mathcal L}
\def \calO {\mathcal O}
\def \calP {\mathcal P}

\def \cA {\mathcal{A}}
\def \cB {\mathcal{B}}
\def \cE {\mathcal{E}}
\def \cG {\mathcal{G}}
\def \cH {\mathcal{H}}

\def \cO {\mathcal{O}}
\def \cP {\mathcal{P}}
\def \cR {\mathcal{R}}
\def \cS {\mathcal{S}}
\def \cT {\mathcal{T}}
\def \cV {\mathcal{V}}

\newcommand{\frakg} {\mathfrak g}

\def \insight {{\sffamily Insight}\xspace}
\def \impact  {{\sffamily Impact}\xspace}
\def \libc    {{\tt libc}\xspace}

\newcommand{\reffig}[1]{Fig.~\ref{#1}}
\newcommand{\component}[1]{{\sf #1}}
\newcommand{\xmltag}[1]{{\tt #1}}

\newcommand{\Name}[1]{{\emph{#1}}\xspace}

\def \Mff     {Metric-FF\xspace}
\def \Sg      {SGPlan\xspace}
\def \add	    {\mbox{effect}^{+}}
\def \del     {\mbox{effect}^{-}}
\def \pre     {\mbox{precond}}

\newcommand{\componentC}{\ensuremath{C}}

\newcommand{\start}{\ensuremath{*}}
\newcommand{\emptyfirewall}{\ensuremath{\emptyset}}

\newcommand{\Figure}[5]{
\begin{figure}[#2]
\centering
\includegraphics[width=#3]{#1}
\caption{#5}
\label{fig:#4}
\end{figure}
}

\newcommand{\menor}{\leqslant} 
\newcommand{\vmenos}{\vspace{-0.3cm}}
\newcommand{\saltito}{\vspace{0.05cm}}

\def \autor        {Carlos SARRAUTE}

\def \grado        {Doctor en Ingenier\'ia Inform\'atica}
\def \titulo       {\textbf{Automatización de Planning de Ataques Informáticos}}
\def \universidad  {Instituto Tecnol\'ogico de Buenos Aires}
\def \lugar        {Buenos Aires, Argentina}
\def \fechadefensa {2 de Julio de 2012}
\def \fechacopyright {2012}
\def \fecha        {Julio 2012}

\def \director     {Gerardo Richarte}
\def \codirector   {Eduardo Bonelli}

\begin{document}
\clubpenalty = 10000
\widowpenalty = 10000

\thispagestyle{empty}

\begin{center}

\begin{figure}[!ht]
\centering
\vspace{-1 cm}
\includegraphics[width=0.6 \linewidth]{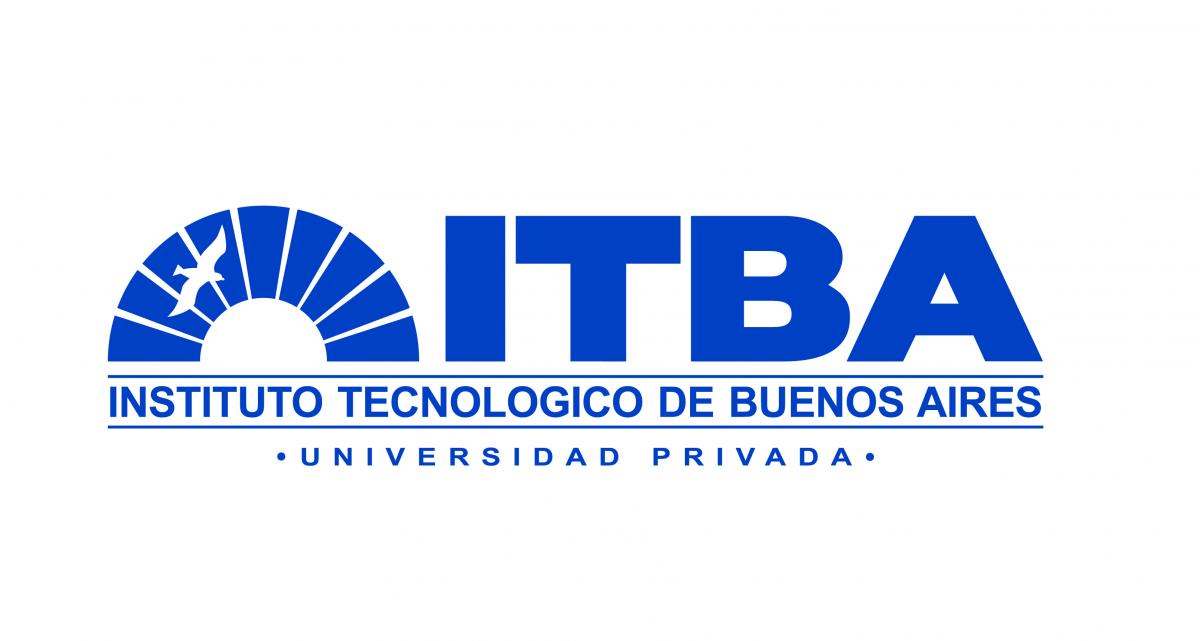}
\end{figure}

Tesis para optar al título de \\
\grado \\
del \\
\universidad \\

\vfill

\textbf{\Large \titulo}

\vfill

Autor: \autor \\
\vskip 2.5ex
Director: \director \\
Co-director: \codirector \\

\vfill

\lugar \\
\fecha \\

\end{center}
\clearpage

 \thispagestyle{empty}%
 \begin{center}
    \hyphenpenalty=10000\large
    \MakeUppercaseWithNewline{\titulo}
 \end{center}
 \vfill
 \begin{center}
    \rmfamily por
 \end{center}
 \vfill
 \begin{center}
    \rmfamily \autor
 \end{center}
 \vfill
 \begin{center}
    \small En cumplimiento parcial de los requisitos \\
	para optar al grado de \\
    \vskip 2.5ex
    \MakeUppercaseWithNewline{\grado} \\
    \vskip 2.5ex
    del \\
    \vskip 2.5ex
    \MakeUppercaseWithNewline{\universidad} \\
 \end{center}
 \vfill
 {\small \lugar \hfill \fechadefensa}
 \vskip0.25in
 \begin{center}
    \rmfamily \copyright\ Copyright \autor, \fechacopyright
 \end{center}
 \clearpage


\title{\textbf{Automated Attack Planning}}

\author{\autor}
\dept{Informatics Engineering}
\faculty{Engineering}
\university{\universidad}
\address{\lugar}

\submitdate{June 4th, 2012}
\defencedate{July 2nd, 2012}
\copyrightyear{\fechacopyright}
\convocation{July}{2012}

\degree{Doctor of Philosophy}
\degreeinitial{Ph.D.}

\supervisor{\director}
\supervisor{\codirector}

\firstreader{J\"org Hoffmann}
\secondreader{Hugo Scolnik}
\thirdreader{Marcelo Frias}

\nodedicationpage
\nolistoftables
\nolistoffigures

\beforepreface

\begin{abstract}

Penetration Testing (short {\em pentesting}) is a methodology for
assessing network security, by generating and executing possible
attacks exploiting known vulnerabilities of operating systems and
applications.
Doing so automatically allows
for regular and systematic testing without a prohibitive amount of
human labor, and makes pentesting more accessible to non-experts. A
key question then is how to automatically generate the attacks.

A natural way to address this issue is as an {\em attack planning}
problem.
In this thesis, we are concerned with the specific context of
regular automated pentesting, and use the term ``attack planning'' in that sense. 
The following three research directions are investigated.


First, we introduce a conceptual model of computer network attacks,
based on an analysis of the penetration testing practices.
We study how this attack model can be represented in the PDDL language.
Then we describe an implementation that integrates 
a classical planner with a penetration testing tool. 
This allows us to automatically generate
attack paths for pentesting scenarios, and to validate these attacks
by executing the corresponding actions -including exploits- against the
real target network. 
We also present another tool that we developed in
order to effectively test the output of the planner:
a simulation platform created to design and
simulate cyber-attacks against large arbitrary target scenarios.


Secondly, we present a custom probabilistic planner.
In this part, we contribute a planning model that captures the uncertainty about the results of the actions, which is modeled as a probability of success of each action. 
We present efficient planning algorithms, specifically designed for this problem, that achieve industrial-scale runtime performance
(able to solve scenarios with several hundred hosts and exploits). 
Proofs are given that the solutions obtained are optimal under certain assumptions.
These algorithms take into account the probability of success of the actions and their expected cost (for example in terms of execution time, or network traffic generated).


Finally, we take a different direction: 
instead of trying to improve the efficiency of the solutions developed,
we focus on improving the model of the attacker.
We model the attack planning problem in
terms of partially observable Markov decision processes (POMDP). This
grounds penetration testing in a well-researched formalism,
highlighting important aspects of this problem's nature. 
POMDPs allow the modelling of information gathering as an integral part of the
problem, thus providing for the first time a means to intelligently
mix scanning actions with actual exploits.

\end{abstract}

\begin{acknowledgments}

First of all I want to thank Gerardo Richarte,
with whom I invented the idea and the project of \emph{attack planning},
and continued to work on its different instantiations;
and who had the generosity of directing me in this special phase
which is completing the Ph.D. thesis.
I learned a lot working with him, but there is something that I want to highlight:
Gera taught me what it means to deeply understand a problem,
and to do frenetic research, the one that doesn't let you sleep.

\medskip

Thanks to Emiliano Kargieman, who introduced me to the world of Hacking
and Information Security, and who also created the conditions
that allowed me to start the Ph.D. journey at ITBA.
To Ariel Futoransky, with whom I shared countless talks and discussions 
around the white board. It is thanks to Futo that several ideas of this thesis
and of all my other research projects took form.

\medskip

My appreciation to Eduardo Bonelli. With his multidisciplinary interest
and academic experience, he was able to put this thesis on track and guide it 
to its final port. To Roberto Perazzo and all the Ph.D. team at ITBA,
who managed to create an environment of freedom and innovation,
that I hope will continue to grow.

\medskip
Thanks to J\"org Hoffman and Olivier Buffet. Meeting J\"org in Atlanta,
and beginning a collaboration with both of you was definitely 
a big inflection point in my thesis, that put me in direct contact
with the world of Artificial Intelligence and Planning.
Thanks for receiving me at INRIA in Nancy, for sharing your knowledge
and teaching me new ways of working (and also for the results obtained!)

\medskip
Thanks to Hugo Scolnik, Marcelo Frias and J\"org (again) for accepting
to be jury of my thesis, and for their reviews and comments.  

\medskip

To Jorge Lucangeli; the time that we worked together was very productive
and enjoyable, I wish it would have last longer.
To Luciano Notarfrancesco; brainstorming a few weeks with Luciano and Gera we produced more ideas
than we were able to implement in several years of work.
To Aureliano Calvo, always willing to chat and discuss new ideas and algorithms.
To Wata, Peter, Gutes, Fruss, Fercho, Jose and all the CoreLabs team.
Thanks to you CoreLabs remained all these years a fruitful and motivating place.

\medskip
To Dragos Ruiu, Rodrigo Branco, Philippe Langlois, Jonathan Brossard,
Matthieu Suiche and all the hackers' community,
for providing great spaces to present and discuss research ideas
(and also of course to make new friends and have a great time!)

\medskip
To James Foster, who patiently revised and corrected all my papers
and presentations.

\medskip

To my professor Adrián Paenza, for inspiring me to be a better person,
and keeping alive the flame of Mathematics (in me and so many others!)

\medskip

To my father Reynaldo, with his passion and enthusiasm he transmitted
me the love for reading, writing, studying, debating, discussing...
as a worker and activist, he taught me to appreciate hard work, critical and scientific thinking,
and guided me to follow a path that gives me today so many satisfactions.

\medskip

To my mother Yuriko, my sister Clara and my brother Sebas, who always accompany me
with their support and affection, and open my head to new worlds.

\medskip
\bigskip

To Nela, León and Leia, my infinite sources of Love and happiness.

\end{acknowledgments}


\afterpreface



\chapter{Introduction} \label{chap:introduction}

\section{Overview of the Thesis}

Penetration Testing (short {\em pentesting}) is a methodology for
assessing network security, by generating and executing possible
attacks exploiting known vulnerabilities of operating systems and
applications (the necessary background is given in Chapter~\ref{chap:penetration_testing}). 
Doing so automatically allows
for regular and systematic testing without a prohibitive amount of
human labor, and makes pentesting more accessible to non-experts. A
key question then is how to automatically generate the attacks.

A natural way to address this issue is as an {\em attack planning}
problem.
In this thesis, we are concerned with the specific context of
regular automated pentesting, as in Core Security's ``Core Insight
Enterprise'' tool. We will use the term ``attack planning'' in that
sense.

The main body of this thesis is divided in three parts. 

\saltito \saltito
\begin{description}
	\item[Part I] 
In the first part we present the 
basic model and a complete implementation that integrates a planner system
with a penetration testing framework and a network simulation tool.

Chapter~\ref{chap:model} introduces a conceptual model of computer network attacks.
This model is based on an analysis of the penetration testing practices,
and of the functionality provided by pentesting frameworks.
This model gives a new perspective
for viewing cyberwarfare scenarios, by introducing conceptual
tools to evaluate the costs of an attack,
to describe the theater of operations, targets, missions, actions,
plans and assets involved in cyberwarfare attacks.

We then present in Chapter~\ref{chap:representations} how this attack model
can be represented in the PDDL language.
To define with precision this representation, we start
with simpler representations based on set theory 
in Section \ref{sec:set-theoretic-representation}
and on first order logic in Section \ref{sec:first-order-representation}.

Chapter~\ref{chap:deterministic} describes an implementation that integrates 
a classical planner with a penetration testing tool. 
This allows us to automatically generate
attack paths for pentesting scenarios, and to validate these attacks
by executing the corresponding actions -including exploits- against the
real target network. We present an algorithm for transforming the information
present in the pentesting tool to the planning domain 
in Section \ref{sec:transform-architecture}, and we show
how the scalability issues of attack graphs can be solved using current
planners. In Section \ref{sec:integration-performance} we make an analysis 
of the performance of our solution, showing
how the model scales to medium-sized networks and the number of actions
available in current pentesting tools.

Finally, we present another tool that we developed in
order to effectively test the output of the planner. 
Chapter~\ref{chap:simulations} is devoted to a simulation platform created to design and
simulate cyber-attacks against large arbitrary target scenarios. 
This simulator has surprisingly low hardware and configuration requirements, while making the simulation a realistic experience from the attacker's standpoint. The scenarios include a crowd of simulated actors: network devices, hardware devices, software applications, protocols, users, etc. 
A novel characteristic of this tool is to simulate vulnerabilities (including 0-days) and exploits, allowing an attacker to compromise machines and use them as pivoting stones to continue the attack. A user can test and modify complex scenarios, with several interconnected networks, where the attacker has no initial connectivity with the objective of the attack. 

Chapters~\ref{chap:representations} and \ref{chap:deterministic} are based on work done 
with Gerardo Richarte and Jorge Lucangeli,
which has been published in the SecArt workshop at the AAAI conference \cite{LucSarRic10}
and presented in the Hackito Ergo Sum conference \cite{Sarraute10}.
Chapter~\ref{chap:simulations} is based on joint work 
with Ariel Futoransky, Fernando Miranda and Jos\'e Orlicki,
and was published in the SIMUTools conference \cite{FutMirOrl09}.

\saltito \saltito

\item[Part II]
The second part of the thesis describes the development of a custom probabilistic planner.
In Chapter~\ref{chap:probabilistic} we contribute a planning model that captures the uncertainty about the results of the actions, which is modeled as a probability of success of each action. We present efficient planning algorithms, specifically designed for this problem, that achieve industrial-scale runtime performance
(able to solve scenarios with several hundred hosts and exploits). 
These algorithms take into account the probability of success of the actions and their expected cost (for example in terms of execution time, or network traffic generated).

Of course planning in the probabilistic setting is far more difficult 
than in the deterministic one.
We do not propose a general algorithm, but a solution suited
for the scenarios that need to be solved in a real-world penetration test.
Two ``primitives'' are presented, which are used as building blocks
in a framework separating the overall problem into two levels of abstraction.
The computational complexity of our planning solution is $\calO ( n \log n) $,
where $n$ is the total number of actions in the case of an attack tree
(with fixed source and target hosts),
and $\calO ( M^{2} \cdot n \log n) $ where $M$ is the number of machines
in the case of a network scenario.
We discuss experimental results obtained with our implementation.
For example, we were able to solve planning
in scenarios with up to 1000 hosts distributed in different networks.

The work reported in this part was presented 
at the FRHACK conference \cite{Sarraute09} 
and the H2HC conference \cite{Sar09proba}.
It was published in the ACM Workshop on Artificial Intelligence and Security  \cite{SarRicLuc11}
(joint work with Gerardo Richarte and Jorge Lucangeli).
A discussion on the exploits' costs was given at the 8.8 Security Conference \cite{Sarraute-8dot8}.

\saltito \saltito

\item[Part III]
The third part takes a different direction: 
instead of trying to improve the efficiency of the solutions developed,
we focus on improving the model of the attacker, thus formulating the
problems that we think will have to be solved in the longer term.

The approach of Chapter~\ref{chap:deterministic} used classical planning and hence ignores all
the incomplete knowledge that characterizes hacking. 
The more recent approach of Chapter~\ref{chap:probabilistic}
makes strong independence assumptions for the sake
of scaling.
In Chapter~\ref{chap:pomdp}, we take the opposite extreme of the trade-off between accuracy
and performance. We tackle the problem in full, in particular
addressing information gathering as an integral part of the attack. 
We achieve this by modeling the problem in terms of partially observable
Markov decision processes (POMDP). 
This grounds penetration testing in a well-researched formalism,
highlighting important aspects of this problem's nature. 
POMDPs allow us to model information gathering as an integral part of the
problem, thus providing for the first time a means to intelligently
mix scanning actions with actual exploits.

As a side effect, this modeling
activity serves to clarify some important aspects of this problem's
nature. A basic insight is that, whereas in Chapter~\ref{chap:probabilistic}
we model the uncertainty as
non-deterministic actions---success probabilities of exploits---this
uncertainty is more naturally modeled as an uncertainty 
about {\em states} (in Chapter~\ref{chap:pomdp}). 
The exploits as such are deterministic in that their
outcome is fully determined by the system configuration.
Once this basic modeling choice is made, all the rest falls into place naturally.

Our experiments are based on a problem generator that is not
industrial-scale realistic, but that allows us to create reasonable test
instances by scaling the number of machines, the number of possible
exploits, and the time elapsed since the last activity of the
pentesting tool. Unsurprisingly, we find that POMDP solvers do not
scale to large networks. However, scaling is reasonable for individual
pairs of machines. As argued in Section~\ref{sec:distinguished-assets}, 
such pairwise strategies can serve as
the basic building blocks in a framework decomposing the overall
problem into two abstraction levels.

The material of Chapter~\ref{chap:pomdp} is joint work with 
J\"org Hoffmann and Olivier Buffet, 
and has been published in the SecArt workshop at IJCAI \cite{SarBufHof11}.
A discussion of the different directions taken in our attack planning research
was given in the H2HC conference \cite{Sarraute-h2hc11}.

\end{description}

\section{The Confluence}

The work of this thesis was performed while I was working
as a researcher in CoreLabs, the research center of a company
that produces software for penetration testing.
This is an important part of the context, since it explains the original
motivation, the original questions that we tried to answer with Gerardo Richarte:
We have this penetration testing tool, how can we make it easier to use?
How can we make it more autonomous?
How can we put more expert knowledge in the software, so it requires less
expertise from the user?

The work started with a strong foothold in the world of industry, 
and thanks to my co-supervisor Eduardo Bonelli it also gained a foothold in the academic world.
For the solutions that we devised, we also asked ourselves:
Has this problem already been studied? How do our solutions compare to the ones found
by other researchers? What are the state-of-the-art tools that we could use?
How can we assure that the solutions that we find are optimal (or at least good)?

This way, the natural flow of the research led us to work 
in what I like to think of as the confluence of Industry and Academics.
Answering questions that come from real-world applications, and in the process
generating ideas and formulating models that have an academic interest.
In my limited experience, I saw that the worlds of Industry and Academics are often 
disjointed, and that there is a lot to gain by building bridges between both worlds.

This work is also placed in another confluence: the problems that we try to solve
come from the field of Information Security, 
and the tools used to provide answers come mainly from the field of Artificial Intelligence.
Being at this intersection also opened up the way to a very interesting experience:
a collaboration with J\"org Hoffman and Olivier Buffet,
recognized experts in the field of Automated Planning,
experienced researchers and great persons from whom I learned a lot.
And fortunately we were not alone at this intersection:
the workshops SecArt (on Intelligent Security) and AISec (on Artificial Intelligence and Security)
provided a great environment to present and discuss some of the results that we obtained.
Another confluence where fruitful bridges are being built.

\bigskip
\medskip

\begin{figure}[!ht]
\centering
\includegraphics[width=\linewidth]{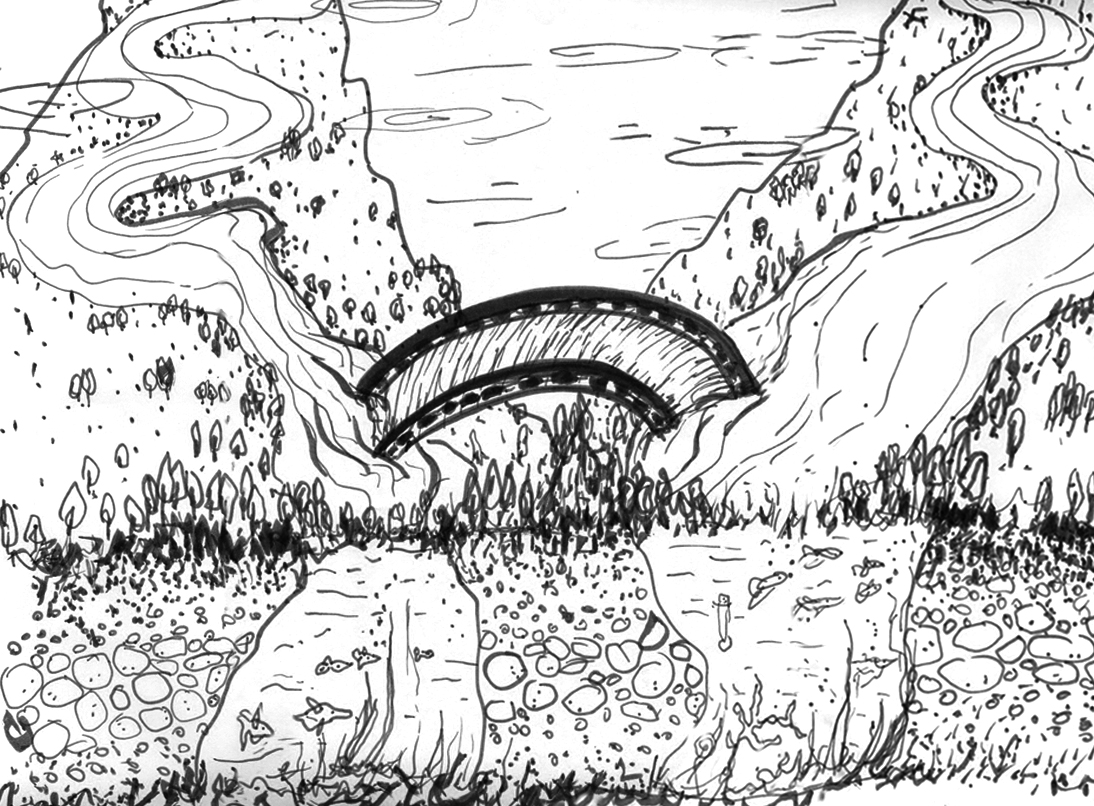}
\caption{The Confluence.}
\label{fig:the-confluence}
\end{figure}


\chapter{Background on Penetration Testing} \label{chap:penetration_testing}

In this preliminary chapter, we give a brief background 
on computer network intrusions and penetration testing frameworks.
The study of these tools, and the need to automate their functionality,
provides the basis for the work of this thesis.

\section{Computer Network Intrusions}

During a network intrusion, an attacker tries to 
gain access to software systems that require authorization
(web servers, database servers, accounting systems).
The intrusion may be illegal (this is what people usually have in mind when 
speaking about intrusions), or may be an authorized audit performed by security professionals.
The latter is called a network penetration test,
of which we give a definition below.

\begin{definition}
A \emph{penetration test}, also called \emph{pentest}, is a method of evaluating the security of a computer system or network by performing a controlled attack. 
The process involves an active analysis of the system for any potential vulnerabilities that could result from poor or improper system configuration, both known and unknown hardware or software flaws, or operational weaknesses in process or technical countermeasures. This analysis is carried out from the position of a potential attacker and involves active exploitation of security vulnerabilities.
\end{definition}

\begin{figure}[ht]
\centering
\includegraphics[width=\linewidth]{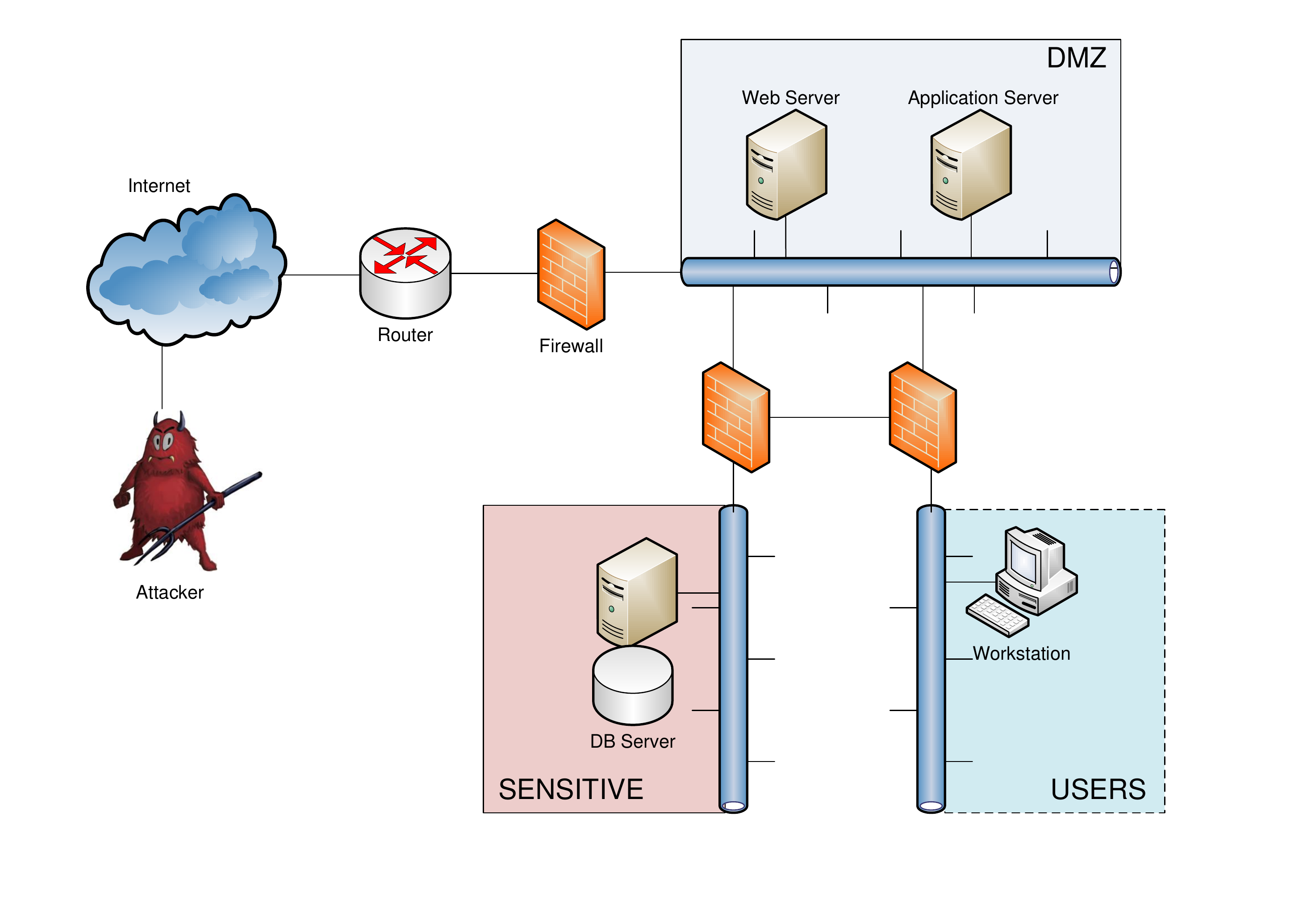}
\caption{Anatomy of a real-world attack --- Sample scenario.}
\label{fig:attack_scenario}
\end{figure}

As networks evolve, and combine a multitude of interconnected technologies, the penetration test
has become an accepted practice to evaluate the global security of a network
(ultimately assessing effectiveness of the deployed security countermeasures).
The interesting point for us is that pentesters basically use the same tools and methodologies
as unauthorized attackers, so we can focus on the former (whose practices are also more documented!)

\section{On Vulnerabilities and Exploits}
\label{sec:exploits}

\subsection{Basic Definitions} \label{sec:exploits-definitions}

\begin{definition}
A \emph{vulnerability} (noun) is a flaw in a system that, if leveraged by an attacker, can potentially
impact the security of said system. 
Also called a security bug, security flaw, or security hole \cite{Arce05}.
It may also be an intentional feature -- in this case it is called a \emph{backdoor}.
\end{definition}

\Figure{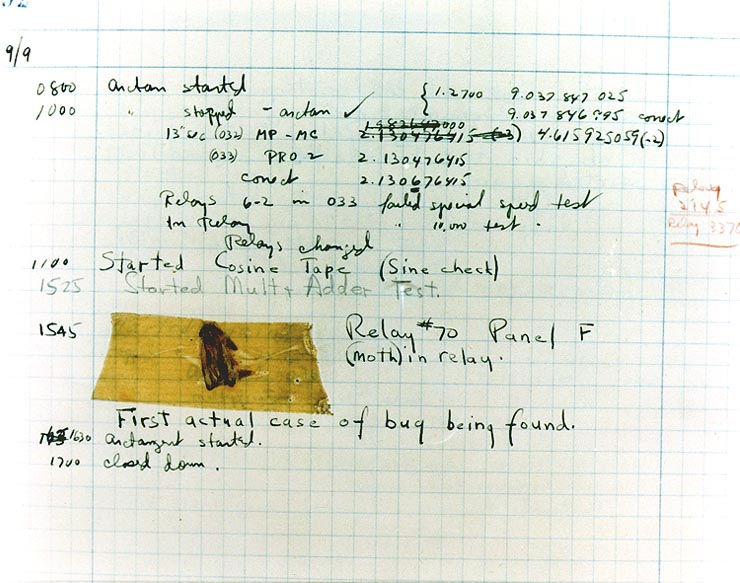}{!ht}{\linewidth}{first_bug}{First Computer Bug.}

\begin{definition}
To \emph{exploit} (verb) is to use or manipulate to one's advantage a 
security vulnerability.
\end{definition}

\begin{definition} \label{def:exploit}
An \emph{exploit} (noun) is a piece of software, a chunk of data, or sequence of commands that take advantage of a bug or vulnerability  in order to cause unintended or unanticipated behavior to occur 
on a computer software, hardware, or electronic device. This frequently includes such things as gaining control of a computer system or allowing privilege escalation or a denial of service attack.
\end{definition}

\subsection{Anatomy of an Exploit}

The exploits are the most important actions during an attack.
According to the literal meaning of ``exploit'', 
it takes advantage and makes use of a hidden functionality.
When used for actual network attacks, exploits execute code that can alter, 
destroy or expose information assets. 
When examining an exploit, three main components can be distinguished.

\subsubsection{Attack Vector}
The attack vector is the mechanism the exploit uses to make a vulnerability manifest,
in other words, how to reach and trigger the bug.
For example, in the case of Apache Chunked Encoding Exploit, the attack vector is the 
TCP connectivity that must be established on port 80 to reach the application.
In a client-side attack, the attack vector might be an email sent to a user,
with a specially crafted attachment that will trigger a vulnerability in the application 
used to open it.
 
\subsubsection{Exploited Vulnerability}
To obtain an unauthorized result, the exploit makes use of a vulnerability. 
This can be a network configuration vulnerability,
or a software vulnerability: a design flaw or an implementation flaw 
(buffer overflow, format string, race condition).

The most classic example is the buffer overflow, first described in 
``Smashing the stack for fun and profit''  by Aleph One \cite{Aleph96}.
The questions for the attacker are: how to insert code and how to modify the execution flow
to execute it?
In the example of a stack based buffer overflow, the code is inserted in a stack buffer
and by overflowing the buffer, the attacker can overwrite the return address and jump to his code.

\subsubsection{Payload}
Once the attacker manages to trigger and exploit a security flaw,
he gains control of the vulnerable program.
The payload is the functional component of the exploit, the code the attacker is interested in running.
Classic payloads allow attackers to:

\begin{itemize}
\item{Add a user account: 
on Unix systems, it was done by adding a line to the system password file (/etc/password)
or changing the password of root.
However such changes are easily detected
and to use the account the attacker needs connectivity through legitimate paths (firewalls can block them). This classic payload is no longer used.
}
\item{Make changes to system configuration:
for example, to add a line to inetd (Internet services daemon),
to open a port and later connect to the system via the newly opened port.
}
\item{Open a shell: the payload consists of opening a shell (a command interpreter), 
that the attacker can use to execute available commands.
These payloads are more difficult to detect, but are also more difficult to write.
See the article of Aleph One \cite{Aleph96} for a report of this technique.
Commonly known as \emph{shellcode}, this was the most popular payload, until the development of
exploitation frameworks such as Metasploit and Core Impact provided
more generic techniques, as we discuss in the next section.
}
\end{itemize}

Writing payloads is a very difficult task, that requires solving multiple restrictions simultaneously.
The payload is a sequence of byte codes, so each payload will only work on 
a specific operating system and on a specific platform.
Depending on the attack vector, the payload may be sent to the vulnerable machine
as an ASCII string (or some protocol field), and thus must respect a particular grammar
(examples: byte 0 is forbidden, only 7-bit ASCII is accepted, only alphanumeric characters are accepted, etc.)
Libraries have been developed to help exploit writers to generate shellcodes.
MOSDEF and InlineEgg are two well known cases, with tools to cope with the restrictions.
The payload is also typically limited in size (for example the buffer size in the case of a buffer overflow),
so the code that the attacker will run must fit in a few hundred bytes.
 If he wants to execute more complex applications, 
he must find another way...

\subsection{Syscall Proxy Agents} \label{sec:syscall-proxy-agent}

We present here a solution to the limitations of payload development described
in the previous section. It is called ``syscall proxy'' and was developed by Caceres et al.
(see \cite{Caceres02} for more details).
The idea is to build a sort of ``universal payload'' that allows an attacker
to execute any system call on the vulnerable host.
By installing a small payload (a thin syscall server),
the attacker will be able to execute complex applications on his local host (a fat client),
with all system calls executed remotely.

\subsubsection{Background on Syscalls}

An application usually interacts with certain resources: a file on a disk, the screen, a
networking card, a printer, etc. Applications can access these resources through \emph{system calls} (\emph{syscalls} for short). These
syscalls are operating system services, usually identified with the lowest layer of communication between a user mode
process and the OS kernel.

Different operating systems implement syscall services differently, sometimes depending on the processor's architecture.
The main groups are UNIX and Windows.

UNIX systems use a generic and homogeneous mechanism for calling system services, 
usually in the form of a ``software interrupt". 
Syscalls are classified by number and arguments are passed either through the
stack, registers or a mix of both. The number of system services is usually kept to a minimum 
(about 380 syscalls are found in a Linux with kernel version 3.0\footnote{Updated as
of June 2012.}),
as more complex functionality is provided on higher user-level functions.

Windows also provides system calls. However, for the sake of simplicity and
generality, the authors of \emph{syscall proxy} decided to 
use ``Windows syscalls" to refer to any function in any dynamic library available 
to a user mode process \cite{Caceres02}.

\subsubsection{Syscall Proxy}

The resources that a process can access, and the kind of access it has to them, define the
``context" in which it is executed.
For example, a process that reads data from a file might do so using the open, read and close syscalls.

\Figure{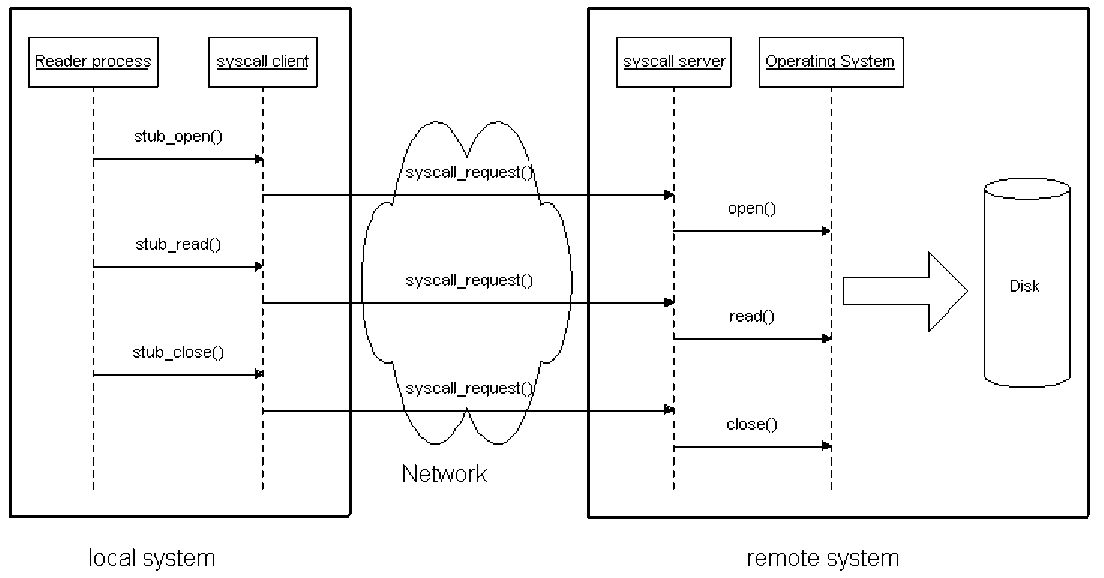}{ht}{12cm}
{syscalls1}{Diagram of a proxy call execution}

Syscall proxying inserts two additional layers between the process and the underlying operating system. These layers
are the \emph{syscall client} layer and the \emph{syscall server} layer.

\begin{description}
	\item[Syscall client (on local system).] 
The syscall client layer acts as a link between the running process and the underlying system services. 
This layer is responsible for forwarding each syscall argument and generating a proper request that the syscall server can
understand. It is also responsible for sending this request to the syscall server, 
usually through the Internet,
and returning back the results to the calling process.

\item[Syscall server (on remote system).]
The syscall server layer receives requests from the syscall client to execute specific syscalls using the underlying
operating system services. This layer marshals back the syscall arguments from the request in a way that the underlying
OS can understand and calls the specific service. After the syscall finishes, its results 
are marshalled\footnote{In computer science, marshalling is the process of transforming the memory representation of an object to a data format suitable for storage or transmission. It is typically used when data must be moved between different parts of a computer program or from one program to another.} and sent back to the client, again through the Internet.
\end{description}

\begin{definition} \label{def:syscall-proxy-agent}
In the context of this chapter, we will refer to a syscall server on a remote system 
as an \emph{agent}.
\end{definition}

In conclusion, the \emph{syscall proxy} technology gives the 
attacker the possibility of executing his tools on a compromised machine,
without the need of actually copying all those tools on the target machine.
This makes the process of installing a \emph{base camp} cleaner and simpler.
This technology is implemented within the agents of ``Core Impact Pro'' and ``Core Insight Enterprise'',
and allows the attacker to perform transparent \emph{pivoting} on compromised machines
(see Section~\ref{sec:pivoting}).

As a final remark, there are multiple connection methods between agents. The originating agent can use:
connect to target (similar to bindshell), connect from target (similar to reverse shell), reuse connection and HTTP tunneling. 
Agents can also be chained together to reach network resources with limited connectivity.

\section{Main Steps of an Attack} \label{sec:main_steps}

Traditionally, the pentesting process is divided in steps,
and an attack will follow the pattern of steps that we describe below.
Of course, this division in steps is arbitrary, and corresponds to an
accepted practice in the field (refer to \cite{ArcGra04,ArcRic03,Richarte03}). 
We will show in subsequent chapters that this methodology can be improved
by using planning techniques before executing the attack -- and whose output
may be to mix actions from different steps in order to run
a faster or more reliable attack.

\subsection{Information Gathering}

A successful attack depends on the ability to gather relevant information about the
target network,
including active IP addresses, operating systems and available services. 
This step is called \emph{information gathering} in the context of pentesting.
It is also called \emph{reconnaissance} in the military context, and be considered as part of the 
OODA loop\footnote{OODA stands for Observe, Orient, Decide, Act.} 
or Boyd cycle (refer to \cite{Boyd87,Higgins90}).

Actions realized during this phase include:
\begin{itemize}
\item{
Network discovery: performed using mechanisms such as ARP, TCP SYN packets, ICMP echo request, 
TCP connect and passive discovery. }
\item{
Port scanning: an exhaustive scan of open and closed ports of all the network hosts.}
\item{
OS identification: consists of recognizing the OS of a remote host by analyzing 
its responses to a set of tests. 
Classical Nmap's fingerprinting database can be combined with a neural network to accurately 
match OS responses to signatures, see \cite{BurSar06}.
Additional OS identification capabilities are available for more specific situations. 
For instance, OS detection utilizing the DCE-RPC and SMB protocols can identify Windows machines more precisely.}
\item{Other techniques available to human attackers are social engineering and 
Google hacking (using publicly available information to gain insight into the target organization).
Although these techniques are difficult (or impossible) to automate,
they can nevertheless be included in the planning phase --
eventually the execution of the resulting plan will require the intervention
of a human attacker to carry out that action.
}
\end{itemize}

\subsection{Attack and Penetrate}

During this phase, the attacker selects and launches remote exploits
making use of data obtained in the Information Gathering step. 
According to the definition given in Section \ref{sec:exploits-definitions},
an exploit is a piece of software that injects code into the vulnerable system's memory
and modifies the execution flow to make the system run the exploit code.
As discussed in Section \ref{sec:syscall-proxy-agent}, the exploit can be thought of
as a way to install an agent on a compromised host.

\subsection{Local Information Gathering}

The Local Information Gathering step collects information about computers that the attacker
has successfully compromised. During this phase, 
the attacker may gather information about the OS, network configuration, users and installed applications;
browse the filesystem on compromised systems;
view rights obtained and interact with compromised systems via shells
and other applications (for example, a remote desktop application).

\subsection{Privilege Escalation}

During the Privilege Escalation phase, the attacker attempts to penetrate deeper into a
compromised computer by running local exploits in an attempt to obtain administrative
privileges.

\subsection{Pivoting} \label{sec:pivoting}

After Privilege Escalation, the attacker can use the newly controlled host as a vantage point
from which to run attacks deeper into the network (i.e. by making use of the
\emph{syscall proxy agents} technology presented in Section~\ref{sec:syscall-proxy-agent}).
By sending instructions to an installed agent, the attacker
can run local exploits to attack systems internally, rather than from across the network.
He can view the networks to which a compromised computer is connected,
and launch attacks from any compromised system to other computers on the same network,
gaining access to systems with increasing levels of security.
That is, the attacker executes the previous steps (Information Gathering and Attacking)
using the new agent as source.

\begin{figure}[!ht]
\centering
\includegraphics[width=\linewidth]{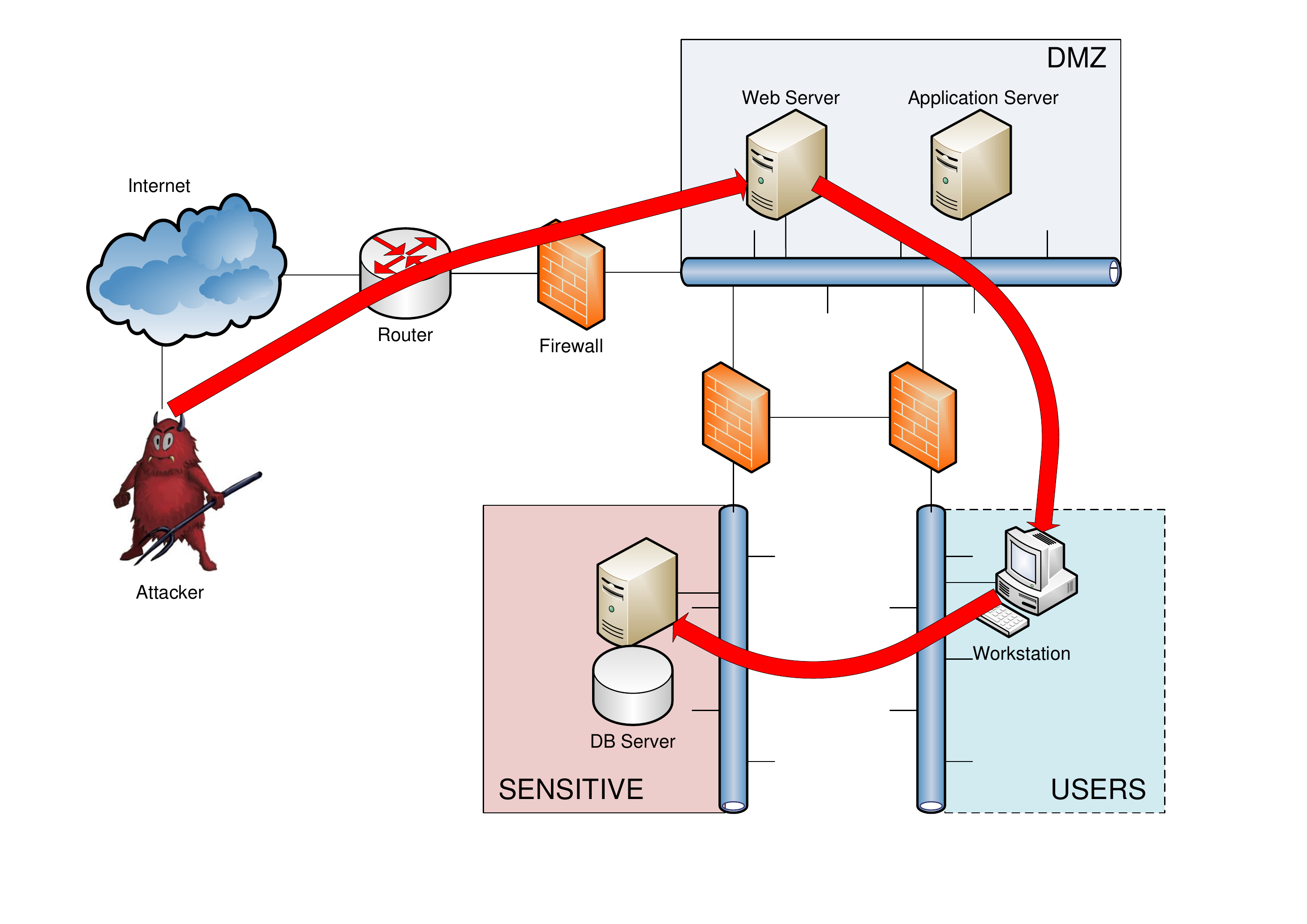}
\vspace{-1 cm}
\caption{Anatomy of a real-world attack --- Pivoting is required.}
\label{fig:attack_scenario4}
\end{figure}

\subsection{Clean Up}

The attacker needs to erase his footprints in order to avoid detection. 
Towards this end, all the executed actions
should minimize the produced noise, for example by making modifications only in memory
and by the avoidance of writing files in the target's filesystem.

\section{The Need for Automation} \label{sec:need-for-automation}

To conclude this background on penetration testing, 
we have to speak about a new kind of information security tool,
whose development started in 2001: the penetration testing frameworks. 
This type of tool facilitates the work of network penetration testers, and make the assessment of network security more accessible to non-experts. 
The main tools available are:
\begin{itemize}
  \item Core Impact (since 2001)
  \item Immunity Canvas (since 2002)
  \item Metasploit (open source project that started in 2003, owned by Rapid7 since 2009)
\end{itemize}
The book \cite{BurKilBea07} provides a survey of computer security tools, 
and in particular of penetration testing tools.

The main difference between these tools and network security scanners such
as Nessus, Qualys Guard or Retina is that pentesting frameworks have the ability to launch
real exploits for vulnerabilities, helping to expose risk by conducting
an attack in the same way a real external attacker would \cite{ArcGra04}.

As pentesting tools have evolved and have become more complex, covering new
attack vectors, and shipping increasing numbers of exploits and information gathering modules,
the problem of controlling the pentesting framework successfully has become an important question.
We detail below the main reasons:
\begin{itemize}
	\item A computer-generated plan for an attack would isolate the user from the
complexity of selecting suitable exploits for the hosts in the target network.
  \item A suitable model to represent these attacks would help to
systematize the knowledge gained during manual penetration tests performed
by expert users, making pentesting frameworks more accessible to non-experts.
  \item The possibility of incorporating the attack planning phase into
the pentesting framework would allow for optimizations based on exploit
running time, reliability, or impact on Intrusion Detection Systems.
  \item Finally, automated attacks allow the user to perform
reproducible tests, which open the path to computing security metrics
whose evolution would be a key indicator of the security posture of the organization.
\end{itemize}

\part{Integrating a Pentesting Tool with Classical Planners}


\chapter{Conceptual Model of Network Attacks} \label{chap:model}

In the first Part of this thesis, we present the basic model of penetration testing,
and a complete implementation that integrates a planner system
with a penetration testing framework and a network simulation tool.

\section{Introduction}

Our work on the attack planning problem applied to pentesting began in 2003
with the construction
of a conceptual model of an attack, distinguishing assets, actions and
goals \cite{FutNotRic03,Richarte03,ArcRic03}. In this attack model,
the assets represent both information and the modifications in the network
that an attacker may need to obtain during an intrusion, whereas the actions
are the basic steps of an attack, such as running a particular exploit
against a target host. This model was designed to be realistic from an
attacker's point of view, and contemplates the fact that the attacker has
an initial incomplete knowledge of the network, and therefore information
gathering should be considered as part of the attack.

This model also has a theoretical value: 
``we understand what we can build.'' 
Our motivation was to provide the Information Security community with a deeper and more detailed 
model of the attacks against computer networks.
It turned out that having such a model is also necessary to communicate with
other communities, such as the Artificial Intelligence and Planning ones.

The main application of this model
is to provide a basis for autonomous attack planning, 
in particular for automating penetration tests.
The description that we give in this chapter is still informal,
and lacks the level of precision required to interact with planning tools.
In Chapter \ref{chap:representations} we give a precise description
of how these ideas can be represented in the PDDL language.
In Chapter \ref{chap:deterministic} we present a deterministic version of the problem,
and use deterministic planning techniques to solve it.
In Chapter \ref{chap:probabilistic} we present a probabilistic version,
and algorithms to solve planning in the probabilistic setting.
Chapter \ref{chap:pomdp} presents a version of the attack planning problem
with uncertainty about the target network, formulated in the theoretical framework of POMDPs.

Another important application of the model is 
attack simulations, 
which can be used by a system administrator to simulate attacks against his network,
evaluate the vulnerabilities of the network and determine which countermeasures
will make it safe.
We present a network simulator based on these ideas in Chapter \ref{chap:simulations}.

In the rest of this chapter we describe the components of our model -- 
probabilistic assets, quantified goals, agents and actions -- and their relations.

\section{Assets}

\begin{definition}
An asset can represent anything that an attacker may need to
obtain during the course of an attack. In particular it 
can represent the knowledge that an agent has of a real object 
or property of the network.
\end{definition}

We describe below some examples of assets (and their parameters):
\begin{description}
\item[AgentAsset (agent, capabilities, host). ]
An AgentAsset represents an {\tt agent} with a collection of {\tt capabilities}
running on a {\tt host}.

\item[BannerAsset (banner, host, port).]
A BannerAsset represents the {\tt banner} that an agent
obtains when trying to connect to a certain {\tt port} on a {\tt host}. 

\item[OperatingSystemAsset (os, host).]
An OperatingSystemAsset represents the knowledge that an agent
has about the operating system of a {\tt host}.

\item[IPConnectivityAsset (source, target).]

\item[TCPConnectivityAsset (source, target, port).]
A TCPConnectivityAsset represents the fact that an agent is able
to establish a TCP connection between a {\tt source} host and a certain
{\tt port} of a {\tt target} host.

\end{description}

In the PDDL language, assets will be represented as predicates 
(see Section~\ref{sec:pddl-representation} for the details of the PDDL representation).

\subsection{Representations and Assumptions}

In Chapters \ref{chap:representations}, \ref{chap:probabilistic} and \ref{chap:pomdp},
we will describe different representations
of this general model. Different assumptions can be made about the model components
in order to make different trade-offs between the complexity of the resulting planning problem
and the realism of the model.

In the simpler case, the assets are deterministic, and can be translated as propositions. 
However this assumption can be relaxed with:

\subsubsection{Probabilistic Assets}

The assets can be probabilistic. This allows us to represent
properties which the attacker guesses are true with a certain probability
or negative properties (which the attacker knows to be false). 
This idea is further explored in Chapters \ref{chap:probabilistic} and \ref{chap:pomdp}.

For example,
an action which determines the operating system of a host using banners 
(OSDetectByBannerGrabber) may give as result an 
OperatingSystemAsset {\tt os=linux} with {\tt probability=0.8} and
a second one with {\tt os=openbsd} and {\tt probability=0.2}. 
Another example, an ApplicationAsset {\tt host=192.168.13.1}
and {\tt application=\#Apache} with {\tt probability=0} means that 
our agent has determined that this host is not running Apache.

\subsubsection{Level of Trust}

We can allocate a certain level of trust to the information that a certain asset represents.
When an action returns an asset, we may trust this information, 
but then trust diminishes with time. An interesting questions is how to estimate that
decrease.

\begin{problem} \label{prob:level-of-trust}
Calculate the decrease of the level of trust for each asset, as a function of time.
\end{problem}

When we first considered this problem, we had no clue on how to solve it.
Several years later, we eventually arrived (with Hoffman and Buffet) to
the POMDP model, wherein this question can be precisely formulated.
In the POMDP formulation (studied in Chapter \ref{chap:pomdp}), 
the attacker does not observe directly the state of the target network,
but rather observes a probability distribution over the states of the system,
which is called a \emph{belief state} (see definition \ref{def:belief-state}).
The problem of how to estimate the initial belief state 
(and thus to provide an answer to Problem \ref{prob:level-of-trust})
is considered in Section \ref{sec:pomdp-model-generation}.

\subsection{The Environment Knowledge} \label{sec:environment-knowledge}

The environment knowledge is a collection of information
about the computer network being attacked or hosting an agent.
Naturally, this information is represented by assets (or predicates in the PDDL language).
By abuse of language, we may speak of the {\em environment} instead
of the environment knowledge.
At the beginning of an attack, the environment contains at least an AgentAsset: 
the {\tt localAgent} which will initiate the attack.

The environment will play an important role during the planning phase
and during the execution phase of an attack, since it 
continuously feeds back the behavior of the agent.
Note also that each agent has its own environment knowledge
and that exchanging assets would be an important part of the
communications between autonomous agents.

As we will discuss in Section~\ref{sec:pddl-representation},
the input for a planner is divided in two files:
the \emph{domain file} which contains the PDDL representation of the attack model
(in particular, the actions that are available to the attacker);
and the \emph{problem file} which is a collection of predicates
about the initial state of the target system.
The environment knowledge is thus translated as the problem file
in the PDDL representation.
It also corresponds to the initial state $s_0 \in \cS$ in the Definition~\ref{def:planning-problem}
of a planning problem.

\section{Goals}

A goal represents a question or a request of the type:
``complete this asset" (every goal has an associated asset).
A goal can be quantified,
and can be associated with a list of actions which may complete his asset.
In the Definition~\ref{def:planning-problem} of a planning problem,
the goals are represented of a set of states $S_g \subseteq \cS$.
Although the goals can be enumerated explicitely,
in the earlier steps of our research we found it more convenient
to consider quantifiers to simplify the representation.

\subsection{Goal Quantifiers}
We considered three types of quantifiers: {\em Any},
{\em All} and {\em AllPossible}. An example will clarify
their meaning: consider that PortAsset has attributes 
({\tt host, port, status}).
The following goals will mean: 

\begin{description}

\item{\tt asset = PortAsset (host=192.168.13.1, status=\#open), \\
\noindent quantifiers = (Any \#port from:1 to:1024)}: \\
find an open port in host 192.168.13.1 between ports 1 and 1024.
To fulfill this goal, an action like PortScan will begin examining
the ports of host 192.168.13.1 until it finds an open port (completes
the PortAsset and returns a success signal) - or reaches
port 1024 (and returns a failure signal). 

\item{\tt asset = PortAsset (host=192.168.13.1, status=\#open), \\
quantifiers = (All \#port from:\#(21,22,23,80))}: \\
find whether all the ports \#(21,22,23,80) are open in host 192.168.13.1.
This time, PortScan will examine the four mentioned ports and return
success only if the four of them are open (and in that case
completes four PortAssets).

\item{\tt asset = AgentAsset (capabilities=\#(TCP,UDP,FileSystem)), \\
quantifiers = (AllPossible \#host from:192.168.1.0/24)}:\\
install agents in all the hosts that you can in netblock 192.168.1.0/24.
An action able to fulfill this goal will be a subclass of Exploit
(for example ApacheChunkedEncodingExploit). To fulfill this goal, the Exploit 
action will try to exploit a vulnerability in all the machines it reaches
in that netblock. It returns an AgentAsset for each compromised host,
and returns success if at least one machine is compromised.
\end{description}

\section{Actions} \label{sec:model-actions}

These are the basic steps which form an attack.
Examples of actions are:
Apache Chunked Encoding Exploit, WuFTPglobbing Exploit,
Banner Graber, OS Detect By Banner,
OS Fingerprint, Network Discovery, IP Connect, TCP Connect.
In this section we review the principal attributes of an action.

\subsection{Action Goal} \label{sec:action-goal}

An action has a goal: when executed successfully the action completes the asset
associated with its goal.
It is also common to speak about the result of an action
(for example to increase access, obtain information, corrupt information, 
gain use of resources, denial of service).

In Chapter~\ref{chap:representations} we will formalize the notion of actions.
In particular, in Definition~\ref{def:planning-domain}, we will consider
a set of actions $\cA$ and for each action $a \in \cA$ the effects of $a$.
Conceptually, with the idea of action goal, 
we are only taking into account the expected result of the action.
Undesired results and other side effects fall into the category of noise
(see Section~\ref{sec:action-noise-produced}).
In the formalization of Chapter~\ref{chap:representations},
the expected result and undesired results will be collectively considered 
as the effects of the action.

\subsection{Action Requirements} \label{sec:action-requirements}

The {\em requirements} of an action are assets that
must have been obtained before the considered action can be executed. 
The requirements are the equivalent of children nodes in \cite{Schneier00}
and subgoals in \cite{MooEllLin01} and \cite{TidLarFit01}.

An abstract action has thus the information on which assets it may satisfy
and which assets it requires before running. These relationships will be
used to construct the attack graph, by chaining actions and connecting them
with the assets they require (see also Section~\ref{sec:tree-construction}
wherein a similar graph is constructed in the context of probabilistic planning).

\begin{figure}[!ht]
\centering
\includegraphics[width=\linewidth]{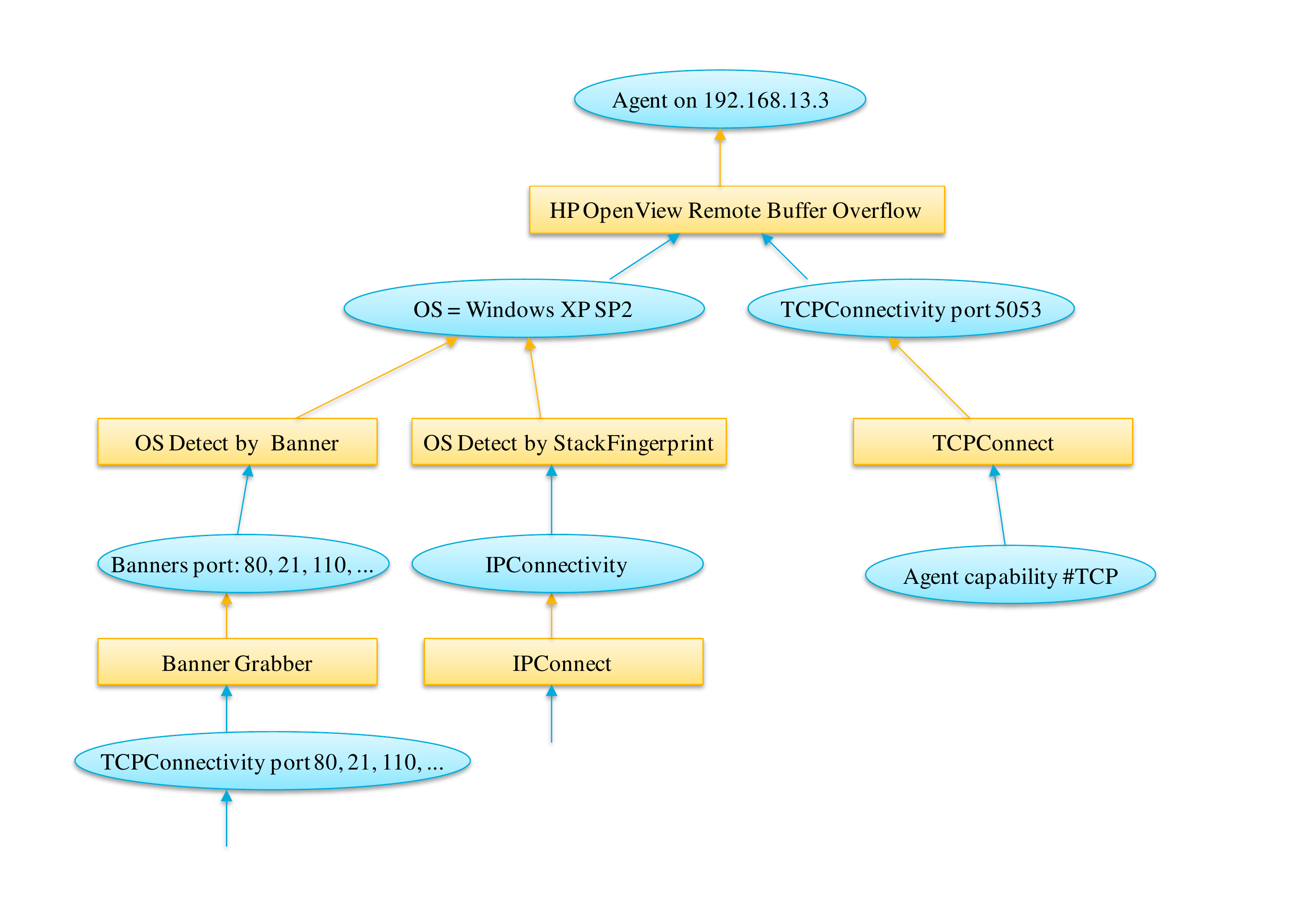}
\vspace{-1 cm}
\caption{Sample Attack Graph, showing the relation between the actions and their requirements.}
\label{fig:attack_graph1}
\end{figure}

Figure~\ref{fig:attack_graph1} shows a sample attack graph, wherein the
actions are depicted as red boxes and assets as blue ellipses. 
This is a layered graph, with alternating layers of actions and assets.
The figure also illustrates the relation between the actions and their requirements.

In the formalization of Chapter~\ref{chap:representations},
the requirements of an action $a \in \cA$ are defined as the preconditions
$\pre(a)$. In the notations of Definition~\ref{def:planning-domain},
an action $a \in \cA$ is applicable to a state $s \in \cS$ if $\pre(a) \subseteq s $.

\subsection{The Execution Phase (or the Action Itself)}

The information about the results and requirements of the actions
is used during a planning phase to obtain an attack plan $\pi$ (see Definition~\ref{def:attack-plan}).
Once the attack plan is obtained, it is used to control a penetration testing framework,
wherein the actions of the plan $\pi$ are effectively executed on the target network.
During this execution phase, the actual code of the actions
(e.g. the code of the exploits or information gathering modules) is executed by the agent.

One interesting point to mention here, is that the action
can be executed in a real network or in a simulated network
(with simulated hosts and network topology). The difference
between working in those two settings will be noticeable 
only during the execution phase. This makes our 
framework easy to adapt for both real and simulated attacks.
Chapter~\ref{chap:simulations} will describe a network simulator
specially crafted to provide realistic simulations of cyber-attacks.

\subsection{Noise Produced and Stealthiness} \label{sec:action-noise-produced}

The execution of the action will produce {\em noise}. This noise
can be network traffic, log lines in IDS, etc. 
Given a list as complete as possible of network sensors,
we have to quantify the noise produced with respect to each of these sensors.
The knowledge of the network configuration and which sensors
are likely to be active, will allow us to calculate a global estimate of 
the noise produced by the action. 
We will come back to the issue of refining the actions' costs in Section~\ref{sec:actions-costs} 
of Chapter~\ref{chap:probabilistic}, wherein we reformulate the present attack model
in the context of probabilistic planning.

With respect to every network sensor, the noise produced can be
classified into three categories: unremovable noise, noise that can
be cleaned in case the action is successful (or another subsequent
action is successful), noise that can be cleaned even in
case of failure. So we can also estimate the noise remaining after
cleanup. Of course, the {\em stealthiness} of an action will refer
to the low level of noise produced.

\subsection{Running Time and Probability of Success}

The {\em expected running time} and {\em probability of success}
depend on the nature of the action, but also
on the environment conditions, so their values must be updated
every time the agent receives new information about the environment.
These values are necessary to take decisions and choose a path
in the graph of possible actions. 

Together with the stealthiness,
these values constitute the cost of the action
and can be used to evaluate sequences of actions.
This will be further discussed in Chapter~\ref{chap:probabilistic} 
in the context of probabilistic planning.

Because of the uncertainties
inherent in the execution environment,
these values can be considered as dependend on the agent's belief
about the target network (we will discuss that in the POMDP model of Chapter~\ref{chap:pomdp}).

\subsection{Exploited Vulnerability}

The action, if aiming to obtain an unauthorized result, will exploit
a vulnerability. The information about the {\em exploited vulnerability} is
not needed by the attacking agent, but is useful for the classification and
analysis of detected intrusions. This vulnerability can be:

\begin{itemize}
\item{software vulnerability: a design flaw or an implementation flaw 
(buffer overflow, format string, race condition).}
\item{network configuration vulnerability.}
\item{trust relationship: this refers to higher level, non autonomous
attack modules: hacking a software provider, getting an insider in a software
provider, inserting backdoors in an open-source project.}
\end{itemize}

\section{Agents}

We can think of the actions as being the verbs in a sequence of 
sentences describing the attack. The agents will then be the subject
of those verbs.
Of course, an attack is always initiated by human attackers,
but during the course of the attack, actions will typically be executed 
by software agents.

\subsection{Human Attackers}

There are different types of attackers. 
They can be classified roughly as:

\begin{itemize}
	\item \textbf{script kiddies}, who attack more or less randomly 
using standard tools downloaded from the Internet;
  \item \textbf{hackers}, who attack computers for challenge, status or 
research;

  \item \textbf{security auditors} (pen testers), who evaluate the security
of a network;
  \item \textbf{government agencies and criminal organizations}, who possess access to 
  the highest skill level and resources for performing attacks.

\end{itemize}

The way we model these different types of attackers is
through the attack parameters: stealthiness, non traceability,
expected running time, expected success; and a skill level
given by the collection of actions available to the attacker.
A script kiddie will not worry about stealthiness or non traceability.
His attacks will have low expected success and require low
skill level. On the other hand, a government agency will use
maximum stealthiness, non traceability and skill level, with
a high expected success. A security auditor will not worry
about non traceability but may require stealthiness to carry out
the penetration test.

There are a number of actions that will require a human agent
to execute them, for example social engineering\footnote
{Note that social engineering could also be performed by an
autonomous agent. An example of this would be a virus who relies
on a suggestive title to be opened by the receiver.}.
But it is important to include them in our model -- 
for the sake of completeness, and also because 
they can be considered during the planning phase as part of an attack.

\subsection{Software Agents}

In general the execution of an action will require the execution
of machine code and therefore involves a software agent $\mathfrak A$
executing this code. This is a generalization of Definition~\ref{def:syscall-proxy-agent},
that refered to agents in the context of a penetration framework such
as ``Core Impact''.

The command of executing this action
might come from the agent itself, from another
software agent or from a human attacker:
we will not distinguish between these cases but just say 
that the action was executed by the agent $\mathfrak A$.
A software agent can take several forms: script, toolkit or other
kinds of programs. Let us point out the {\em autonomous agents}
who are able to take decisions and continue the attack
without human intervention.

\subsection{Communication between Agents}

This framework supports the interactions between agents, 
which collaborate to achieve the objective.
The agents establish communication channels amongst themselves
to exchange knowledge, gained information and
missions to be executed.
For example, each agent has a collection of actions. 
Agents can learn new actions and  exchange actions with each other through the 
communication channels.

The communication between human attackers can take place
through unlimited types of channels (direct interaction, telephone,
email, chat, irc, written mail). We will not enter into details here.
Examples of communication channels between software agents
are POP3 and SMTP, HTTP/HTTPS, DNS, TCP, and covert channels like Loki.

\subsection{Agent Mission}

We contemplate different types of organization between the agents.
One scenario is given by a ``root agent'' who plans the attack and
then gives the other agents orders (for executing actions), eventually
creating new agents if necessary, and asks the agents for 
feedback about action results in order to decide further steps.

Another scenario is when the root agent delegates responsibilities
to the other agents, giving them higher level {\em missions}.
To fulfill the mission, the agent will have to do its own planning
and communicate with other agents. This scenario is likely
to arise when stealthiness is a priority: communication is very
expensive and it becomes necessary to rely on the agents to
execute their missions without giving feedback (or the smallest
amount of feedback, or delayed feedback because of 
intermittent communication channels).

\section{Building an Attack}

An attack involves a group of agents, executing a series of actions
in order to fulfill a goal (or a set of goals).
Notice that this goal can change during the course of the attack.

The {\em target} is the logical or physical entity which is the objective
of the attack. Usually, the target is a computer or a 
computer network or some information hosted on a computer.
The target can also change during the course of the attack.
It is also possible that an attack has no specific target at all
(for example, a script kiddie running a specific exploit
against all the computers he reaches, until it succeeds).

The complete graph of all combinations of actions determines which goals
we (as attackers) can reach. Considering the complete graph
of possible actions, to build an attack will consist in 
finding a path to reach the objective (this implies in particular
finding a path through the physical networks to reach the target).

\subsection{Attack Parameters}

We will try to find the best path to reach the objective,
and to evaluate this we must take into account the attack
parameters: non traceability, tolerated noise, expected success,
execution time, zero-dayness.
These parameters have initial values for 
the whole attack, but they can vary from agent to agent,
for example an agent might create other agents with a different profile.

\begin{description}

\item[Non traceability:]  refers to the ability to dissimulate the 
origin of the attack. We could also call it ``deniability''.
To achieve non traceability, special modules must be
executed, who will insert intermediate agents
(we call them ``pivots'' or stepping stones) between
the originating agent and the agents executing the
objective or partial objectives.

\item[Tolerated noise:] is the level of noise that we allow
our agents to make. It can vary from agent to agent,
for example an agent executing a DNS spoofing module
would benefit from other agents simultaneously
executing a DNS flooding (and generating a high level
of noise).

\item[Expected success:] determines the priority which
will be given to successfully executing the actions, 
over the other parameters. If set to the maximum value,
the agent will try to execute the projected action,
even though the noise generated might be higher
than the tolerated noise level.

\item[Execution time:] each agent will be given a
limit of time to execute each action. This is necessary
to plan the attack, as it usually consists of a series of dependent
steps.

\item[Zero-dayness:] specifies whether the agent is allowed to use
zero-day exploits (a valuable resource that should be used only
for critical missions).

\end{description}

\subsection{Evaluating Paths}

A path is a sequence of actions (in a specific order and without branchings).
To be able to choose between different paths,
we have to evaluate the paths in terms of the attack parameters:
the probability of success, the noise probably produced,
the running time and the traceability.

For the probability of success, we consider
the success of each action as independent of the previous ones,
and that all the actions in the paths are necessary,
so the probability of success of the path is the product of
the probabilities of success of each action.

For the running time of the path, we consider three
time estimations (minimum, average, maximum) 
that we have for each module and sum them to obtain
the path's minimal, average and maximal running time.

The stealthiness of the path, that we define as the 
probability of not being discovered, diminishes with
each executed action. As with the probability of success,
we consider them independent and compute the stealthiness
of the path as the product of the stealthiness of the actions.

The traceability is harder to estimate. It depends 
basically on the number of hops (or pivots or stepping stones)
introduced, and this is how we compute it, although each can contribute
a different amount to the global ``non traceability'' of the path of actions.

\section{Applications}

\subsection{Attack Planning}

This is the main application that we are considering in this thesis,
and will be further studied in Chapters \ref{chap:representations}, 
\ref{chap:deterministic}, \ref{chap:probabilistic} and \ref{chap:pomdp}.

\subsection{Simulations and Analysis of Network Security}

As we have mentioned in Section~\ref{sec:model-actions}, our framework
can be used to build attacks against a simulated network.
Of course, the quality of this simulation will depend on how accurately
we simulate the machines. Using VMwares we obtain a slow and accurate
simulation, for faster simulations a tradeoff must be made between accuracy
and speed.

The system administrator can simulate different types of attackers
by using different attack parameters and different collections of available actions,
and evaluate the response of his network to these attackers.
For example, he can start with an attacker with a minimal portfolio of actions,
and gradually add actions to the arsenal of his simulated attacker
until there is a successful attack which goes undetected by the IDS.
This will give him an indication of which attack actions he should
defend his network against.

Also consider that the system administrator has a set of measures
which make certain attack actions less effective (in our framework, a measure
may reduce the probability of success of an attack action, or increase
the noise it produces, for example by adding a new IDS).
He can then use the simulation to see if his system becomes safe
after all the measures are deployed, or to find a minimal set
of measures which make his system safe.

As opposed to VMwares, rudimentary simulations of machines allow
us to simulate a huge amount of machines. This can be used to
investigate the dissemination of worms (considered as autonomous
agents with a minimalist set of actions) in large networks.
This is further studied in Chapter \ref{chap:simulations}.

\section{Related Work} \label{sec:attack-graphs-related-work}

\subsection{Description of Security Incidents}

In \cite{HowLon98}, Howard and Longstaff describe an incident taxonomy,
which was the result of a project to establish a
``common language'' in the field of computer security.
In \cite{LinJon97}, Lindqvist and Jonsson also work on the classification 
of intrusions.
We try to use the high-level terms proposed by Howard and Longstaff,
in particular, the attributes: vulnerability, action, target, result and objective.

One common flaw of these classifications is that they
exclusively adopt the point of view of the system owners
or the intrusion detection sensors. But an attack is 
always composed of several steps, some of which may
be invisible to the sensors.
We will add to the attributes considered in \cite{HowLon98} and \cite{LinJon97}, some
attributes which are relevant only to the attacker (or risk assessment team).
Thus a first result of our framework will be a widening of the concepts
used to describe security intrusions.

\subsection{Attack Models}

In \cite{Schneier99} and \cite{Schneier00}, 
Bruce Schneier proposes describing attacks against a system
using ``attack trees'', where each node requires the execution of
children nodes, and the root node represents the goal of the attack.
There are two types of nodes: OR nodes and AND nodes.

In \cite{TidLarFit01} the authors propose an attack specification language,
which extends the attack trees model. Each node has preconditions
(system environment properties which facilitate the execution of 
the attack node), subgoals (these are the children nodes), and
postconditions (changes in systems and environments). 
\cite{MooEllLin01} is also based on the attack trees model. The model is extended
by attack patterns and attack profiles. These authors' objective is to
provide a means for documenting intrusions. Their models are purely
descriptive and doe not allow the construction or prediction of new attacks.

\subsection{Attack Graphs}

Since the actions have requirements (preconditions) and results, given a
goal, a graph of the actions/assets that lead to this goal can be
constructed. This graph is related to the {\em attack graphs}\footnote{Nodes
in an attack graph identify a stage of the attack, while edges represent
individual steps in the attack.} studied in \cite{PhiSwi98,JajNoeBer05,NoeEldJaj09}
and many others. In \cite{LipIng05} the authors reviewed past papers
on attack graphs, and observed that the ``first major limitation of these
studies is that most attack graph algorithms have only been able to generate
attack graphs on small networks with fewer than 20 hosts''.
For example, Figure~\ref{fig:attack_graph2} shows an attack graph for 
a small network, comprising 14 hosts, taken from \cite{NoeJaj04}.
This figure illustrates the fact that attack graphs become rapidly too complex to be represented
in their integrity, whether visually or as a data structure in the computer's memory.

\Figure{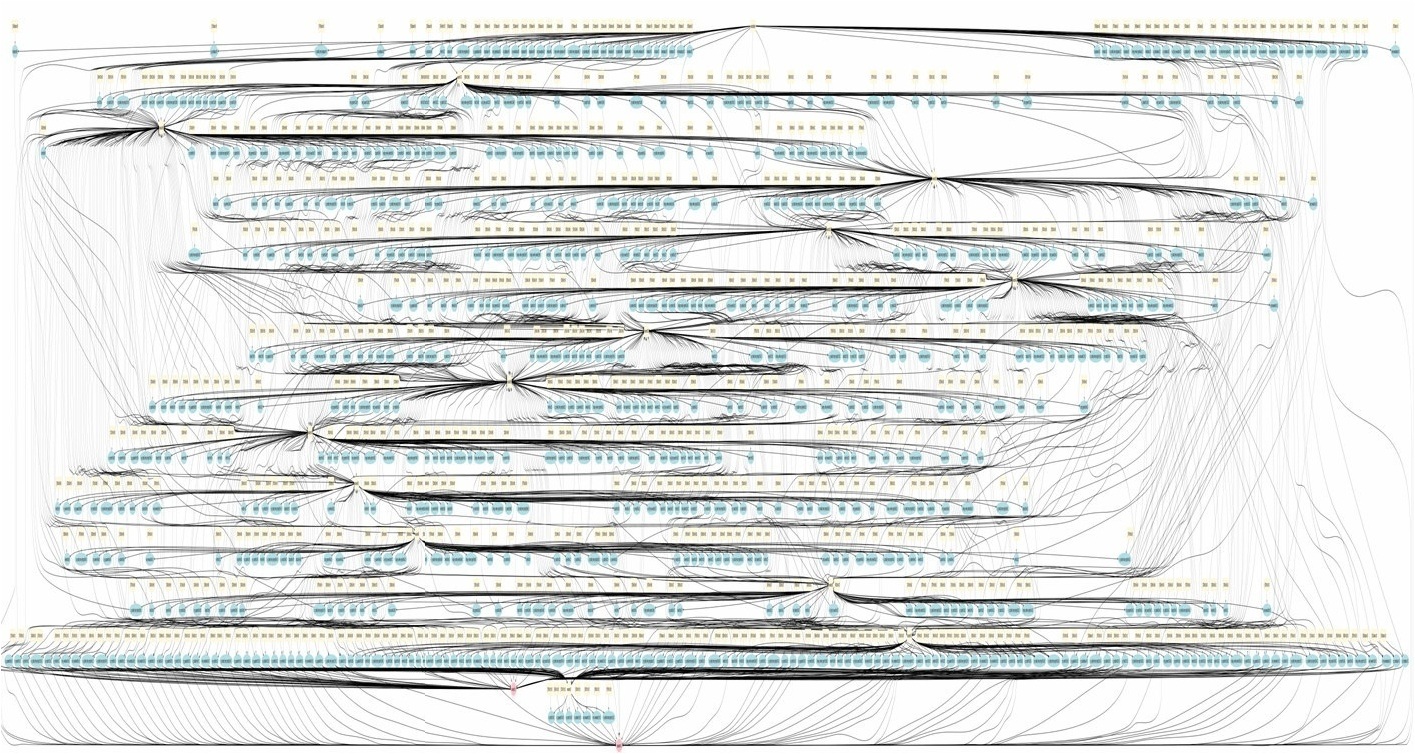}{!ht}{\linewidth}{attack_graph2}
{Attack Graph for a small network (14 hosts).}

In \cite{SwiPhiGay98} and \cite{JhaWinShe02} the authors propose using attack graphs 
to determine the security of networked systems. 
There are two main differences with our model. Firstly, the system they propose is 
an analysis system, which requires as input a great amount of information:
a database of common attacks broken down into atomic steps, specific network configuration
and topology information, and an attacker profile. Our model is a system
for building attacks, starting with possibly zero information about the network,
and gathering this information as the attack takes place.
Secondly, the attack graph we construct for planning purposes differs from
the attack graph of \cite{SwiPhiGay98} and \cite{JhaWinShe02}. In particular, it has a smaller size,
which allow us to effectively find a plan for real world instances of the problem.
Additionally we introduce several new dimensions to the graph, like quantified goals, 
probabilistic assets and a complex cost function.

\section{Summary}

In this chapter we presented a conceptual model 
to evaluate the costs of an attack, to describe the theater
of operations, targets, missions, actions, plans and assets involved in 
cybernetic attacks.

A contribution of this model concerns the costs of the actions.
We show that the cost is given by a tuple a values: not only the
probability of success, but also the stealthiness (which depends
on the noise produced), time consumed, non traceability and
zero-dayness. 
The noise produced is particularly relevant, and when we first published these ideas,
we had not seen it in other models\footnote{More recent work does take the \emph{noise}
into account, for instance the paper \cite{ElsKohMen11} published at the SecArt workshop in 2011.}.
These dimensions, considered
as attack parameters, also allow us to model different types of attackers.

The most important application of this model is automated planning.
Integrated in a tool like Core Impact, it leads the way to automated
penetration testing. Used against simulated networks, it is a 
tool for evaluating the vulnerabilities of a network.

Working on this model has opened a lot of questions and
directions for future work:
how to estimate the probability of success and noise produced by
an action, how to modify these values after an execution,
how to combine the different dimensions of the cost in order
to obtain a total order between costs,
how to choose the agents who will execute the actions,
when to create a new agent on a specific host,
how to decide the profiles (or personalities) of the agents,
the use of planning techniques, 
and the applications of the simulation scenario.
It also led us to review current penetration testing practices
and opened new dimensions for planning attacks and creating new tools.


\chapter{Planning Representations} \label{chap:representations}

Along this thesis, we will develop several representations of 
the attack model that we informally presented in Chapter \ref{chap:model}.
To understand and formulate precisely the differences between
those representations, we will consider them in the general model
of {\em state-transition systems} and enunciate the assumptions that are relaxed in 
each case (in Section~\ref{sec:state-transition-system}).

Another important point of this chapter
is how the attack model of Chapter \ref{chap:model}
can be represented in the PDDL language.
To define with precision this representation, we start
by recalling in Section~\ref{sec:classical-representations} 
the simpler representations based on set theory 
and first order logic.
We then provide in Section~\ref{sec:pddl-representation} a detailed description
of the PDDL representation of the attack planning problem.

We conclude this chapter with two exercises concerning the expressivity
of this representation and the complexity of the corresponding planning problem.

\section{Basic Formalization} \label{sec:state-transition-system}

To formalize the attack model, we will use a general model of dynamic systems, 
the model of {\em state-transition systems} \cite{GhaNauTra04}.
We give below the basic definition, then give a list of common assumptions,
and discuss how they can be relaxed to obtain different versions of the model.

\subsection{Formalization as a Dynamic System}

\begin{definition}
In general, a state-transition system is a tuple
$ \Sigma = \langle \cS, \cA, \cE, \gamma \rangle $ where:

\begin{itemize}

\item{ $ \cS = \{ s_1, s_2, \ldots \} $ is a finite set of states; }

\item{ $ \cA = \{ a_1, a_2, \ldots \} $ is a finite set of actions; }

\item{ $ \cE = \{ e_1, e_2, \ldots \} $ is a finite set of events; }

\item{ $ \gamma : \cS \times \cA \times \cE \rightarrow 2^{\cS} $ is a state-transition function.}

\end{itemize}
\end{definition}

A state represents a {\em``state of the world''},
which is the combined knowledge that the agents have
of the environment (network topology, operating systems,
running services, agents distribution, etc).

Actions are transitions that are controlled by the plan executor.
If $a \in \cA$ is an action and $\gamma(s,a) \neq \emptyset$,
the action $a$ is {\em applicable} to state $s$.

Events are transitions that are contingent: instead of being controlled
by the plan executor, they correspond to the internal dynamics of the system.

\subsection{General Assumptions}\label{sec:assumptions}

The following is a list of common assumptions made when 
modeling planning problems. These assumptions restrict the
complexity of the problem. 

\begin{description}

\item[Assumption A1 (Fully Observable $\Sigma$).] \hfill \\
The system $\Sigma$ is fully observable, the attacker has 
complete knowledge about the state of $\Sigma$.

\item[Assumption A2 (Deterministic $\Sigma$).] \hfill \\
The system $\Sigma$ is deterministic. For every state $s$ and every action $a$,
$ | \gamma (s,a) | \leq 1$.
If an action is applicable to a state, its application brings a deterministic
system to a single other state.

\item[Assumption A3 (Static $\Sigma$).] \hfill \\
The system $\Sigma$ is static if the set of events $\cE$ is empty.
The system stays in the same state until the controller applies some action.

\item[Assumptions A4 (Restricted Goals).] \hfill \\
The planner handles only {\em restricted goals} 
that are specified as an explicit goal state $s_g$ or set of goal states $S_g$.
The objective is any sequence of state transitions that ends at one of the
goal states.
Extended goals such as states to be avoided and constraints on state trajectories
or utility functions are nor handled under this assumption.

\item[Assumptions A5 (Sequential Plans).] \hfill \\
A solution plan to a planning problem is a linearly ordered finite sequence of actions.

\item[Assumptions A6 (Implicit Time).] \hfill \\
Actions and events have no duration. In state-transition systems time is not
explicitly represented.

\item[Assumptions A7 (Offline Planning).] \hfill \\
The planner is not concerned with any change that may occur in $\Sigma$ {\em while}
it is planning.
\end{description}

\subsection{Extending the Model} \label{sec:relaxing-assumptions}

Several models can be obtained by relaxing some of these restrictive assumptions.
In the development of the thesis, we will make the following relaxations
to obtain different versions of the model.

\begin{description}
\item[Relaxing Assumption A1 (Fully Observable $\Sigma$).] \hfill \\
	If the system is partially observable, then the observation of $\Sigma$
	will not fully disambiguate which state $\Sigma$ is in.
	For each observation $o$, ther may be more than one state $s$ 
	that produce the observation $o$.
	This situation will be studied with the model based on
	Partially Observable Markov Decision Processes (POMDP) in Chapter \ref{chap:pomdp}.

\item[Relaxing Assumption A2 (Deterministic $\Sigma$).] \hfill \\
  In a nondeterministic system, each action can lead to different possible states,
  so the planner may have to consider alternatives.	 
  Usually nondeterminism requires assumption A5 as well, because a plan
  must encode ways for dealing with alternatives.
  Nondeterministic systems will be studied in Chapter \ref{chap:probabilistic}
  about probabilistic algorithms and Chapter \ref{chap:pomdp} about
  the POMDP model.

\item[Relaxing Assumption A4 (Restricted Goals).] \hfill \\
  Extended goals allow one to specify to the planner requirements not only
  about the final state, but also on the states traversed, for example
  critical states to be avoided, states that the system should go through,
  states it should stay in, and other constraints on its trajectories.
  Another possibility is to specify a utility function to be optimized.
  This is used in the POMDP model, wherein the objective of the planner
  is to maximize the expected reward.
  
\item[Relaxing Assumption A5 (Sequential Plans).] \hfill \\
  A plan may be a mathematical structure that can be richer than a 
  simple sequence of actions.
  Relaxing assumption A5 is often required when other assumptions are relaxed,
  as we have seen in the case of nondeterministic systems (assumption A2).
  In the POMDP model, the output of the planner is a policy that specifies
  which action to execute for each belief state.
  
\item[Relaxing Assumption A6 (Implicit Time).] \hfill \\
  To handle time explicitly will require an extension of the model:
  in the PDDL model (Chapter \ref{chap:deterministic}) the execution
  time of an action $a \in \cA$ is specified as a numerical effect of $a$.
  
  In the POMDP model (Chapter \ref{chap:pomdp}), the execution time of an action $a$ is encoded
  in the reward function $r(s,a)$.
  
\end{description}

\subsection{Number of States}

We give below some estimations of the size of the state space $\cS$
in several examples.

\begin{example}
To get a grasp on the number of states involved, 
let's begin with a very simple example.
Let $\mathfrak{S}$ be a scenario with $M$ machines. 
Suppose that the only information that the attacker may obtain 
about each machine is its OS information. The only action that the attacker
can execute with each machine, is to recognize its Operating System.
That is, for each machine $m$, the attacker may have the OperatingSystemAsset for $m$ or not.
The number of states in this scenario is
$$
| \cS | = 2^M.
$$
\end{example}

\begin{example}
To make the previous example slightly more realistic, suppose that for each machine,
there are $k$ bits of information that the attacker may obtain: 
the knowledge of the Operating System and of $k-1$ applications that may be running on that machine.
The number of states is
$$
| \cS | = (2^k)^M
$$
which is exponential on the number of machines.
\end{example}

\begin{example}
Suppose now that the attacker can learn which Operating System version is running on a host, 
the applications running (and their versions), and the connectivity between machines. 
Let $O_s$ be the set of Operating Systems for each machine (including the ``unknown" operating system), $A_{pp}$ the set of applications
and $V_s$ the possible versions of applications (to simplify, we suppose them
to be uniform for all the applications). Then the number of states is
$$
| \cS | = ( | O_s | \times | V_s | ^ { | A_{pp} | }  ) ^ M \times 2 ^ { M \times M }.
$$
\end{example}

As these examples show, representing explicitly all the states of a network, even in a simple scenario,
becomes quickly unfeasible when the number $M$ of machines grow. 
{\em Implicit} representations are needed, that will enable us
to describe subsets of $\cS$ in a way that can be easily searched.

Another approach that enabled us to deal with the exponential growth of
the number of states (as a function of $M$) was to decompose the problem 
into different levels of abstraction,
and search for approximate solutions. We will present such 
ideas in Chapter~\ref{chap:probabilistic} (used for a probabilistic planner)
and in Chapter~\ref{chap:pomdp} (used for a planner based on POMDPs).

\section{Classical Planning Representations} \label{sec:classical-representations}

\Figure{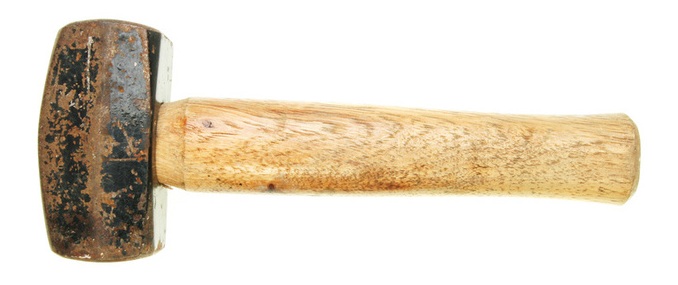}{!ht}{11cm}{hammer}{Picture of a Classical Planner.}

Classical planning, also called STRIPS planning,
refers to planning in a restricted state-transition system.
The importance of classical planning is given by the fact
that several general purpose and efficient planners exist for this setting.
In this sense, classical planners can be considered as the ``hammers'' of planning,
which are more suitable for real-world applications \cite{Geffner10}.

\begin{definition}
A {\em restricted state-transition system} is one
that meets the restrictive assumptions $\langle \text{A1}, \ldots, \text{A7} \rangle$
given in Section \ref{sec:assumptions}.
It is a deterministic, static, and fully observable state-transition system
with restricted goals and implicit time.
Such a system is denoted $\Sigma = \langle \cS, \cA, \gamma \rangle$.
Since the system is deterministic, if $a$ is applicable to $s$ then
$\gamma(s,a)$ contains one state $s'$. 
To simplify notations, we will say that $\gamma(s,a) = s'$ instead of $\gamma(s,a) = \{ s'\}$.
The transition function is thus noted as:
\begin{align*}
\gamma : \cS \times \cA &\rightarrow \cS \\
\gamma(s,a) &=
\begin{cases}
 s' &\mbox{if } a \mbox{ is applicable to } s \\
 \mbox{undefined} & \mbox{otherwise}
\end{cases}
\end{align*}
\end{definition}

\begin{definition} \label{def:planning-problem}
A {\em planning problem} for a restricted state-transition system 
$\Sigma = \langle \cS, \cA, \gamma \rangle$
is defined as a triple $\cP = \langle \Sigma, s_0, S_g \rangle$,
where
\begin{itemize}
	\item $s_0 \in \cS$ is the initial state,
	\item $S_g \subseteq \cS$ corresponds to a set of goal states. 
\end{itemize}
\end{definition}

\noindent
Note that the initial state $s_0 \in \cS$ corresponds to what we called the ``environment 
knowledge'' of the attacker in Chapter~\ref{chap:model}.

\begin{problem}(Planning in the restricted model)
\noindent
Given $\Sigma = \langle \cS, \cA, \gamma \rangle$,
an initial state $s_0$ and a set of goal states $S_g$,
find a sequence of actions $( a_1, a_2, \ldots, a_k )$
corresponding to a sequence of state transitions $(s_0, s_1, \ldots, s_k)$
such that $s_1 = \gamma(s_0, a_1), \ldots, s_k = \gamma(s_{k-1}, a_k)$,
and $s_k \in S_g$. 
\end{problem}

\subsection{Set-Theoretic Representation} \label{sec:set-theoretic-representation}

The set-theoretic representation uses only a finite set of proposition
symbols that represent propositions about the world.
This representation is more concise and readable than 
enumerating all the states and transitions explicitly.
It can be precisely formulated, and provides the foundation for the
classical representation. We recall the basic definitions from \cite{GhaNauTra04}.

\begin{definition} \label{def:planning-domain}
Let $ L = \{ p_1, \ldots, p_n \} $ be a finite set of proposition symbols.
A {\em set-theoretic planning domain} is a restricted state-transition system
$\Sigma = \langle \cS, \cA, \gamma \rangle$ such that:

\begin{itemize}

\item{ $ \cS \subseteq 2^L $, i.e, each state $s$ is a subset of $L$.
The state $s$ tells us which propositions currently hold.
If $p \in s$, then $p$ holds in the state of the world represented by $s$. }
\item{ Each action $a \in \cA$ is a triple of subsets of $L$. 
We will write it as
\begin{equation*}
a = ( \pre(a), \del(a), \add(a) )
\end{equation*}}
\item The set $\pre(a)$ is called the {\em preconditions} of $a$.
\item The set $\del(a)$ is called the {\em delete effects} of $a$,
and the set $\add(a)$ is called the {\em add effects} of $a$. 
We require that $\del(a) \cap \add(a) = \emptyset$.
\item{ An action $a \in \cA$ is applicable to a state $s \in \cS$ if $\pre(a) \subseteq s $.}
\item{ For every $s \in \cS$ and action $a$ applicable to $s$, the set
$ ( s - \del(a)) \cup \add(a) \in \cS $.}
\item{ The state-transition function is
\begin{equation*}
\gamma(s,a) = 
\begin{cases}
( s - \del(a)) \cup \add(a)  &\mbox{if }  a \mbox{ is applicable to } s \\
\mbox{undefined} &\mbox{otherwise}.
\end{cases}
\end{equation*}
}
\end{itemize}
\end{definition}

\noindent
Note that the effects of an action $a \in \cA$ (the add effects and delete effects)
correspond to what we called the action goal (in Section~\ref{sec:action-goal})
and the noise produced by the action (in Section~\ref{sec:action-noise-produced}).
The preconditions of an action $a$ correspond to the requirements
of the action (as defined in Section~\ref{sec:action-requirements}).

\begin{example}\label{set-theoretic-example}
This is a very small example involving only two hosts {\tt h1} and {\tt h2}.
The attacker can detect the operating system of these hosts,
and install an agent if he has already detected the OS.
He can also uninstall agents.

\medskip
\noindent
$L$ = \{ {\tt win-h1}, {\tt linux-h2}, {\tt agent-h1}, {\tt agent-h2} \} where \\
\indent {\tt win-h1} means that (the attacker detected that) host {\tt h1} is running windows; \\
\indent {\tt linux-h2} means that host {\tt h2} is running linux; \\
\indent {\tt agent-h1} means that host {\tt h1} has been compromised (agent installed); \\
\indent {\tt agent-h2} means that host {\tt h2} has been compromised.

\medskip
\noindent
$\cA$ = \{ {\tt detect-h1}, {\tt detect-h2}, {\tt install-h1}, {\tt install-h2},
 {\tt uninstall-h1}, \\ {\tt uninstall-h2} \} where: \\
\indent {\tt detect-h1} = ( \{ \}, \{ \}, \{ {\tt win-h1} \} ); \\
\indent {\tt detect-h2} = ( \{ \}, \{ \}, \{ {\tt linux-h2} \} ); \\
\indent {\tt install-h1} = ( \{ {\tt win-h1} \}, \{ \}, \{ {\tt agent-h1} \} ); \\
\indent {\tt install-h2} = ( \{ {\tt linux-h2} \}, \{ \}, \{ {\tt agent-h2} \} ); \\
\indent {\tt uninstall-h1} = ( \{ {\tt agent-h1} \}, \{ {\tt agent-h1} \}, \{  \} ); \\
\indent {\tt uninstall-h2} = ( \{ {\tt agent-h2} \}, \{ {\tt agent-h2} \}, \{  \} ).

\medskip
\noindent
$\cS = \{ s_0, \ldots, s_8 \} $ where: \\
\indent $s_0 = $ \{ \}; \\
\indent $s_1 = $ \{ {\tt win-h1} \}; \\
\indent $s_2 = $ \{ {\tt linux-h2} \}; \\
\indent $s_3 = $ \{ {\tt win-h1}, {\tt linux-h2} \}; \\
\indent $s_4 = $ \{ {\tt win-h1}, {\tt agent-h1} \}; \\
\indent $s_5 = $ \{ {\tt linux-h2}, {\tt agent-h2} \}; \\
\indent $s_6 = $ \{ {\tt win-h1}, {\tt linux-h2}, {\tt agent-h1} \}; \\
\indent $s_7 = $ \{ {\tt win-h1}, {\tt linux-h2}, {\tt agent-h2} \}; \\
\indent $s_8 = $ \{ {\tt win-h1}, {\tt linux-h2}, {\tt agent-h1}, {\tt agent-h2} \}.

\end{example}

\begin{definition}
A {\em set-theoretic planning problem} is a triple $ \calP = \langle \Sigma, s_0, g \rangle $ where:
\begin{itemize}
\item{$ s_0 \in \cS $ is the {\em initial state}.
}
\item{$ g \subseteq L $ is a set of propositions called {\em goal propositions} that
give the requirements that a state must satisfy to be a goal state.
The set of goal states is $ S_g = \{ s \in \cS \; | \; g \subseteq s \}$.
}
\end{itemize}

\end{definition}

\noindent
Note that the initial state $s_0 \in \cS$ corresponds to what we called the ``environment 
knowledge'' of the attacker in Chapter~\ref{chap:model}.

\begin{definition} \label{def:attack-plan}
A {\em plan} is any sequence of actions $ \pi = \langle a_1, \ldots, a_k \rangle $, where $ k \geq 0$.
The length of the plan is $ | \pi | = k $.
\end{definition}

\noindent
The state produced by applying $\pi$ to a state $s$ is the state produced
by applying the actions of $\pi$ in the given order. We extend
the state-transition function as follows:
$$
\gamma(s, \pi) = 
\begin{cases}
s & \mbox{if } k = 0 \\
\gamma( \gamma(s, a_1), \langle a_2, \ldots, a_k \rangle ) & \mbox{if } 
  k > 0 \mbox{ and } a_1 \mbox{ is applicable to } s \\
\mbox{undefined} & \mbox{otherwise}
\end{cases}
$$

\begin{definition}
Let $ \cP = \langle \Sigma, s_0, g \rangle $ be a planning problem.
A plan $\pi$ is a {\em solution} for $\cP$ if $ g \subseteq \gamma(s_0, \pi) $.
A solution $\pi$ is {\em minimal} if no other solution plan for $\cP$
contains fewer actions than $\pi$.
\end{definition}

\subsubsection{State Reachability}

We give below some basic definitions and facts concerning state reachability.

\begin{definition}
If $s$ is a state, then the set of all {\em successors} of $s$ is
\begin{equation*}
\Gamma(s) = \left\{ \gamma(s,a) \; | \; a \in \cA \text{ and } a \text{ is applicable to } s  \right\}
\end{equation*}
\end{definition}

\begin{definition}
Let $\Gamma^2(s) = \Gamma(\Gamma(s)) = \bigcup \{ \Gamma(s') \; | \; s' \in \Gamma(s) \} $,
and similarly for $\Gamma^3, \ldots, \Gamma^n$, then the set of states
{\em reachable} from $s$ is the transitive closure:
\begin{equation}
\hat{\Gamma} (s) = \Gamma(s) \cup \Gamma^2(s) \cup \Gamma^3(s) \cup \ldots
\end{equation}
\end{definition}

\begin{definition}
An action $a$ is said to be {\em relevant} for a goal $g$ if and only if
$g \cap \add(a) \neq \emptyset$ and $g \cap \del(a) = \emptyset$.
In other words, these conditions state that $a$ can contribute to producing
a state in $S_g = \{ s \in \cS \; | \; g \subseteq s \}$.
\end{definition}

\begin{definition}
The {\em regression set} of a goal $g$, for an action $a$ that is relevant for $g$,
is the minimal set of propositions required in a state in order to apply $a$ and get $g$:
\begin{equation}
\gamma^{-1} (g,a) = (g - \add(a)) \cup \pre(a)
\end{equation}
Thus for any state $s$,
\begin{equation}
\gamma(s,a) \in S_g \; \iff \; \gamma^{-1}(g,a) \subseteq s
\end{equation}
\end{definition}

\subsection{First Order Logic Representation} \label{sec:first-order-representation}

This representation generalizes the set-theoretic representation
using first-order logic notation.
We start with a first-order language $\calL$ in which
there is a finite number of predicate symbols and constant symbols.
A state is a set of ground atoms of $\calL$.

\begin{definition}
A planning operator $o$ is a triple (name($o$), precond($o$), effects($o$))
whose components are:

\begin{itemize}
\item{
The name of the operator, name($o$), is a syntactic expression of the form
$ n ( x_1, \ldots, x_k ) $ where $n$ is a unique symbol
and $ x_1 \ldots, x_k $ are all the variable symbols that appear anywhere in $o$.
}
\item{
The preconditions and effects are generalizations of the set-theoretic
preconditions and effects: instead of being sets of propositions, they are
sets of literals (atoms and negations of atoms).
}
\end{itemize}
\end{definition}

\begin{definition}
For any set of literals $L$, $L^+$ is the set of all atoms in $L$
and $L^-$ is the set of atoms whose negations are in $L$.
If $o$ is an operator, then $\mbox{precond}^+(o)$ and $\mbox{precond}^-(o)$
are its positive and negative preconditions, respectively,
and $\mbox{effect}^+(o)$ and $\mbox{effect}^-(o)$ are its positive and negative effects, respectively.
\end{definition}

\begin{definition}
An {\em action} is any ground instance of a planning operator.
If $a$ is an action and $s$ is a state such that
$\mbox{precond}^{+}(a) \subseteq s $ and $ \mbox{precond}^{-}(a) \cap s = \emptyset $,
then $a$ is applicable to $s$ and the result of applying $a$ to $s$ is the state
$$
\gamma(s,a) = ( s - \mbox{effects}^{-}(a)) \cup \mbox{effects}^{+}(a) 
$$
\end{definition}

\begin{definition}
Let $\calL$ be a first-order language that has finitely many predicate symbols
and constant symbols. A {\em classical planning domain} in $\calL$
is a restricted state-transition system 
$\Sigma = (S, A, \gamma)$ such that:

\begin{itemize}

\item{ $ S \subseteq 2^{ \{ \mbox{\scriptsize all ground atoms of } \calL \} } $ .}

\item{ $ A = \{ \mbox{all ground instances of operators in } \calO \} $ .}

\item{ The state-transition function is $\gamma(s,a) = ( s - \mbox{effects}^{-}(a)) \cup \mbox{effects}^{+}(a) $ if $ a\in A $ is applicable to $ s \in S $ and undefined otherwise.
}

\item{ $S$ is closed under $\gamma$, that is for every $s \in S$ and action $a$ applicable to $s$, 
$ \gamma(s,a) \in S $.
}

\end{itemize}

\end{definition}

\section{The PDDL Representation in Detail}
\label{sec:pddl-representation}

We use an extension of the classical representation that
allows the use of typed variables and relations.
This extension improves the efficiency of a planning system
by reducing the number of ground instances it needs to create.
More concretely we use the PDDL planning language notation \cite{Der98,FoxLon03}.

The PDDL description language serves as the bridge between the pentesting
tool and the planner. Since exploits have strict platform and connectivity
requirements, failing to accurately express those requirements in the PDDL
model would result in plans that cannot be executed against real networks.
This forces our PDDL representation of the attack planning problem to be
quite verbose.

On top of that, we take advantage of the optimization abilities of planners
that understand numerical effects\footnote{Numerical effects allow the
actions in the PDDL representation to increase the value of different metrics
defined in the PDDL scenario. The planner can then be told to find a plan
that minimizes a linear function of these metrics.}, and have the PDDL
actions affect different metrics commonly associated with penetration
testing such as running time, probability of success or possibility of
detection (stealth).

We will focus on the description of the {\em domain.pddl} file, which
contains the PDDL representation of the attack model. We will not delve
into the details of the {\em problem.pddl} file, since it consists of a
list of networks and machines, described using the predicates to be presented
in this section.

The PDDL requirements of the representation are {\bf :typing}, so that
predicates can have types, and {\bf :fluents}, to use numerical effects.
We will first describe the types available in the model, and then list
the predicates that use these types. We will continue by describing the
model-related actions that make this predicates true, and then we will
show an example of an action representing an exploit. We close this section
with an example PDDL plan for a simple scenario.

\subsection{Types}

Table \ref{tab:types} shows a list of the types that we used. Half of
the object types are dedicated to describing in detail the operating
systems of the hosts, since the successful execution of an exploit
depends on being able to detect the specifics of the OS.

\begin{table}[ht]
\begin{center}
\begin{tabular}{|l|l|}
\hline
network     & operating\_system \\
host        & OS\_version \\
port        & OS\_edition \\
port\_set   & OS\_build \\
application & OS\_servicepack \\
agent       & OS\_distro \\
privileges  & kernel\_version \\
\hline
\end{tabular}
\normalsize
\end{center}
\caption{List of object types}
\label{tab:types}
\end{table}

\subsection{Predicates}

The following are the predicates used in our model of attacks. Since exploits also
have non-trivial connectivity requirements, we chose to have a detailed
representation of network connectivity in PDDL. We need to be able to express
how hosts are connected to networks, and the fact that exploits may need
both IP and TCP or UDP connectivity between the source and target hosts, usually on
a particular TCP or UDP port. These predicates express the different forms
of connectivity:

{ \small \begin{verbatim}
(connected_to_network ?s - host ?n - network)
(IP_connectivity ?s - host ?t - host)
(TCP_connectivity ?s - host ?t - host ?p - port)
(TCP_listen_port ?h - host ?p - port)
(UDP_listen_port ?h - host ?p - port)
\end{verbatim} 
}

These predicates describe the operating system and services of a host:

{ \small \begin{verbatim}
(has_OS ?h - host ?os - operating_system)
(has_OS_version ?h - host ?osv - OS_version)
(has_OS_edition ?h - host ?ose - OS_edition)
(has_OS_build ?h - host ?osb - OS_build)
(has_OS_servicepack ?h - host ?ossp - OS_servicepack)
(has_OS_distro ?h - host ?osd - OS_distro)
(has_kernel_version ?h - host ?kv - kernel_version)
(has_architecture ?h - host ?a - OS_architecture)
(has_application ?h - host ?p - application)
\end{verbatim} 
}

\subsection{Actions}

We require some ``model-related'' actions that make true the aforementioned
predicates in the right cases.

{ \small \begin{verbatim}
(:action IP_connect
:parameters (?s - host ?t - host)
:precondition (and (compromised ?s)
  (exists (?n - network)
    (and (connected_to_network ?s ?n) 
      (connected_to_network ?t ?n))))
:effect (IP_connectivity ?s ?t)
)

(:action TCP_connect
:parameters (?s - host ?t - host ?p - port)
:precondition (and (compromised ?s)
  (IP_connectivity ?s ?t)
  (TCP_listen_port ?t ?p))
:effect (TCP_connectivity ?s ?t ?p)
)

(:action Mark_as_compromised
:parameters (?a - agent ?h - host)
:precondition (installed ?a ?h)
:effect (compromised ?h)
)
\end{verbatim}
}

Two hosts on the same network possess IP connectivity, and two hosts have
TCP (or UDP) connectivity if they have IP connectivity and the target host
has the correct TCP (or UDP) port open. Moreover, when an exploit is successful an
\emph{agent} is installed on the target machine, which allows control over
that machine. An installed agent is hard evidence that the machine is vulnerable,
so it marks the machine as compromised\footnote{Depending on the exploit
used, the agent might have regular user privileges, or superuser ({\em root})
privileges. Certain local exploits allow a low-level (user) agent to be
upgraded to a high-level agent, so we model this by having two different
\emph{privileges} PDDL objects.}.

The penetration testing framework we used has an extensive test suite that collects
information regarding running time for many exploit modules. We obtained
average running times from this data and used that information as the numeric
effect of exploit actions in PDDL. The metric to minimize in our PDDL scenarios
is therefore the total running time of the complete attack.

Finally, the following is an example of an action: an exploit that will attempt
to install an agent on target host {\em t} from an agent previously installed
on the source host {\em s}. To be successful, this exploit requires that
the target runs a specific OS, has the service {\em ovtrcd} running and
is listening on port 5053.

{ \small \begin{verbatim}
(:action HP_OpenView_Remote_Buffer_Overflow_Exploit
:parameters (?s - host ?t - host)
:precondition (and (compromised ?s)
  (and (has_OS ?t Windows)
    (has_OS_edition ?t Professional)
    (has_OS_servicepack ?t Sp2)
    (has_OS_version ?t WinXp)
    (has_architecture ?t I386))
  (has_service ?t ovtrcd)
  (TCP_connectivity ?s ?t port5053)
)
:effect(and (installed_agent ?t high_privileges)
 	(increase (time) 10)
))
\end{verbatim} 
}

In our PDDL representation there are several versions of this exploit, one
for each specific operating system supported by the exploit. For example,
another supported system for this exploit looks like this:

{ \small \begin{verbatim}
(:action HP_OpenView_Remote_Buffer_Overflow_Exploit
:parameters (?s - host ?t - host)
:precondition (and (compromised ?s)
  (and (has_OS ?t Solaris)
    (has_OS_version ?t V_10)
    (has_architecture ?t Sun4U))
  (has_service ?t ovtrcd)
  (TCP_connectivity ?s ?t port5053)
)
:effect(and (installed_agent ?t high_privileges)
 	(increase (time) 12)
))
\end{verbatim} 
}

The main part of the {\em domain.pddl} file is devoted to the description
of the actions. In our sample scenarios, this file has up to 28,000
lines and includes up to 1,800 actions. The last part of the {\em domain.pddl}
file is the list of constants that appear in the scenario, including the
names of the applications, the list of port numbers and operating system
version details.

\subsection{An Attack Plan}

We end this section with an example plan obtained by running \Mff on a
scenario generated with this model. The goal of the scenario is to compromise
host 10.0.5.12 in the target network. This network is similar to the test
network that we will describe in detail in Section \ref{sec:integration-performance}. The plan
requires four pivoting steps and executes five different exploits
in total. The \emph{localagent} object
represents the pentesting framework running in the machine of the user/attacker.
The exploits shown are real-world exploits
currently present in the pentesting framework.

{ \small \begin{verbatim}
 0: Mark_as_compromised localagent localhost
 1: IP_connect localhost 10.0.1.1
 2: TCP_connect localhost 10.0.1.1 port80
 3: Phpmyadmin Server_databases Remote Code Execution
        localhost 10.0.1.1
 4: Mark_as_compromised 10.0.1.1 high_privileges
 5: IP_connect 10.0.1.1 10.0.2.2
 6: TCP_connect 10.0.1.1 10.0.2.2 port80
 7: PHP memory_limit exploit
        10.0.1.1 10.0.2.2
 8: Mark_as_compromised 10.0.2.2 high_privileges
 9: IP_connect 10.0.2.2 10.0.3.2
10: SNMPc Network Manager Trap Packet Remote Buffer 
        Overflow 10.0.2.2 10.0.3.2
11: Mark_as_compromised 10.0.3.2 high_privileges
12: IP_connect 10.0.3.2 10.0.4.2
13: Snort TCP Stream Integer Overflow
        10.0.3.2 10.0.4.2
14: Mark_as_compromised 10.0.4.2 high_privileges
15: IP_connect 10.0.4.2 10.0.5.12
16: TCP_connect 10.0.4.2 10.0.5.12 port445
17: Novell Client NetIdentity Agent Buffer Overflow
        10.0.4.2 10.0.5.12
18: Mark_as_compromised 10.0.5.12 high_privileges
\end{verbatim} 
}

\section{Expressivity of the PDDL Representation}

Choosing to use the PDDL representation to model our problem turned out to be
a very happy choice. First of all, because the PDDL language was created for the
International Planning Competition, and using it we could experiment with a 
set of different planners designed to take PDDL input.
Secondly because the PDDL language has been widely extended \cite{FoxLon03,YouLit04}, 
and the set of problems that be represented using this language is very wide.
We will illustrate its expressivity with a simple exercise:
to model the Attack Trees proposed by Bruce Schneier \cite{Schneier99}.

\begin{example}
Schneier's article \cite{Schneier99} takes as an example a simple attack tree against a physical safe.
The attack is represented in a tree structure, with the goal as the root node and different ways 
of achieving that goal as leaf nodes. 
There are AND nodes and OR nodes:
the OR nodes are alternatives,
and the AND nodes represent different steps toward achieving the same goal.

This representation does not distinguish between actions and assets.
Let us model the same attack in the PDDL language, using predicates for the assets.

The actions of this example are described follow. The leaf nodes of the tree
are marked as ``possible'' or ``impossible'' in the attack tree,
this is translated as preconditions, where (possible) is a true predicate
in the initial conditions of the problem.

\begin{multicols}{2}

{ \small \begin{verbatim}
(:action PickLock
:precondition (impossible)
:effect (open_safe)
)

(:action UseLearnedCombo
:precondition (safe_combination)
:effect (open_safe)
)

(:action CutOpenSafe
:precondition (possible)
:effect (open_safe)
)

(:action InstallImproperly
:precondition (impossible)
:effect (open_safe)
)

(:action FindWrittenCombo
:precondition (impossible)
:effect (safe_combination)
)

(:action GetComboFromTarget
:precondition 
   (safe_combination_from_target)
:effect (safe_combination)
)

(:action Threaten
:precondition (impossible)
:effect (safe_combination_from_target)
)

(:action Blackmail
:precondition (impossible)
:effect (safe_combination_from_target)
)

(:action Eavesdrop
:precondition (and
   (conversation_eavesdropped)
   (target_states_combo)
)
:effect (safe_combination_from_target)
)

(:action Bribe
:precondition (possible)
:effect (safe_combination_from_target)
)

(:action ListenToConversation
:precondition (possible)
:effect (conversation_eavesdropped)
)

(:action GetTargetToStateCombo
:precondition (impossible)
:effect (target_states_combo)
)
\end{verbatim} 
}
\end{multicols}

The assets, formulated as predicates, are:
{ \small \begin{verbatim}
(:predicates 
   (possible)
   (impossible)
   (open_safe)
   (safe_combination)	
   (safe_combination_from_target)
   (conversation_eavesdropped)
   (target_states_combo)
)
\end{verbatim} 
}

This is the domain definition of the problem. The initial conditions 
and the goal are:

{ \small \begin{verbatim}
(define (problem attack_tree_figure1) 

(:domain AttackTree)

(:init (possible))

(:goal (open_safe))
)
\end{verbatim} 
}

When executing the planner FF with this problem, we get the following plan
{ \small \begin{verbatim}
ff: search configuration is  best-first on 1*g(s) + 5*h(s) where
    metric is  plan length

advancing to distance:    1
                          0

ff: found legal plan as follows

step    0: CutOpenSafe
\end{verbatim} 
}
\end{example}

\begin{example}
Let's see a continuation of the previous example. Consider that we distinguish 
between actions that require special equipment and those which don't.
We also add a cost to each leaf action (in thousands of dollars).
This is done via the cost function and adding numerical effects 
to the actions.
The domain with these modifications is now:

\begin{multicols}{2}
{ \small \begin{verbatim}
(define (domain AttackTree)

(:requirements :TYPING :FLUENTS)

(:functions 
   (cost) 
)

(:predicates 
   (none)
   (special_equipment)
   (open_safe)
   (safe_combination)	
   (safe_combination_from_target)
   (conversation_eavesdropped)
   (target_states_combo)
)

(:action PickLock
:precondition (special_equipment)
:effect (and
   (open_safe)
   (increase (cost) 30))
)

(:action UseLearnedCombo
:precondition (safe_combination)
:effect (open_safe)
)

(:action CutOpenSafe
:precondition (special_equipment)
:effect (and
   (open_safe)
   (increase (cost) 10))
)

(:action InstallImproperly
:precondition (none)
:effect (and
   (open_safe)
   (increase (cost) 100))
)

(:action FindWrittenCombo
:precondition (none)
:effect (and
   (safe_combination)
   (increase (cost) 75))
)

(:action GetComboFromTarget
:precondition 
   (safe_combination_from_target)
:effect (and
   (safe_combination)
   (increase (cost) 0))
)

(:action Threaten
:precondition (none)
:effect (and
   (safe_combination_from_target)
   (increase (cost) 60))
)

(:action Blackmail
:precondition (none)
:effect (and
   (safe_combination_from_target)
   (increase (cost) 100))
)

(:action Eavesdrop
:precondition (and
   (conversation_eavesdropped)
   (target_states_combo))
:effect (safe_combination_from_target)
)

(:action Bribe
:precondition (none)
:effect (and
   (safe_combination_from_target)
   (increase (cost) 20))
)

(:action ListenToConversation
:precondition (special_equipment)
:effect (and
   (conversation_eavesdropped)
   (increase (cost) 20))
)

(:action GetTargetToStateCombo
:precondition (none)
:effect (and
   (target_states_combo)
   (increase (cost) 40))
)
)
\end{verbatim} 
}
\end{multicols}

And the initial conditions and the goal of the problem are now:
{ \small \begin{verbatim}
(define (problem attack_tree_figure1) 

(:domain AttackTree)

(:init
   (none)
   (= (cost) 0)
)

(:goal (open_safe))

(:metric MINIMIZE (cost))
)
\end{verbatim} 
}
In particular we suppose that the attacker hasn't got the special equipment required by some actions,
and the objective is to obtain the goal (open\_safe) while minimizing the cost function.

The output of the planner Metric-FF is the following plan:
{ \small \begin{verbatim}
ff: search configuration is  best-first on 1*g(s) + 5*h(s) where
    metric is ((1.00*[RF0](cost)) - () + 0.00)

advancing to distance:    1
                          0

ff: found legal plan as follows

step    0: FindWrittenCombo
        1: UseLearnedCombo
\end{verbatim} 
}
It is noteworthy that this plan doesn't minimize the total cost, since the given plan has a
cost of \$ 75k, whereas the optimal plan (below) has a cost of \$ 20k.
{ \small \begin{verbatim}
step    0: Bribe
        1: GetComboFromTarget
        2: UseLearnedCombo
\end{verbatim} 
}
This plan was not found by Metric-FF because in its heuristic search,
the planner only considered plans with 2 actions (which are sufficient to obtain the goal)
before looking for an optimal plan within that horizon.
\end{example}

To conclude the example, it showed that Schneier's Attack Trees can be perfectly
modeled in PDDL with fluents, and it also showed us a limitation of Metric-FF,
that failed to find the optimal plan even in this very small example.

\section{The Hypergraph Representation}

In this section we explore briefly the Hypergraph representation,
and use it to prove that a version of the Attack Planning
Problem is NP-hard. Of course this comes as no surprise,
and the following section can thus be considered as an exercise
performed during the early phases of our research.

\subsection{Basic concepts}

We recall some basic definitions.

\begin{definition}
A \textit{directed hypergraph}, or simply \textit{hypergraph},
is a pair $ H = (V,E) $ where $V$ is the set of nodes,
and $E$ is the set of hyperarcs.
A \textit{hyperarc} is a pair $ e = (T(e), h(e)) $,
where $ T(e) \subseteq V $ is the tail of $e$ 
while $h(e) \in V$ is the head of $e$.
\end{definition}

The nodes represent the attack assets.
The actions are modeled as hyperarcs. 
The assets contained in the tail $T(e)$ of an action are the requirements.
The head of the hyperarc is the result of the action.

Action templates are instantiated.

\begin{definition}
A \textit{directed hyperpath}, or simply \textit{hyperpath}, $P_{St}$
from the source set $S \subseteq V$ to the target node $t \in V$
is a minimal acyclic sub-hypergraph of $H$ containing the nodes in $S$ and the node $t$,
such that each node (except the nodes in $S$) has exactly one entering hyperarc.
\end{definition}

\subsection{NP-Hardness of the Hypergraph Formulation}

Suppose that the hypergraph $H$ is given.
The planning problem is to find an optimal hyperpath.
The following problem is NP-hard \cite{FelRuh99}.

\begin{problem} (Directed Steiner Tree problem)

\noindent
Instance: A directed graph $G = (V,E)$, 
a node $s \in V$, a subset of nodes $K = \{i_1,\dots, i_k\} \subseteq V$ 
and weights $w: A \longrightarrow \Z_{+}$.

\noindent
Task: Find a directed subtree $T$, rooted at node $s$ and of minimum weight 
that contains all nodes in $K$.
\end{problem}

We state the following version of the Attack Planning problem.

\begin{problem} (Attack Planning, hypergraph formulation)

\noindent
Instance: A directed hypergraph $G = (V,E)$, nodes $s$, $t$
and weights $w: E \longrightarrow  \Z_{+}$.

\noindent
Task: Find a subhypergraph $G_1$ of minimum weight that contains 
a directed hyperpath from node $s$ to node $t$.

We note that if $G_1$ contains a hyperarc $(\{i_1,\dots,i_k\},j)$, 
it must contain nodes $\{i_1,\dots,i_k, j\}$.
\end{problem}

\begin{theorem} The version of Attack Planning Problem stated above 
is NP-hard.
\end{theorem}
\begin{proof}
The Attack Planning Problem contains the Directed Steiner Tree Problem
as a special case. More precisely, given an instance $(G, s, K, c)$ 
of the Directed Steiner Tree Problem 
we construct the following hypergraph $G^{\prime}$:
we add an extra node $t$ to graph $G$ and we add a hyperarc 
$(\{i_1,\dots,i_k\}, t)$ from all nodes in $K$ to node $t$ of zero cost.
That is, let $V^{\prime} = V \cup \{t\}$, 
let $A^{\prime} = A\cup \{(\{i_1,\dots,i_k\}, t)\}$ 
and let $G^{\prime} = (V^{\prime}, A^{\prime})$.  
The cost $c^{\prime}$ of each arc of $A^{\prime}$ 
is the same as before and the cost of the hyperarc is zero.

Then, solving the Attack Planning Problem on 
$(G^{\prime}, s, t, c^{\prime})$ implies solving 
the Directed Steiner Tree Problem on $(G, s, K, c)$. 
\end{proof}


\chapter{Integration with Classical Planners} \label{chap:deterministic}

Assessing network security is a complex and difficult task. Attack graphs
have been proposed as a tool to help network administrators understand
the potential weaknesses of their networks 
(see Section~\ref{sec:attack-graphs-related-work}). 
However, one problem has not yet
been addressed by previous work on this subject; namely, how to actually
execute and validate the attack paths resulting from the analysis of
the attack graph. In Chapter \ref{chap:representations} we presented 
a complete PDDL representation of an attack model.
In the rest of this chapter, we present an implementation that integrates a planner into
a penetration testing tool. 

This allows us to automatically generate
attack paths for penetration testing scenarios, and to validate these attacks
by executing the corresponding actions -including exploits- against the
real target network. We present an algorithm for transforming the information
present in the penetration testing tool to the planning domain 
in Section \ref{sec:transform-architecture}, and we show
how the scalability issues of attack graphs can be solved using current
planners. In Section \ref{sec:integration-performance} we make an analysis 
of the performance of our solution, showing
how the model scales to medium-sized networks and the number of actions
available in current penetration testing tools.

\section{Attack Planning in the Real World}

In medium-sized networks, building complete attack graphs quickly becomes
unfeasible (their size increases exponentially with the number of machines
and available actions). To deal with the attack planning problem, 
a proposed approach \cite{SarWei08,Sarraute09} is to translate the model
into a PDDL representation and use classical planning algorithms to find attack paths.
Planning algorithms manage to find paths in the attack graph without constructing
it completely, thus helping to avoid the combinatorial explosion \cite{BluFur97}.
A similar approach was presented at SecArt'09 \cite{GhoGho09}, but the authors'
model is less expressive than the one used in this work, as their objective
was to use the attack paths to build a minimal attack graph, and not to
carry out these attacks against real networks.

In the following sections we present an implementation of these ideas. 
We have developed
modules that integrate a pentesting framework with an external planner,
and execute the resulting plans back in the pentesting framework, against
a real network. We believe our implementation proves the feasability of
automating the attack phases of a penetration test, and allows us to think
about continuing this line of work in order to automate the whole process. 
We show how our model, and
its PDDL representation, scales to hundreds of target nodes and available
exploits, numbers that can be found when assessing medium-sized networks
with current pentesting frameworks.

The rest of the chapter is structured as follows: 
in Section \ref{sec:transform-architecture} we
present a high-level description of our solution, describing the steps needed
to integrate a planner with a penetration testing framework. 
We have already described the PDDL representation in detail
in Section \ref{sec:pddl-representation}
explaining how the ``real world''
view that we take forces a particular implementation of the attack planning
problem in PDDL. Section \ref{sec:integration-performance} presents the results of our
scalability testing, showing how our model manages medium-sized networks
using current planners.

\section{Architecture of our Solution}
\label{sec:transform-architecture}

In this section we describe the components of our solution, and how
they fit together to automate an attack. Figure \ref{fig:architecture}
shows the relationship between these different components. The {\em penetration
testing framework} is a tool that allows the user/attacker to execute exploits
and other pre/post exploitation modules against the target network. Our
implementation is based on Core Impact\footnote{As mentioned in the previous
section, Metasploit is an open-source alternative.}. The \emph{planner}
is a tool that takes as input the description of a \emph{domain} and a
\emph{scenario}, in PDDL\footnote{Refer to \cite{FoxLon03} for a description
of the PDDL planning language.}. The domain contains the definition of
the available actions in the model, and the scenario contains the definition
of the objects (networks, hosts, and their characteristics), and the goal
which has to be solved.

\Figure{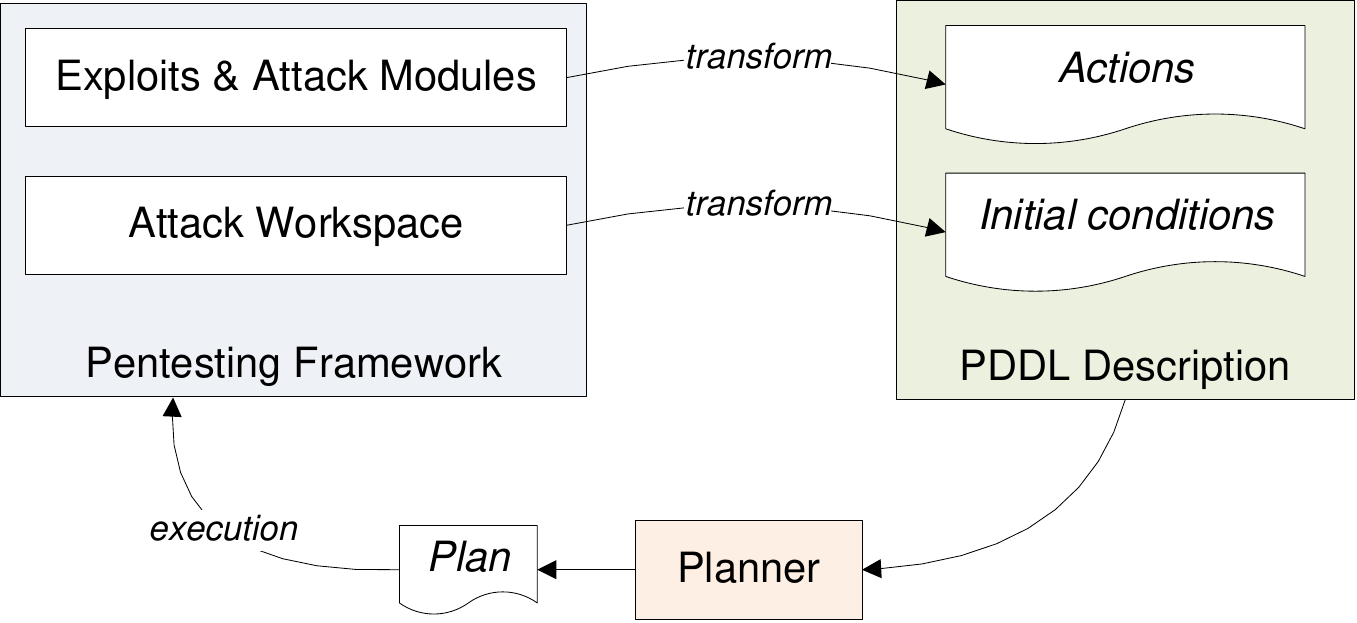}{t}{0.90 \linewidth}{architecture}
{Architecture of our solution.}

The {\em attack workspace} contains the information about the current attack
or penetration test. In particular, the discovered networks and hosts,
information about their operating systems, open/closed ports, running services
and compromised machines. In the current version of our solution we assume
that the workspace has this network information available, and that no
network information gathering is needed to generate a solvable plan. We
will address this limitation in Section \ref{sec:conclusion} when we discuss
future work.

\subsection{The ``Transform'' Algorithm} \label{sec:transform-algorithm}

The {\em transform} algorithm generates the PDDL representation of the
attack planning problem, including the initial conditions, the operators
(PDDL actions), and the goal. From the pentesting framework we extract the description of
the operators, in particular the requirements and results of the exploits,
which will make up most of the available actions in the PDDL representation.
This is encoded in the {\em domain.pddl} file, along with the predicates
and types (which only depend on the details of our model).

From the attack workspace we extract the information that constitutes the
initial conditions for the planner: networks, machines, operating systems,
ports and running services. This is encoded in the {\em problem.pddl} file,
together with the goal of the attack, which will usually be the \emph{compromise}
of a particular machine.

A common characteristic of pentesting frameworks is that they
provide an incomplete view of the network under attack. The pentester has
to infer the structure of the network using the information that he sees
from each compromised machine. The \emph{transform} algorithm takes this
into account, receiving extra information regarding host connectivity.

\subsection{The Planner}

The PDDL description is given as input to the {\em planner}.
The advantage of using the PDDL language is that we can experiment with
different planners and determine which best fits our particular problem.
We have evaluated our model using both \Sg \cite{CheWahHsu06} and \Mff \cite{Hoffmann02}.

The planner is run from inside the pentesting framework, as a pluggable
module of the framework that we call \emph{PlannerRunner}. The output of
the planner is a {\em plan}, a sequence of actions that lead
to the completion of the goal, if all the actions are successful. We make
this distinction because even with well-tested exploit code, not all exploits
launched are successful. The plan is given as feedback to the pentesting
framework, and executed against the real target network.

\section{Performance and Scalability Evaluation}
\label{sec:integration-performance}

This model, and its representation in PDDL, are intended to be used to
plan attacks against real networks, and execute them using a pentesting
framework. To verify that our proposed solution scales up to the domains
and scenarios we need to address in real-world cases, we carried out extensive
performance and scalability testing -- to see how far we could take the
attack model and PDDL representation with current planners. We focused
our performance evaluation on four metrics:

\begin{itemize}

\item Number of machines in the attacked network

\item Number of pivoting steps in the attack

\item Number of available exploits in the pentesting suite

\item Number of individual predicates that must be fulfilled to accomplish the goal

\end{itemize}

The rationale behind using these metrics is that we needed our solution
to scale up reasonably with regard to all of them. For example, a
promising use of planning algorithms for attack planning lies in
scenarios where there are a considerable number of machines to
take into account, which could be time-consuming for a human attacker.

Moreover, many times a successful penetration test needs to reach the
innermost levels of a network, sequentially exploiting many machines in
order to reach one which might hold sensitive information.
We need our planning solution to be able to handle these cases where
many pivoting steps are needed.

Pentesting suites are constantly updated with exploits for new
vulnerabilities, so that users can test their systems against the latest
risks. The pentesting tool that we used currently\footnote{As of March, 2010.}
has about 700 exploits, of which almost 300 are the remote exploits that
get included in the PDDL domain. Each remote exploit is represented as
a different operator for each target operating system, so our PDDL domains
usually have about 1800 operators, and our solution needs to cope with
that input.

Finally, another promising use of planning algorithms for attack planning
is the continuous monitoring of a network by means of a constant pentest.
In this case we need to be able to scale to goals that involve compromising
many machines.

We decided to use the planners \Mff\footnote{Latest version available (with additional improvements).} \cite{Hoffmann02} 
and \Sg\footnote{\Sg version 5.22.} \cite{CheWahHsu06} since we
consider them to be representative of the state of the art in classical planners.
The original FF planner was the baseline planner for IPC'08\footnote{The
International Planning Competition, 2008.}. \Mff adds numerical effects
to the FF planner. We modified the reachability analysis in \Mff
to use type information, as in FF, to obtain better memory usage.

\Sg combines \Mff as a base planner with a constraint partitioning scheme
which allows it to divide the main planning problem in subproblems; these
subproblems are solved with a modified version of \Mff, and the individual
solutions combined to obtain a plan for the original problem. This method,
according to the authors, has the potential to significantly reduce the
complexity of the original problem \cite{CheWahHsu06}. It was successfully
used in \cite{GhoGho09}.

\subsection{Generating the Test Scenarios}

We tested both real and simulated networks, generating the test scenarios
using the same pentesting framework we would later use to attack them. For the
large-scale testing, we made use of a network simulator \cite{FutMirOrl09}.
This simulator allows us to construct sizable networks\footnote{We tested up to
1000 nodes in the simulator.}, but to still view each machine independently
and, for example, to execute distinct system calls in each of them. The simulator
integrates tightly with the pentesting framework, to the point where the
framework is oblivious to the fact that the network under attack is simulated
and not real.

This allowed us to use the pentesting tool to carry out all the steps of
the test, including the information gathering stage of the attack. Once
the information gathering was complete, we converted the attack workspace
to PDDL using our \emph{transform} tool.

\Figure{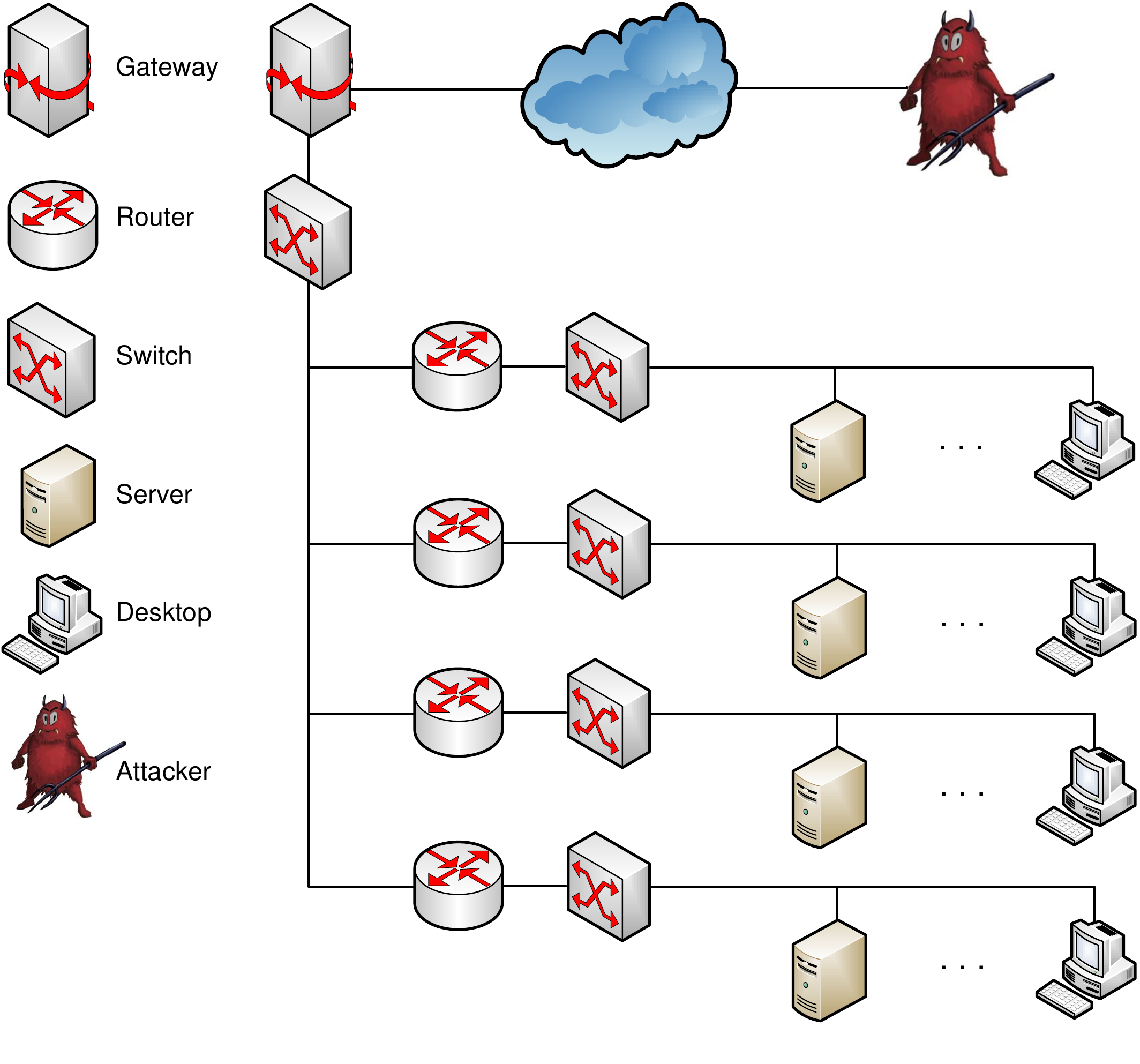}{!ht}{0.85 \linewidth}{test_net}
{Test network for scalability evaluation.}

We generated two types of networks for the performance evaluation. To
evaluate the scalability in terms of number of machines, number of
operators, and number of goals; the network consists of five subnets with
varying numbers of machines, all joined to one main network to which the
user/attacker initially has access. Figure \ref{fig:test_net} shows the
high-level structure of this simulated network.

To evaluate the scalability in terms of the number of pivoting steps
needed to reach the goal, we constructed a test network where the attacker
and the target machine are separated by an increasing number of routers,
and each subnetwork in between has a small number of machines.

\begin{table}[ht]
\begin{center}
\begin{tabular}{|l|c|c|c|}
\hline
\textbf{Machine type} & \textbf{OS version} & \textbf{Share} & \textbf{Open ports} \\
\hline
Windows desktop & Windows XP SP3 & 50\% & 139, 445 \\
\hline
Windows server & Windows 2003 Server SP2 & 14\% & 25, 80, 110, 139, \\
 & & & 443, 445, 3389 \\
 \hline
Linux desktop & Ubuntu 8.04 or 8.10 & 27\% & 22 \\
\hline
Linux server & Debian 4.0 & 9\% & 21, 22, 23, 25, \\
 & & & 80, 110, 443 \\
\hline
\end{tabular}
\normalsize
\end{center}
\caption{List of machine types for the test networks}
\label{tab:test_machines}
\end{table}

The network simulator allows us to specify many details about the simulated
machines, so in both networks, the subnetworks attached to the
innermost routers contain four types of machines: Linux desktops and servers,
and Windows desktops and servers. Table \ref{tab:test_machines} shows 
the configuration for each of the four machine types, and the share of each
type in the network. For server cases, each machine randomly removes one
open port from the canonical list shown in the table, so
that all machines are not equal and thus not equally exploitable.

\subsection{Experimental Results}

As we expected, both planners generated the same plans in all cases, not
taking into account plans in which goals were composite and the same actions
could be executed in different orders. This is reasonable given
that \Sg uses \Mff as its base planner. We believe that the performance
and scalability results are more interesting, since a valid plan for an
attack path is a satisfactory result in itself.

Figures \ref{fig:scale_time} to \ref{fig:goals_mem} show how running time
and memory consumption scale for both planners, with respect to the four
metrics considered\footnote{Testing was performed on a Core i5 750 2.67
GHz machine with 8 GB of RAM, running 64-bit Ubuntu Linux; the planners
were 32-bit programs.}. Recall that, as explained in Section \ref{sec:pddl-representation},
each exploit maps to many PDDL actions.

\Figure{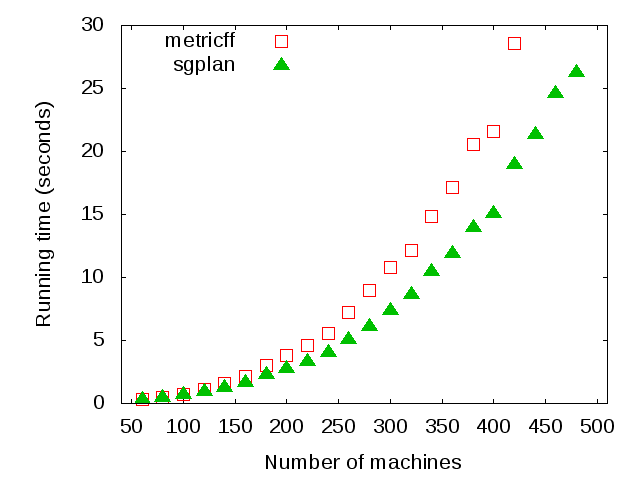}{!t}{11cm}{scale_time}
 {Running time, increasing number of machines.
 (Fixed values: 1600 actions, 1 pivoting step).}
\Figure{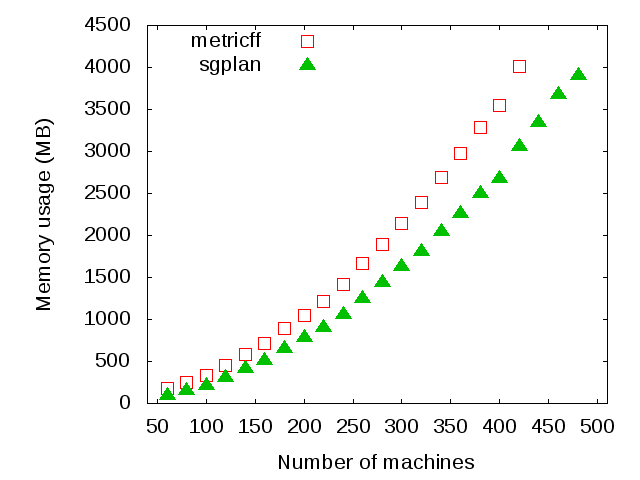}{!t}{11cm}{scale_mem}
 {Memory usage, increasing number of machines.}

As illustrated by Figures \ref{fig:scale_time} and \ref{fig:scale_mem},
both running time and memory consumption increase superlinearly with the
number of machines in the network. We were not able to find exact specifications
for the time and memory complexities of \Mff or \Sg, though we believe
this is because heuristics make it difficult to calculate a complexity
that holds in normal cases. Nonetheless, our model, coupled with the \Sg
planner, allows us to plan an attack in a network with 480 nodes in 25 seconds
and using less than 4 GB of RAM. This makes attack planning practical for
pentests in medium-sized networks.

Moving on to the scalability with regard to the \emph{depth} of the attack
(Figures \ref{fig:depth_time} and \ref{fig:depth_mem}), it was surprising
to verify that memory consumption is constant even as we increase the depth
of the attack to twenty pivoting steps, which generates a plan of more than
sixty steps. Running time increases slowly, although with a small number
of steps the behaviour is less predictable. The model is therefore not
constrained in terms of the number of pivoting steps.

\Figure{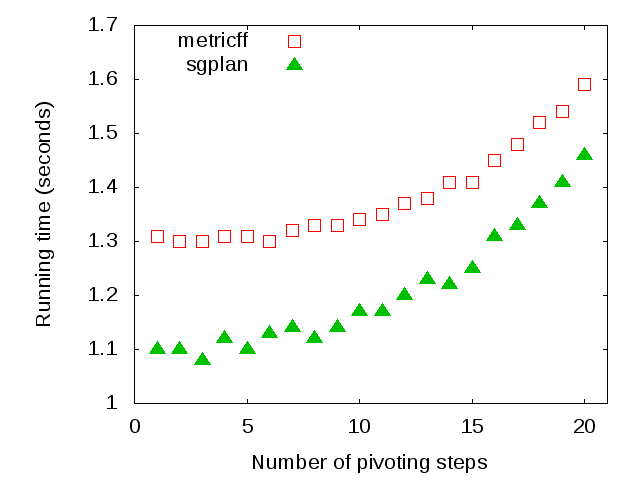}{p}{11cm}{depth_time}
 {Running time versus increasing number of pivoting steps.
 (Fixed values: 1600 actions, 120 machines).}
\Figure{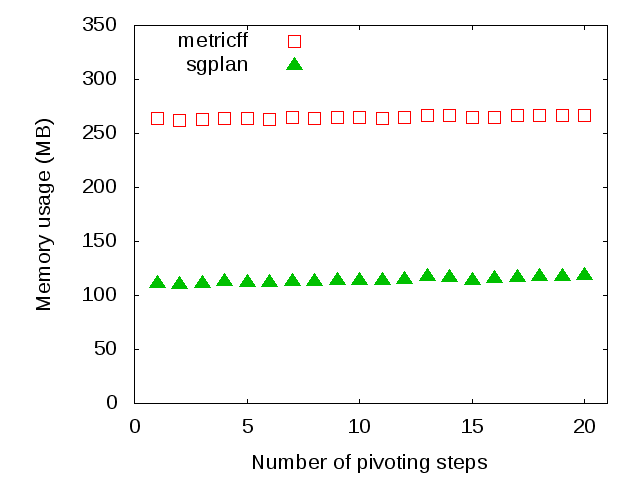}{p}{11cm}{depth_mem}
 {Memory usage versus increasing number of pivoting steps.}

With regard to the number of operators (i.e. exploits) (Figures \ref{fig:operators_time}
and \ref{fig:operators_mem}), both running time and memory consumption
increase almost linearly; however, running time spikes in the largest cases.
Doubling the number of operators, from 720 to 1440 (from 120 to 240 available exploits),
increases running time by 163\% for \Mff and 124\% for \Sg. Memory
consumption, however, increases only 46\% for \Mff, and 87\% for \Sg. In
this context, the number of available exploits is not a limiting factor
for the model.

Interestingly, these three tests also verify many of the claims made by
the authors of \Sg. We see that the constraint partition used by their planner
manages to reduce both running time and memory consumption, in some cases
by significant amounts (like in Figure \ref{fig:depth_mem}).

\Figure{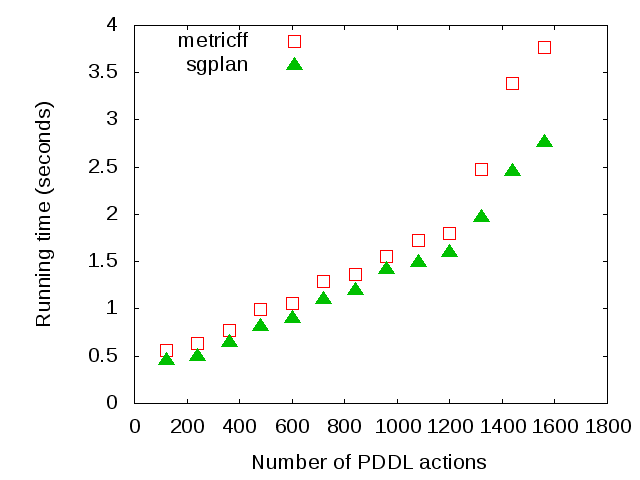}{p}{11cm}{operators_time}
 {Running time versus increasing number of actions.
 (Fixed values: 200 machines, 1 pivoting step).}
\Figure{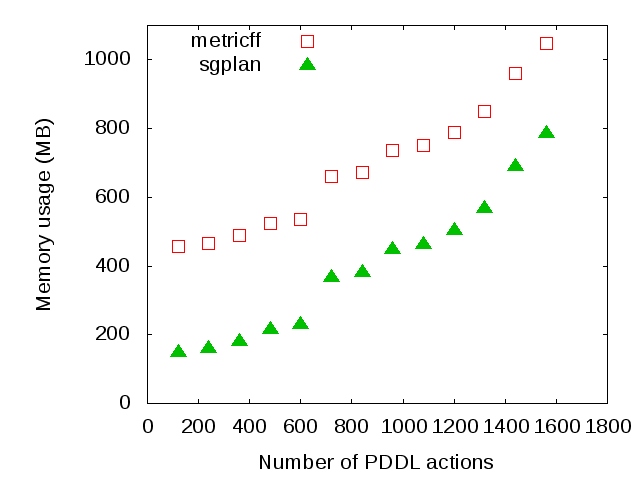}{p}{11cm}{operators_mem}
 {Memory usage versus increasing number of actions.}

The results for the individual number of predicates in the overall goal
(Figures \ref{fig:goals_time} and \ref{fig:goals_mem}) are much more surprising. 
While \Sg runs faster than \Mff in most of the cases, \Mff consumes significantly
less memory in almost half of them. We believe that as the goal gets
more complex (the largest case we tested requests the individual compromise
of 100 machines), \Sg's constraint partition strategy turns into a liability,
not allowing a clean separation of the problem into subproblems. By falling
back to \Mff, our model can solve, in under 6 seconds and using slightly more
than 1 GB of RAM, attack plans where half of the machines of a 200-machine
network are to be compromised.

\Figure{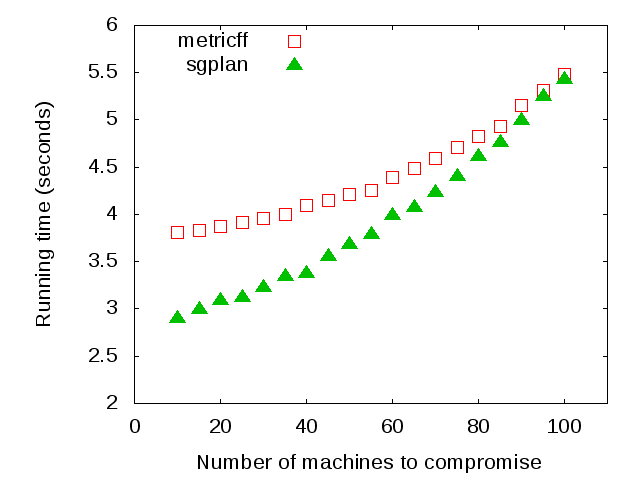}{p}{11cm}{goals_time}
 {Running time versus increasing number of predicates in the goal.
 (Fixed values: 200 machines, 1 pivoting step for each compromised machine, 1600 actions).}
\Figure{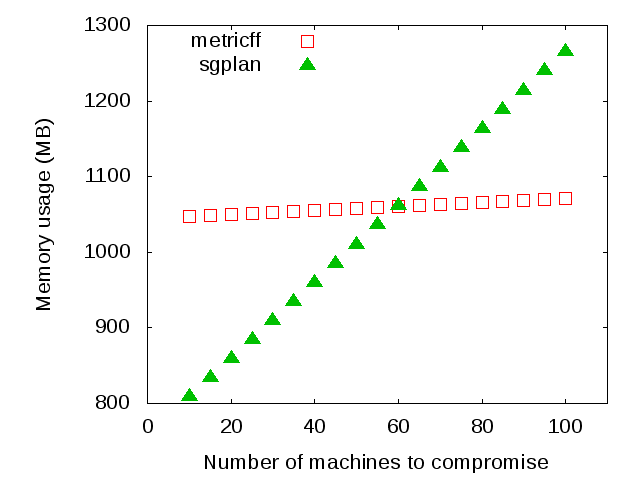}{p}{11cm}{goals_mem}
 {Memory usage versus increasing number of predicates in the goal.}

\subsection{Improving the Memory Usage of Metric-FF}

During our initial tests with \Mff, we were not able to
solve even medium-sized scenarios without running out of memory on a
machine with 4 GB of RAM. We discovered that \Mff was performing a very
inefficient reachability analysis before the actual planning, thus causing
said memory problems. More specifically, the reachability analysis allocated
four integer arrays with room for all the possible ``states'' of each predicate.
These states corresponded to all possible combination of constants for
each argument of the predicate. The problem was that \Mff was not using
type information, and therefore it considered \emph{all} constants, regardless
of type, for each argument of each predicate. Due to the fact that we need
to represent many characteristics of the target machine and operating system
for each exploit, our domain contains a lot of constants, representing
port numbers, operating system versions, editions, distributions, service
packs, and kernel versions.

\Mff allocated $\sum_{i = 1}^{p} (c^{a_i})$ 32-bit integers for each array,
where $p$ is the number of predicates (20 in our scenarios), $c$ is the
number of constants (around 300 in our scenarios), and $a_i$ is the arity
of predicate $i$ ($max(a_i) = 3$ in our scenarios). We modified \Mff to
only allocate space for constants of the correct type, using type information
as originally implemented in the planner FF, and obtained much better results.
The reachability analysis for our standard domain would consume over 4 GB
of RAM, and we managed to reduce that to a little over 2 MB, by significantly
reducing the number of constants associated with each argument of each predicate.

\section{Related Work}
\label{sec:deterministic-related}

Work on attack modeling applied to penetration testing had its origin in
the possibility of programmatically controlling pentesting tools such as
Metasploit or Core Impact. This model led to the use of attack graphs.
Earlier work on attack graphs such
as \cite{PhiSwi98,RitAmm00,SheHaiJha02} were based on the complete
enumeration of attack states, which grows exponentially with the number
of actions and machines. As we mentioned in Section \ref{sec:attack-graphs-related-work}
the survey of \cite{LipIng05} shows that the major limitations of past
studies of attack graphs is their lack of scalability to medium-sized networks.

One notable exception is the Topological Vulnerability Analysis (TVA) project conducted in 
George Mason University described in \cite{JajNoeBer05,NoeJaj05,NoeEldJaj09}
and other papers, which has been designed to work in real-size networks.
The main differences between our approach and TVA are the following:

\begin{itemize}

\item{ {\bf Input.} In TVA the model is populated with information from
third party vulnerability scanners such as Nessus, Retina and FoundScan,
from databases of vulnerabilities such as CVE and OSVDB and other software.
All this information has to be integrated, and will suffer from the drawbacks
of each information source, in particular from the false positives generated
by the vulnerability scanners about potential vulnerabilities.

In our approach the conceptual model and the information about the target
network are extracted from a consistent source: the pentesting framework exploits
and workspace. The vulnerability information of an exploit is very precise:
the attacker can execute it in the real network to actually compromise systems.
}

\item{ {\bf Output.} The attack graph generated by TVA is presented to the end-user
through an interactive graphical interface. Whereas in our approach, the
output of the planner is a sequence of actions that are directly executed
by the pentest tool on the target network, and is shown as an actual agent
installed in the target host.
}

\item{ {\bf Monotonicity.} TVA assumes that the attacker's control over
the network is monotonic \cite{AmmWijKau02}. In particular, this implies
that TVA cannot model Denial-of-Service (DoS) attacks, or the fact that
an unsuccessful exploit may crash the target service or machine. It is
interesting to remark that the monotonicity assumption is the same used
by FF \cite{Hoffmann01} to create a relaxed version of the planning
problem, and use it as a heuristic
to guide the search through the attack graph. By relying on the planner
to do the search, we do not need to make this restrictive assumption.
}
\end{itemize}

\section{Summary and Future Work}
\label{sec:conclusion}

\cite{FutNotRic03} proposed a model of computer network attacks which
was designed to be realistic from an attacker's point of view.
We have shown in this chapter that this model scales up to medium-sized networks:
it can be used to automate attacks (and penetration tests) against networks
with hundreds of machines.

The solution presented shows that it is not necessary to build the complete
attack graph (one of the major limitations of earlier attack graph studies).
Instead we rely on planners such as \Mff and \Sg to selectively explore
the state space in order to find attack paths.

We have successfully integrated these planners with a pentesting framework,
which allowed us to execute and validate the resulting plans against a
test bench of scenarios. We presented the details of how to transform the
information contained in the pentesting tool to the planning domain\footnote{Our
implementation uses Core Impact, but the same ideas can be extended to
other tools such as the open-source project Metasploit.}.

One important question that for future work on this subject is how
to deal with incomplete knowledge of the target network (actually
we deal with that issue in Chapters \ref{chap:probabilistic} and \ref{chap:pomdp}). 
The architecture that we presented supports running non-classical planners, so one possible approach
is to use probabilistic planning techniques, where actions have different
outcomes with associated probabilities. For example, a step of the attack
plan could be to discover the operating system details of a particular host,
so the outcome of this action would be modeled as a discrete probability
distribution.

Another approach would be to build a ``metaplanner'' that generates hypotheses
with respect to the missing bits of information about the network, and
uses the planner to test those hypotheses. Continuing the previous example,
the metaplanner would assume that the operating system of the host was
Windows and request the planner to compromise it as such. The metaplanner
would then test the resulting plan in the real network, and verify or discard
the hypothesis.


\chapter{Simulation of Network Scenarios} \label{chap:simulations}

In this chapter we present a simulation platform called \insight, created to design 
and simulate cyber-attacks against large arbitrary target scenarios.
This platform was used to test the implementation described in Chapter~\ref{chap:deterministic}, namely the integration of a classical planner
with a pentesting framework. To be able to test our implementation,
we needed to execute attack paths in networks with hundreds of machines,
something that could only be done on a simulated network (given our current infrastructure).

The simulation platform that we developed has surprisingly low hardware and configuration requirements, while making the simulation a realistic experience from the attacker's standpoint. The scenarios include a crowd of simulated actors: network devices, hardware devices, software applications, protocols, users, etc. 

A novel characteristic of this tool is to simulate vulnerabilities (including 0-days) and exploits, allowing an attacker to compromise machines and use them as pivoting stones to continue the attack. A user can test and modify complex scenarios, with several interconnected networks, where the attacker has no initial connectivity with the objective of the attack. 

We give a concise description of this new technology, and its possible uses in the security research field, such as pentesting training, study of the impact of 0-day vulnerabilities, evaluation of security countermeasures, and as a risk assessment tool
(besides providing a testbed for our planning implementation).

\section{Motivation}

Computer security has become a necessity in most of today's computer uses and practices.
However it is a wide topic and security issues can arise from almost everywhere: binary flaws 
(e.g., buffer overflows \cite{Aleph96}), 
Web flaws (e.g., SQL injection, remote file inclusion), protocol flaws (e.g., TCP/IP flaws \cite{Bellovin89}), 
not to mention hardware, human, cryptographic and other well known flaws.
 
Although it may seem obvious, it is useless to secure a network with a hundred firewalls 
if the computers behind it are vulnerable to client-side attacks. The protection provided by an 
{Intrusion Detection System} (IDS) is worthless against new vulnerabilities and 0-day attacks. 
As networks have grown in size, they have implemented a wider variety of more complex
configurations and include new devices (e.g. embedded devices) and technologies. 
This has created new flows of information and control, and therefore new attack vectors. 
As a result, the job of both black hat and white hat communities has become more difficult 
and challenging. 

The previous examples are just the tip of the iceberg.
Computer security is a complex field and
it has to be approached with a \emph{global} view, considering all parts of the 
\emph{whole} picture simultaneously: 
network devices, hardware devices, software applications, protocols, users, \emph{et cetera}. 
The simulation platform \insight has been created with that goal in mind.

\subsection{Design Restrictions}

In practice, the simulation of complex networks requires resolving the tension between the
\emph{scalability} and \emph{accuracy} of the simulated subsystems, devices and data.
This is a complex issue, and to find a satisfying solution for this trade-off 
we have adopted the following design restrictions:
\begin{enumerate}
	\item Our goal is to have a simulator on a single desktop computer, running hundreds 
of simulated machines, with simulated traffic realistic \emph{only} from the attacker's 
standpoint. 

\item Attacks within the simulator are \emph{not} launched by real attackers in the wild 
(e.g. script-kiddies, worms, black hats). As a consequence, the simulation does not have 
to handle exploiting details such as stack overflows or heap overflows. 
Instead, attacks are executed from an attack framework by \insight users who know 
they are playing in a simulated environment.
\end{enumerate}

\subsection{Main Features}

To demonstrate that our approach is valid, we have developed a proof of concept
called \insight. This program introduces a platform for executing 
attack experiments and tools for constructing these attacks.
We show that users of this tool are able to design 
and adapt attack-related technologies, and have better tests to 
assess their quality.
Attacks are executed from an attack framework which includes many 
information gathering and exploitation modules.
Modules can be scripted, modified or even added.

One of the major \insight features is the capability to simulate \emph{exploits}. 
An exploit is a piece of code that attempts to compromise a computer system via a specific vulnerability.
There are many ways to exploit security holes. If a computer programmer makes a programming 
mistake in a computer program, it is sometimes possible to circumvent security. 
Some common exploiting techniques are stack exploits, heap exploits,
format string exploits, etc. 

To simulate these techniques in detail is very expensive. 
The main problem is to maintain the complete state (e.g., memory, stack, heap, CPU registers) 
for every simulated machine. 
From the attacker's point of view, 
an exploit can be modeled as a magic string sent to a target machine to unleash a hidden feature 
(e.g., reading files remotely) with a probabilistic result. 
This is a lightweight approach, 
and we have sacrificed some of the realism in order to support     
very large and complex scenarios. For example, $1,000$ virtual machines and network devices 
(e.g., hubs, switches, IDS, firewalls) can be simulated on a single 
Windows desktop, each one running their own simulated OS, applications, vulnerabilities and file systems. 
Certainly, taking into account available technologies, it is not feasible to use a complete virtualization server (e.g., VMware) running thousands of images simultaneously. 

As a result, the main design concept of our implementation is 
to focus on the attacker's point of view, and \emph{to simulate on demand}. 
In particular, the simulator only generates information as requested by the attacker. By performing this 
on-demand processing, the main performance bottleneck comes from the ability of the attacker to request information 
from the scenario. Therefore, it is not necessary, for example, to simulate the complete TCP/IP packet traffic 
over the network if nobody is requesting that information. 
A more lightweight approach is to send data between network sockets writing in the memory 
address space of the peer socket, and leaving the full packet simulation as an option.

\section{Background and Related Work}

Using simulated networks as a research tool to gather knowledge regarding the techniques, 
strategy and procedures 
of the black hat community is not a new issue. Solutions such as \emph{honeypots} and 
\emph{honeynets} \cite{honeynet:2006, spitzner02} were primarily designed to attract malicious 
network activities and to collect data. 
A precise definition for the term honeypot is given 
by The Honeynet Project \cite{honeynet04}:

\begin{definition}
	A honeypot is an information system resource whose value 
	lies in unauthorized or illicit use of that resource.
\end{definition}

Over the last decade a wide variety of honeypot systems have been
built \cite{honeypot03, Bailey05theinternet, song01, yegneswaran04design, Provos04}, 
both academic and commercial. Honeypots have emerged as an 
interesting tool for gathering knowledge on new methods used by attackers, 
and the underlying strength of the approach lies in its simplicity. Typically, 
honeypots offer minimal interaction with the attacker, emulating only small 
portions of the behavior of real networks. However, this simplicity is also a weakness: 
none of these systems execute kernel or application code that attackers seek 
to compromise, and only a few ones maintain a per-flow and per-protocol state to 
allow richer emulation capabilities. Thus, honeypots are most useful for capturing 
indiscriminate or large-scale attacks, such as worms, viruses or botnets, rather than 
very focused intrusions targeting a particular host \cite{potemkin05}.

In Table \ref{insightvshoneypot} we show the main differences
with our approach. In particular, we are interested 
in the ability to compromise machines, and use them as 
\emph{pivoting stones}\footnote{ In a network attack, to \emph{pivot} means to use a compromised machine as a stepping stone to reach further networks and machines, making use of its trust relationships. } to build complex multi-step attacks.

\begin{table}[ht]
\begin{center}
\small
\begin{tabular}{ p{6cm} p{6cm} }
\textbf{Honeypot-like tools} & \textbf{\insight} \\
	\hline\hline
		Design focus: to detect, understand and monitor real cyber-attacks. & 
		Design focus: to reproduce or mimic cyber-attacks, penetration test training, what-if and 0-day scenarios. \\
	\hline
		Attacks are launched by real attackers: worms, botnets, script-kiddies, black-hats. & 
		Attacks are launched by the \insight users: pentest and forensic auditors, security researchers. \\
	\hline
		Simulation up to transport layer. & 
		Simulation up to application layer, including vulnerabilities and exploits. \\
	\hline
		Stateless or (a kind of) per-flow and per-protocol state. & 
		Applications and machines' internal states. \\
	\hline
		No exploit simulation. No pivoting support. & 
		Full exploit and agent (shellcode) simulation. Ability to pivot through a chain of 
		agents.\\
\end{tabular}
\end{center}
\caption{Honeypots vs. \insight.}
\label{insightvshoneypot}
\end{table}

\normalsize

In contrast, ``high interaction honeypots'' and virtualization technologies (e.g., VMware, Xen, Qemu) 
execute native system and application code, but the price of this 
fidelity is quite high. For example, the RINSE approach \cite{rinse05} is implemented over the iSSFNet network 
simulator, which runs on parallel machines to support real-time simulation of large-scale networks.
All these solutions share the same principle of simulating almost every aspect of a real machine or real network, but share similar problems too: expensive configuration cost 
and expensive hardware and software licenses. 
Moreover, most of these solutions are not fully compatible with standard network protection (e.g., firewalls, IDSs), suffering a lack of integration between all security actors in complex cyber-attack scenarios.

\emph{Security assessment} and \emph{staging} are other well known security practices. It is common, for example in web application development, to duplicate the production environment on a staging environment (accurately mimicking or mirroring the first) to anticipate changes and their impact. The downside is that it is very difficult to adopt this approach in the case of network security due to several reasons. It would require the doubling of the hardware and software licenses and (among other reasons) there are no means to automatically configure the network.

Other interesting approaches to solve these problems include the framework developed by Bye et al. \cite{byeschmidt2008}. While they focus on distributed denial of service attacks (DDoS) and defensive IDS analysis, we focus on offensive strategies to understand the scenarios and develop countermeasures. Also Loddo et al. \cite{loddosaiu2008} have integrated \emph{User Mode Linux} \cite{dike2006} and \emph{Virtual Distributed Ethernet} \cite{davoli2005} to create a flexible and very detailed network laboratory and simulation tool. The latter project has privileged accuracy and virtualization over scalability and performance.

The \emph{Potemkin Virtual Honeyfarm} \cite{potemkin05} is another interesting prototype. 
It improves high-fidelity honeypot scalability by up to six times while still closely emulating the execution behavior of individual Internet hosts. Potemkin uses quite sophisticated on-demand techniques for instantiating 
hosts\footnote{Including \emph{copy-on-write} file system optimizations implemented also in \insight, as we are going to see it in \S\ref{filesystem}.}, but this approach focuses on attracting real attacks and it shows the same honeypot limitations to reach this goal. 
As an example, to capture e-mail viruses, a honeypot must posses an e-mail address, must be scripted to read mail (executing attachments like a naive user) and, most critically, \emph{real e-mail users} must be influenced to add the honeypot to their address books. 
Passive malware (e.g., many spyware applications) may require a honeypot to generate explicit requests, and focused malware (e.g., targeting only financial institutions) may carefully select its victims and never touch a large-scale honeyfarm. In each of these cases there are partial solutions, and they require careful engineering to truly mimic the target environment.

In conclusion, new trends in network technologies make cyber-attacks more difficult to understand, 
learn and reproduce, and the current tools to tackle these problems 
have some deficiencies when facing large complex scenarios. 
In spite of that, it is possible to overcome 
the problems described above using the lightweight software simulation tool we present.

\section{Insight Approach and Overview}

\begin{figure}[th]
\begin{center}
\includegraphics[width=10cm]{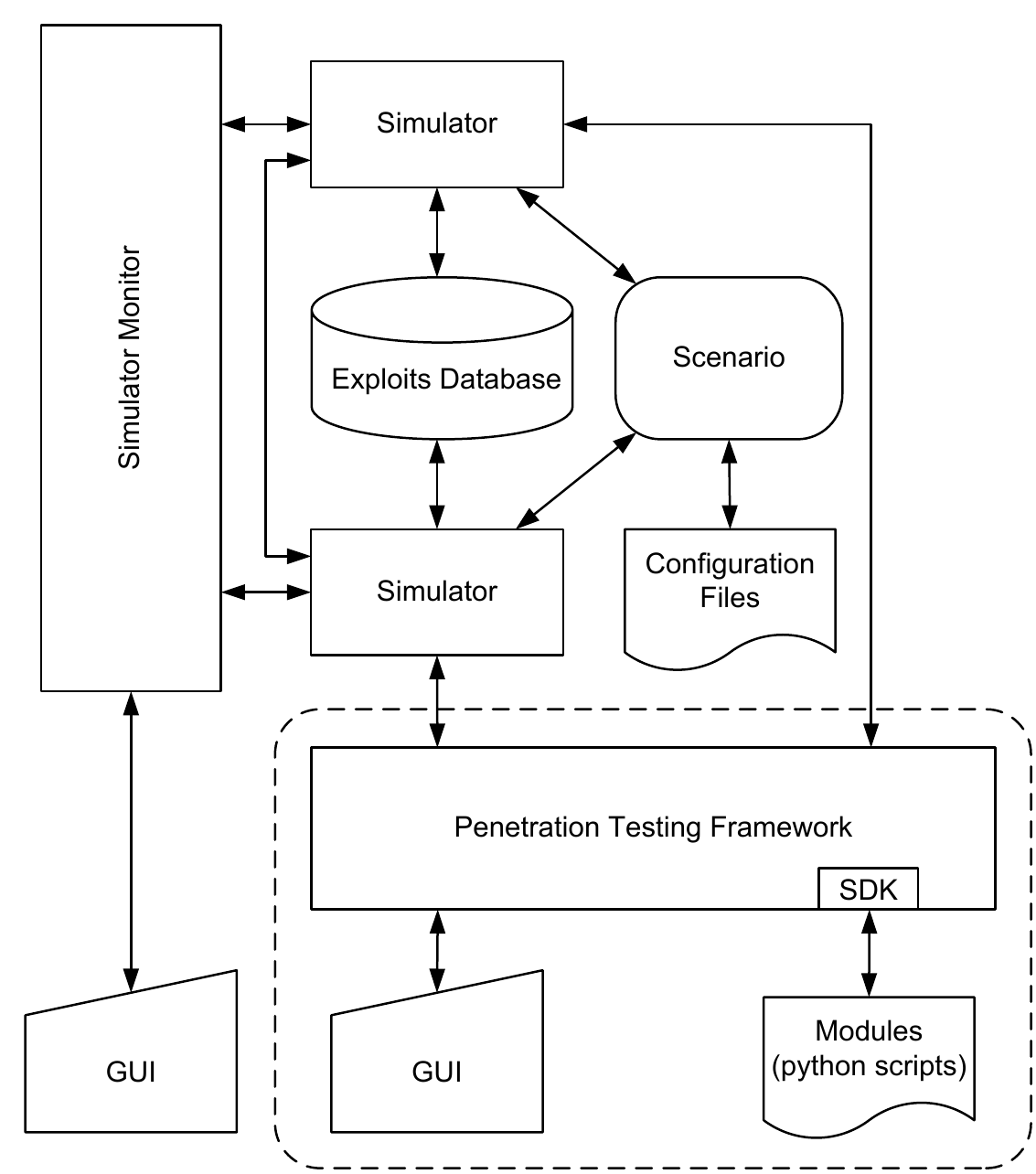} 
\end{center}
\caption{\insight architecture layout.}
\label{fig:arch}
\end{figure}

A diagram of the \insight general architecture is showed in \reffig{fig:arch}. 
The \component{Simulator} subsystem is the main component. It performs all 
simulation tasks on the simulated machines, such as system call execution, 
memory management, interrupts, device I/O management, etcetera. 

At least one \component{Simulator} 
subsystem is required, but the architecture allows several ones, each running 
in a real computer (e.g., a Windows desktop). In this example, there are two simulation subsystems, 
but more could be added in order to support more virtual hosts. 

The simulation proceeds in a lightweight fashion. It means, for example, that 
not all system calls for all OSes are supported by the simulation. Instead of 
implementing the whole universe of system calls, \insight handles a reduced and 
generic set of system calls, shared by all the simulated OSes. Using this approach, 
a specific OS system call is mapped to an \insight syscall which works similarly 
to the original one. For example, the Windows sockets API is based on the 
Berkeley sockets API model used in Berkeley UNIX, but both implementations are 
slightly different\footnote{For additional details look at the Winsock API documentation 
(available from \url{http://msdn.microsoft.com}), which includes a section called Porting 
Socket Applications to Winsock.}. Similarly, 
there are some instances where \insight sockets have to diverge from strict adherence 
to the Berkeley conventions, usually due to implementation difficulties in the 
simulated environment. In spite of this (and ignoring the differences between OSes), all 
sockets system calls of the real world have been mapped to this unique simulated API.

Of course, there are some system calls and management tasks closely 
related to the underlying OS which were not fully supported, such as UNIX {\sf fork} 
and {\sf signal} syscalls, or the complete set of functions implemented by the Windows SDK. 
There is a trade-off between precision and efficiency, and the decision of which syscalls 
were implemented was made with the objective of 
maintaining the precision of the simulation 
\emph{from the attacker's standpoint}.

The exploitation of binary vulnerabilities\footnote{\insight supports simulation for 
binary vulnerabilities. Other kind of vulnerabilities (e.g. client-side and SQL injections) will be implemented in future versions.} 
is simulated with a \emph{probabilistic} 
approach, to keep the attack model simple, lightweight, and to avoid tracking 
anomalous conditions (and their countermeasures), such as buffer overflows, 
format string vulnerabilities, exception handler overwriting---among other well known vulnerabilities 
\cite{AnlHeaLin07}. This probabilistic approach allows us to mimic 
the unpredictable behavior when an exploit is launched against a targeted machine.

Let us assume that a simulated computer was initialized with an 
underlying vulnerability (e.g. it hosts a vulnerable OS). 
In this case, the exploit payload is replaced by a 
special ID or ``magic string'', which is sent to the attacked 
application using a preexistent TCP communication channel. 
When the attacked application receives this ID, \insight will decide if the exploit 
worked or not based on a probability distribution that depends on the exploit 
and the properties describing the simulated computer (e.g., OS, patches, open services). 
If the exploit is successful, then \insight will grant the control 
in the target computer through the \emph{agent abstraction}, 
which will be described in \S \ref{sec:attack_model}.

The probabilistic attack model is implemented by the \component{Simulator} subsystems, 
and it is supported by the \component{Exploits Database}, a special configuration file 
which stores the information related to the vulnerabilities. This file has a XML tree structure, 
and each entry has the whole necessary information needed by the simulator to compute 
the probabilistic behavior of a given simulated exploit. For example, a given 
exploit succeeds against a clean XP SP2 with 83\% probability if port 21 is open, but 
crashes the system if it is a SP1. We are going to spend some time looking at the probability
distribution, how to populate the exploits database, and the \insight attack model in the next sections.

Returning to the architecture layout showed in \reffig{fig:arch}, all simulator subsystems are coordinated 
by a unique \component{Simulator Monitor}, which deals with management and administrative operations, 
including administrative tasks (such as starting/stopping a simulator instance) and providing statistical 
information for the usage and performance of these.

A set of \component{Configuration Files} defines the \emph{snapshot} of a virtual 
\component{Scenario}. Similarly, a scenario snapshot defines the instantaneous status 
of the simulation, and involves a crowd of simulated actors: servers, workstations, 
applications, network devices (e.g. firewalls, routers or hubs) and their present status. 
Even users can be simulated using this approach, and this is especially interesting in 
client-side attack simulation, where we expect some careless users opening 
our poisoned crafted e-mails. 

Finally, at the bottom right of the architecture diagram, we can see the 
\component{Penetration Testing Framework}, an external system which interacts with 
the simulated scenario in real time, sending system call requests through a communication channel 
implemented by the simulator. This attack framework is a free tailored version of the 
\impact solution\footnote{Available from \url{http://trials.coresecurity.com/.}}, however 
other attack tools are planned to be supported in the future (e.g., Metasploit 
\cite{Moore06}). 

The attacker actions are coded as \impact script files (using Python) 
called \emph{modules}, which have been implemented using the attack framework SDK, as shown 
in the architecture diagram. 
The framework Python modules include several tools for common tasks (e.g. information 
gathering, exploits, import scenarios). The attacks are executed in real 
time against a given simulated scenario; a simulation component can provide scenarios of thousands 
of computers with arbitrary configurations and topologies.  
\insight users can design new scenarios and they have scripts to manage the creation and modification of the 
simulated components, and therefore iterate, import and reproduce cyber-attack experiments.

\section{The Simulated Attack Model}\label{sec:attack_model}

One of the characteristics that distinguish the scenarios simulated by \insight is the ability to compromise machines, and use them as pivoting stones to build complex multi-step attacks.
To compromise a machine means to install an {agent} that will be able to execute arbitrary system calls (syscalls) as a user of this system.

The agent architecture is based on the solution called \emph{syscall proxy} 
(see \cite{Caceres02} for more details).
The idea of syscall proxying is to build a sort of \emph{universal payload} that allows an attacker
to execute any system call on a compromised host.
By installing a small payload (a thin syscall server) on a vulnerable machine,
the attacker will be able to execute complex applications on his local host,
with all system calls executed remotely. This syscall server is called an \emph{agent}.

In the \insight attack model, the use of syscall proxying introduces two additional layers between a process run by the attacker and the compromised OS. These layers are the \emph{syscall client} layer and the \emph{syscall server} layer.

The syscall client layer runs on the attacker's \component{Penetration Testing Framework}.
It acts as a link between the process running on the attacker's machine and the system services on a remote host simulated by \insight. 
This layer is responsible for forwarding each syscall argument and generating a proper request that the agent can understand. It is also responsible for sending this request to the agent and sending back the results to the calling process.

The syscall server layer (i.e. the agent that runs on the simulated system)
receives requests from the syscall client to execute specific syscalls using the OS services. After the syscall finishes, its results are marshalled and sent back to 
the client.

\subsection{Probabilistic Exploits}

In the simulator security model, a vulnerability 
is a mechanism used to access an otherwise restricted communication channel. 
In this model, a real exploit payload is replaced by an ID or ``magic string'' 
which is sent to a simulated application. If this application 
is defined to be vulnerable (and some other requirements are fulfilled), then an agent 
will be installed in the computer hosting the vulnerable application.

The simulated exploit payload includes the aforementioned magic string.  
When the \component{Simulator} subsystem receives this information, it looks up for 
the string in the \component{Exploits Database}. If it is found, then the simulator will decide 
if the exploit worked or not and with what effect based on a probability distribution that depends on 
the effective scenario information of that computer and the specific exploit. Suppose, 
for example, that the \component{Penetration Testing Framework} assumes (wrongly) the attacked machine 
is a Red Hat Linux 8.0, but that machine is indeed a Windows system. In this hypothetical situation, 
the exploit would fail with 100\% of probability. On the other side, if the attacked 
machine is effectively running an affected version of Red Hat Linux 9.0, then the probability of success could be 75\%,
or as determined in the exploit database.

\subsection{Remote Attack Model Overview}

\begin{figure}[t]
\begin{center}
\includegraphics[width=13cm]{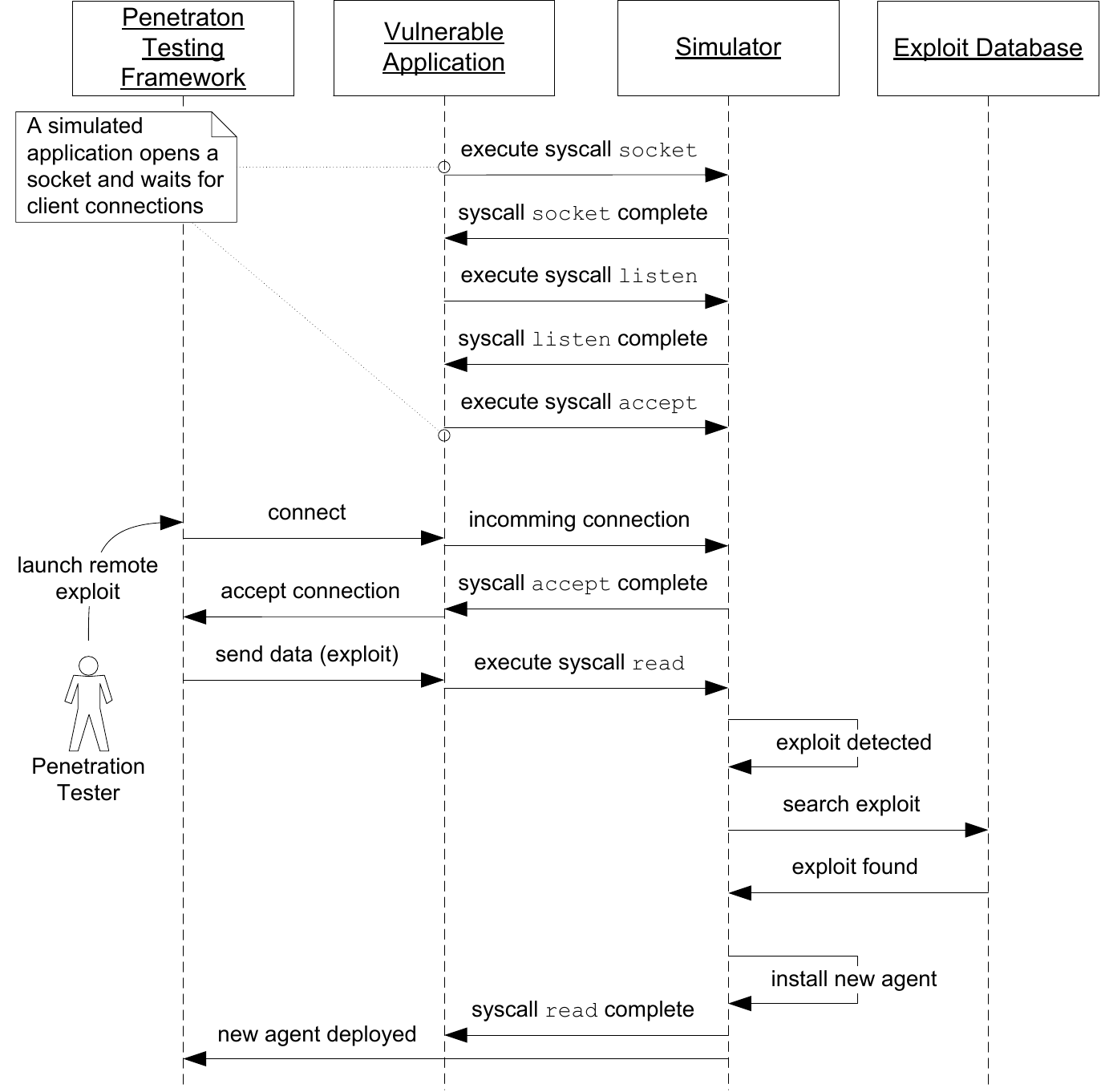}
\end{center}
\caption{Remote attack model.}
\label{fig:remote_attack_model}
\end{figure}

In \reffig{fig:remote_attack_model} we can see the sequence of events which occurs when an attacker 
launches a remote exploit against a simulated machine. The rectangles 
in the top are the four principal components involved: The \component{Penetration Testing Framework}, 
the \component{Simulator} and the \component{Exploits Database} are the subsystems explained in \reffig{fig:arch}; 
the \component{Vulnerable Application} is a simulated application or service which is 
running inside an \insight scenario and has an open port. 
In the diagram the declared components are represented as named rectangles, 
messages are represented as solid-line arrows, and time is represented as a vertical 
progression.

When an exploit is launched against a service running in a simulated machine, 
a connection is established between the \component{Penetration Testing Framework} 
and the service\footnote{This connection is established, for example, by a real Windows socket 
or a simulated TCP/IP socket, see \S\ref{st:sockets}.}. 
Then, the simulated exploit payload is sent to the application. 
The targeted application reads the payload by running the system call {\sf read}.
Every time the syscall {\sf read} is invoked, the \component{Simulator} subsystem analyzes if a 
magic string is present in the data which has just been read. When a magic string 
is detected, the \component{Simulator} searches for it in the \component{Exploits Database}. 
If the exploit is found, a new agent is installed in the compromised machine. 

The exploit payload also includes information of the OS that the 
\component{Penetration Testing Framework} knows about the attacked machine: OS version, system architecture, service packs, 
etcetera. All this information is used to compute the probabilistic function and allows the 
\component{Simulator} to decide whether the exploit should succeed or not.

\subsection{Local Attack Model Overview}

\insight can also simulate local attacks: If an attacker gains control over a machine 
but does not have enough privileges to complete a specific action, a local attack 
can deploy a new agent with higher privileges.  

\begin{figure}[t]
\begin{center}
\includegraphics[width=13cm]{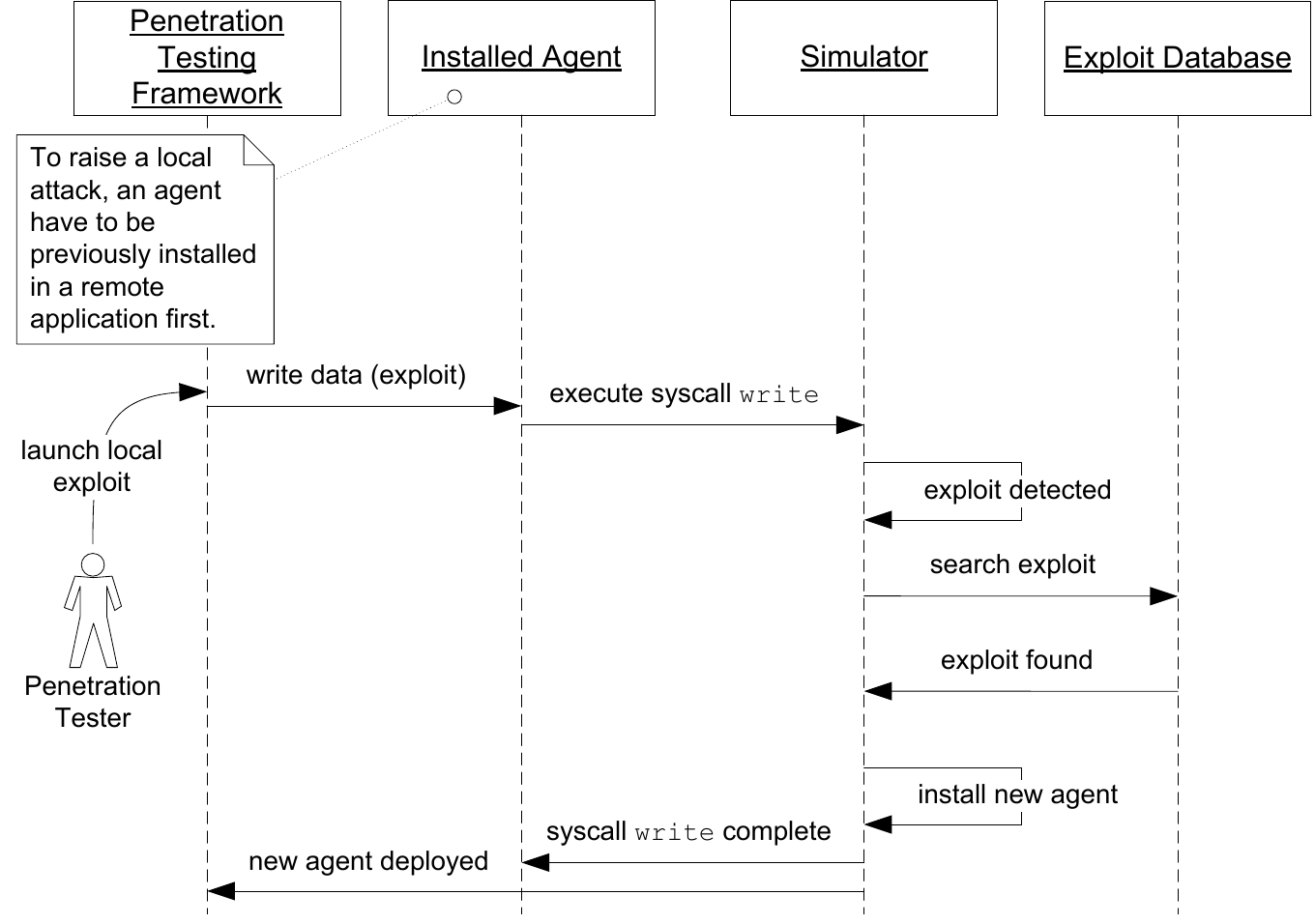}
\end{center}
\caption{Local attack model.}
\label{fig:local_attack_model}
\end{figure}

In \reffig{fig:local_attack_model} we can see the sequence of events which occurs when a local attack is launched against a given  machine. A running agent has to be present in the targeted machine in order to launch a local exploit. All local simulated attacks are executed by the \component{Simulator} subsystem identically: The \component{Penetration Testing Framework} will write the exploit magic string into the agent standard input, using the {\sf write} system call, and the \component{Simulator} will eventually detect the magic string intercepting that system call.

In a similar way as the previous example, the exploit magic string is searched in the database and a new agent (with higher privileges) is installed with probabilistic chance.

\section{Detailed Description}

One of the most challenging issues in the \insight architecture is to resolve the tension between realism and performance. The goal was to have a simulator on a single desktop computer, running hundreds of simulated machines, with a simulated traffic realistic from a penetration test point of view. But there is a trade-off between realism and performance and we are going to discuss some of these problems and other architecture details in the following sections.

\subsection{The Insight Development Library}

New applications can be developed for the simulation platform using a minimal 
\emph{C standard library}, a standardized collection 
of header files and library routines used to implement common operations such as: 
input, output and string handling in the C programming language.
 
This library---a partial \libc---implements the most common functions (e.g., read, write, open), allowing 
any developer to implement his own services with the usual compilers and 
development tools (e.g., gcc, g++, MS Visual Studio). For example, a web server could be implemented, linked with the 
provided \libc and plugged within the \insight simulated scenarios.

The provided \libc supports the most common system calls, but it is still incomplete 
and we were unable to compile complex open source applications. In spite of this, 
some services (e.g., a small DNS) and network tools (e.g., ipconfig, netstat) 
have been included in the simulation platform, and new system calls are planned to be 
supported in the future.

\subsection{Simulating Sockets}
\label{st:sockets}

\begin{figure}[ht]
\begin{center}
\includegraphics[width=10cm]{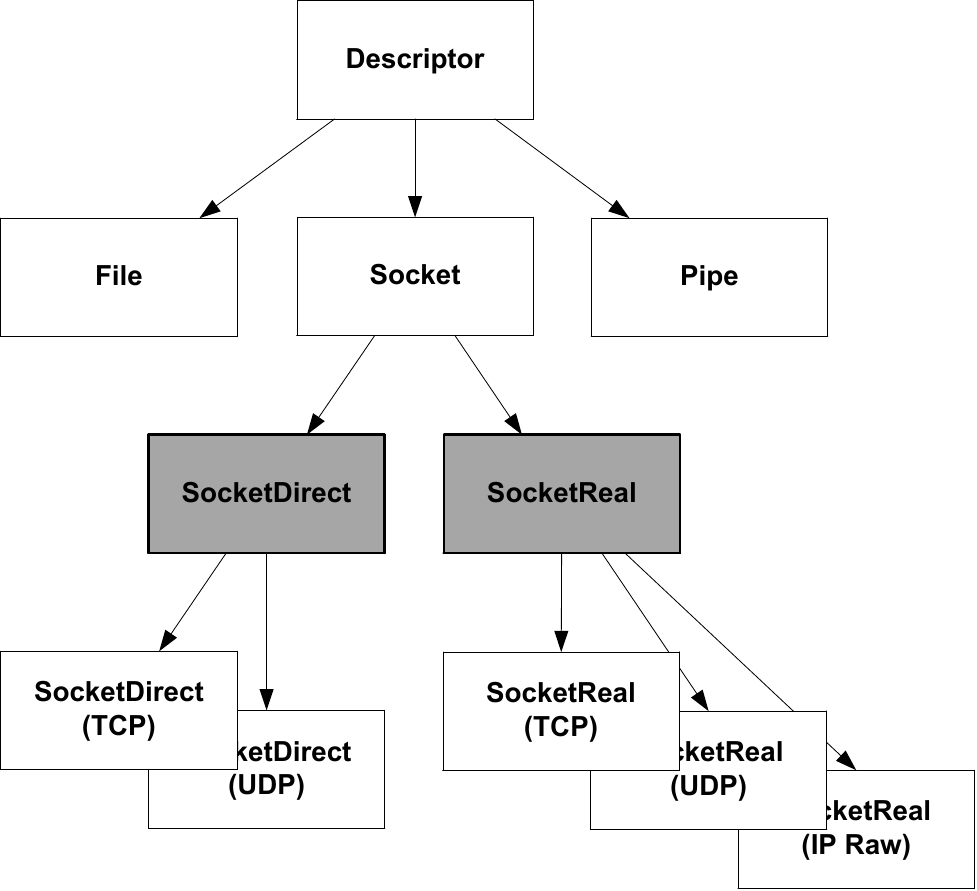}
\end{center}
\caption{Descriptors' object hierarchy tree.}
\label{fig:desc}
\end{figure}

A hierarchy for file descriptors has been developed as shown in \reffig{fig:desc}. 
File descriptors can refer (but they are not limited) to files, directories, sockets, or pipes. 
At the top of the hierarchy, the tree root shows the descriptor object which typically provides 
the operations for reading and writing data, closing and duplicating file descriptors, 
among other generic system calls.

The simulated sockets implementation spans in two kinds of 
supported sockets subclasses:
\begin{enumerate}
	\item SocketDirect. This variety of sockets is optimized for the simulation on one computer. 
	Socket direct is fast: as soon as a connection is established, the client keeps a file descriptor
	pointing directly to the server's descriptor. Routing is only executed during the connection and
	the protocol control blocks (PCBs) are created as expected, but they are only used during connection establishment.
	Reading and writing operations between direct sockets are carried out using shared memory. 
	Since both sockets can access the shared memory area like regular working memory, this is a very 
	fast way of communication.
	\item SocketReal. In some particular cases, we are interested in having full socket functionality. 
	For example, the communication between \insight and the outside world is made using real sockets. As a result, 
	this socket subclass wraps a real BSD socket of the underlying OS.
\end{enumerate}

Support for routing and state-less firewalling was also implemented, 
supporting the simulating of attack payloads that connect back to the attacker, 
accept connections from the attacker or reuse the attack connection.

\subsection{The Exploits Database}

When an exploit is raised, \insight has to decide whether the attack is successful 
or not \emph{depending on the environment conditions}. For example, an exploit can require 
either a specific service pack installed in the target machine to be successful, 
or a specific library loaded in memory, or a particular open port, among others requirements. 
All these conditions 
vary over the time, and they are basically unpredictable from the attacker's standpoint. 
As a result, the behavior of a given exploit has been modeled using a probabilistic approach.

In order to determine the resulting behavior of the attack, \insight uses the \component{Exploits Database} showed 
in the architecture layout of \reffig{fig:arch}. It has a XML tree structure. 
For example, if an exploit succeeds against a clean XP professional SP2 with 83\% probability, or crashes the machine with 0.05\% probability in other case; this could be expressed as follows:

\small
\begin{verbatim}
<database>
  <exploit id="sample exploit">
    <requirement type="system">
      <os arch="i386" name="windows" />
      <win>XP</win>
      <edition>professional</edition>
      <servicepack>2</servicepack>
    </requirement>
    <results>
      <agent chance="0.83" />   
      <crash chance="0.05" what="os" /> 
      <reset chance="0.00" what="os" /> 
      <crash chance="0.00" what="application" /> 
      <reset chance="0.00" what="application" /> 
    </results>
  </exploit>
  <exploit> ... </exploit>
  <exploit> ... </exploit>
  ...
</database>
\end{verbatim}
\normalsize

The conditions needed to install a new agent are described in the \xmltag{requirements} section. It is possible to use several tags in this section, they specify the conditions which have influence on the execution of the exploit (e.g., OS required, a specific application running, an open port). The \xmltag{results} section is a list of the relevant probabilities. 
In order, these are the chance of: 
\begin{enumerate}
	\item successfully installing an agent, 
	\item crashing the target machine, 
	\item resetting the target machine, 
	\item crashing the target application, 
	\item and the chance of resetting the target application.
\end{enumerate}

To determine the result, we follow this procedure: processing the lines in order, for each positive probability, 
choose a random value between 0 and 1. If the value is smaller than the chance attribute, 
the corresponding action is the result of the exploit.

In this example, we draw a random number to see if an agent is installed. 
If the value is smaller than 0.83, an agent is installed and the execution of the exploit is finished. 
Otherwise, we draw a second number to see if the OS crashes. 
If the value is smaller than 0.05, the OS crashes and the attacked machine becomes useless, 
otherwise there is no visible result.
Other possible results could be: raising an IDS alarm, writing some log in a network device (e.g. firewall, IDS or router) or capturing a session id, cookie, credential or password. 

The exploits database allows us to model the probabilistic behavior 
of any exploit from the attacker's point of view, but how do we populate our database? 
A paranoid approach would be to assign a probability of success of 100\% to every exploit. 
In that way, we would consider the case where an attacker 
can launch each exploit as many times as he wants, and will finally 
compromise the target machine with 100\% probability 
(assuming the attack does not crash the system).

A more realistic approach is to use statistics from real networks.
Currently we are using the framework presented by Picorelli \cite{Pico06} 
in order to populate the probabilities in the exploits database. This framework was 
originally implemented to assess and improve the quality of real exploits in QA environments. 
It allows us to perform over 500 real exploitation tests daily on several running configurations, 
spanning different target operating systems with their own setups and applications that 
add up to more than 160 OS configurations. In this context, a given exploit is executed against:
\begin{itemize}
	\item All the available platforms
	\item All the available applications
\end{itemize}

All these tests are executed automatically using low end hardware, VMware servers, OS images and snapshots. 
The testing framework has been designed to improve testing time and coverage, and we have modified it 
in order to collect statistical information of the exploitation test results.

\subsection{Scheduler}

The scheduler's main task is to assign the CPU resources to the different 
simulated actors (e.g. simulated machines and process).
The scheduling iterates over the hierarchy machine-process-thread as a tree 
(like a depth-first search), each machine running its processes in round-robin.

In a similar way, running a process is giving all its threads the order to run until a system call 
is needed. Obviously, depending on the state of each thread, they run, 
change state or finish execution. The central issue is that threads execute systems calls and 
then (if possible) continue their activity until they finish or another system call is required. 

\insight threads are simulated within real threads of the underlying OS. 
Simulated machines and processes are all running within one or several working processes (running hundreds of threads), 
and all of them are coordinated by a unique scheduler process called the \emph{master process}.
Thanks to this architecture, there is a very low loss of performance due to context switching\footnote{Because 
descriptors and pointers remain valid when switching from one machine to the other.}.

\subsection{File System}
\label{filesystem}

In order to handle thousands of files without wasting a huge amount of disk space, 
the file system simulation is accomplished by mounting shared file repositories. 
We are going to refer to these repositories as \emph{template file systems}. 
For example, all simulated Windows XP systems could share a file repository with 
the default installation provided by Microsoft. These shared templates would have 
reading permission only. Thus, if a virtual machine needs to read or change a file, it 
will be copied within the local file system of the given machine.  

This technique is known as \emph{copy-on-write}. The fundamental idea is to allow multiple 
callers to ask for resources which are initially indistinguishable, giving them pointers 
to the same resource. This function can be maintained until a caller tries to modify its copy 
of the resource, at which point a true private copy is created to prevent the changes from 
becoming visible to everyone else. All of this happens transparently to the callers. 
The primary advantage is that no private copy needs to be created if a caller never makes any modification.

Additionaly, with the purpose of improving the simulator's performance, a file cache 
is implemented: 
the simulator saves the most recent accessed files (or block of files) in memory. In high scale 
simulated scenarios, it is very common to have several machines doing the same task at (almost) 
the same time\footnote{For example, when the simulation starts up, all UNIX machines would read 
the boot script from {\sf /etc/initd} file.}. If the data requested by these kind of tasks are 
in the file system cache, the whole system performance will improve, because less disk accesses
would be required, even in scenarios of hundreds or thousands of simulated machines.

\section{Performance Analysis}

To evaluate the performance of the simulator we run a test including a scenario with
an increasing number of complete LANs with 250 computers each, simultaneously emulated. The tests only involve the
execution of a network discovery on the complete LANs through a TCP connection to port 80. 
An original pen-testing module used for information was executed with no modifications, this was 
a design goal of the simulator, to use real unmodified attack modules when possible.

\begin{table}[ht]
\small
\begin{center}
\begin{tabular}{c c c c}
\multicolumn{4}{c}{{\bf Performance of the simulator }} \\
\hline
\hline
LANs & Computers & Time (secs) & Syscalls/sec \\
\hline
1 & 250 & 80 & 356 \\
2 & 500 & 173 & 236 \\
3 & 750 & 305 & 175 \\
4 & 1000 & 479 & 139 \\
\end{tabular}
\end{center}
\caption[Evolution of the system performance as the simulated scenario grows, running a network discovery module, connecting to a predefined port.]
{Evolution of the system performance as the simulated scenario grows, running a network discovery module, connecting to a predefined port. This benchmark was run on a single Intel Pentium D 2.67Ghz, 1.43GB RAM.}
\label{tab:performance}
\end{table}

\normalsize
We can observe the decrease of system calls processed per second as we increase the number of simulated computers as \insight was run on a single real computer with limited resources. Nevertheless, the simulation is efficient because system calls are required on demand by the connections of the module gathering the information of the networks through TCP connections.

\section{Applications} 

We have created a playground to experiment with cyber-attack scenarios which has
several applications. The most important are:

\begin{description}
\item [Attack Planning.] This is the main application. In Chapter \ref{chap:deterministic}
we used this simulator to test 
attack planning algorithms in a variety of scenarios.

\item[Data collection and visualization.] Having the complete network scenario 
in one computer allows an easy capture and log of system calls and network traffic. 
This information is useful for analyzing and debugging real pen-test tools and their 
behavior in complex scenarios. Some efforts have been made to visualize attack 
pivoting and network information gathering using the platform presented.

\item [Pentest training.]

Our simulation tool is already being used in Pentest courses.
It provides reproducible scenarios, where students can practice the different steps of a pentest: information gathering, attack and penetrate, privilege escalation, local information gathering and pivoting.

The simulation allows the student to grasp the essence of pivoting.
Setting up a real laboratory where pivoting makes sense is an expensive task,
whereas our tool requires only one computer per student
(and in the case of a network / computer crash, the simulation environment can be easily reset).
Configuring new scenarios with more machines or more complex topologies is easy,
as a scenario wizard is provided.

In Pentest classes with \insight, the teacher can check the logs to see if students used the right tools with the correct parameters.
He can test the students' ability to plan and see if they avoided performing unnecessary actions.
The teacher can also identify their weaknesses as pentesters and plan new exercises to work on these.
The students can be evaluated: success, performance, stealth and quality of reports can be measured.

\item [Worm Spreading Analysis.] The lightweight design of the platform allows 
the simulation of socket/network behavior of thousands of computers, providing a good 
framework for research on worm infestation and spreading. It should be possible to develop 
very accurate applications to mimic worm behavior using the \insight \texttt{C} 
programming API. There are available abstract modeling  \cite{chen:infocom2003} or high-fidelity 
discrete event \cite{songjie2005} studies
but no system call level recreation of attacks 
like the one we propose as future application of this platform.

\item [Analysis of countermeasures.] Duplication of the production configuration on 
a simulated staging environment accurately mimicking or mirroring the security aspects 
of an organization's network allows the anticipation of software/hardware changes and 
their impact on security. For example, you can answer questions like ``Will the network 
avoid attack vector $A$ if firewall rule $R$ is added to the complex rule set $S$  
of firewall $F$?''

\item [Impact of 0-day vulnerabilities.]

The simulator can be used to study the impact of 0-days (vulnerabilities that
have not been publicly disclosed) in your network. 
How is that possible, if we do not know current 0-days?
But we can model the existence of 0-day vulnerabilities based on statistics.
In our security model, the specific details of the vulnerability 
are not needed to study the impact on the network, just that it may exist with 
a measurable probability.

\begin{figure}[ht]
\begin{center}
\includegraphics[width=12cm]
{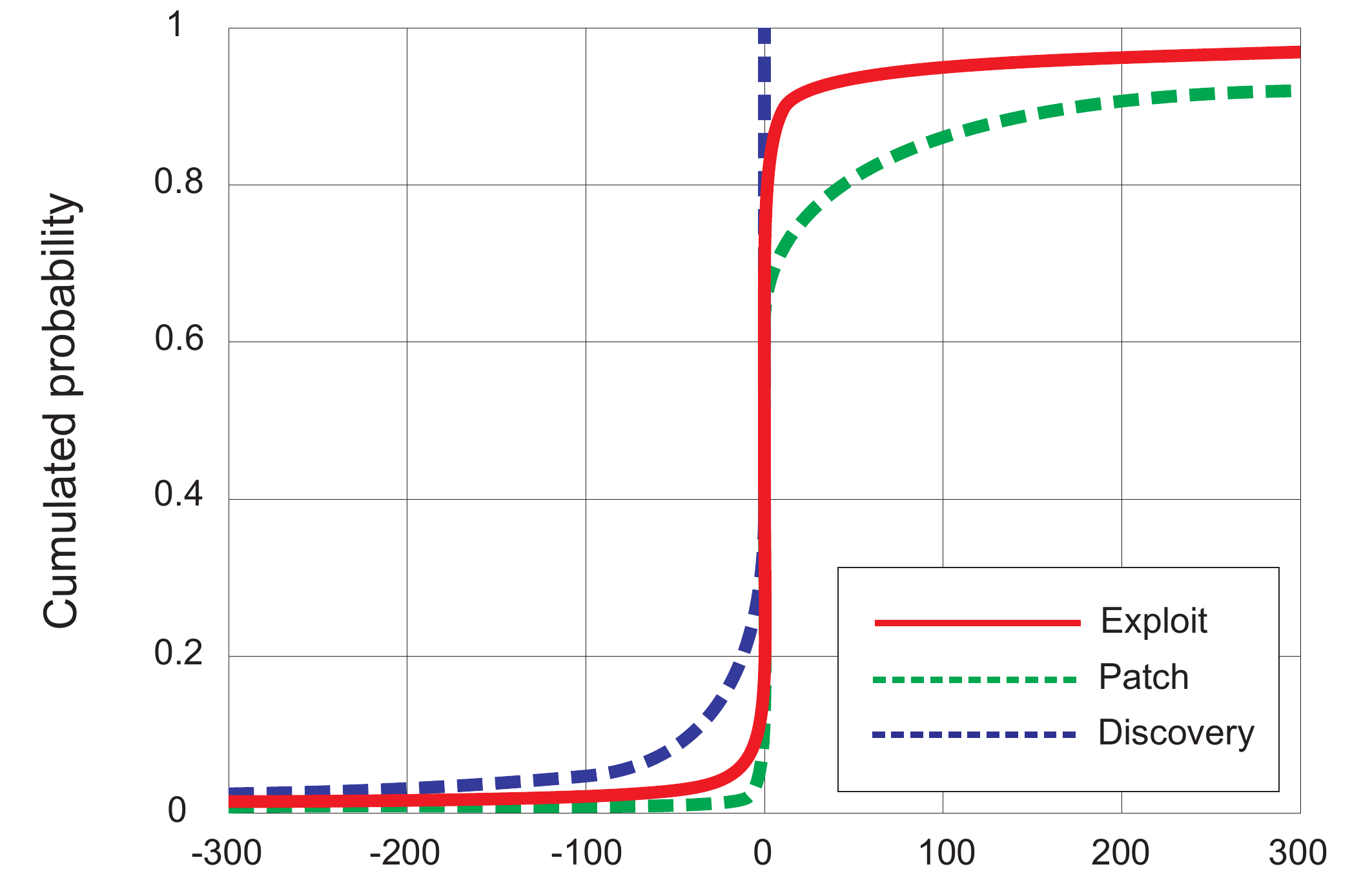}
\end{center}
\caption{Probabilities before disclosure.}
\label{fig:prob_disclosure}
\end{figure}

That information can be gathered from public vulnerability databases:
the discovery date, exploit date, disclosure date and patch date
are found in several public databases of vulnerabilities and exploits
\cite{cert, securityfocus, secunia, frsirt}.

The risk of a 0-day vulnerability is given by the probability of an attacker discovering and exploiting it. Although we do not have data about the security underground,
the probabilities given by public information are a lower bound indicator.

As shown in \cite{frei06}, the risk posed by a vulnerability 
exists before the discovery date,
augments as an exploit is made available for the vulnerability,
and when the vulnerability is disclosed.
The risk only diminishes as a patch becomes available 
and users apply the patches (and workarounds).

The probability of discovery, and the probability of an exploit being developed, 
can be estimated as a function of the time before disclosure
(see \reffig{fig:prob_disclosure} taken from \cite{frei06}).
For Microsoft products, we have visibility of upcoming disclosures of vulnerabilities:
every month (on patch Tuesday) on average $9.40$ patches are released (high and medium risk).
Based on those dates we estimate the probability that the vulnerabilities 
were discovered and exploited during the months before disclosure.

\end{description}

\section{Summary}

We have created a playground to experiment with cyber-attack scenarios.
The framework is based on a probabilistic attack model---that model is 
also used by attack planning tools developed in our lab.
By making use of the proxy syscalls technology, and simulating multiplatform agents,
we were able to implement a simulation that is both realistic and 
lightweight, allowing the simulation of networks with thousands of hosts.

The framework provides a global view of the scenarios.
It is centered on the attacker's point of view,
and designed to increase the size and complexity of simulated scenarios,
while remaining realistic for the attacker.

The value of this framework is given by its multiple applications:
\begin{itemize}
	\item Systematic study of \emph{Attack Planning} techniques.
	\item Evaluation of network security.
	\item Evaluation of security countermeasures.
	\item Anticipating the risk posed by 0-day vulnerabilities.
	\item Pentest training.
	\item Worm spreading analysis.
	\item Data generation to test visualization techniques.
\end{itemize}

\part{Development of a Probabilistic Attack Planner}


\chapter{Probabilistic Attack Planning} \label{chap:probabilistic}

We have presented in Chapter~\ref{chap:deterministic} an approach to the \textit{attack planning} problem 
based on modeling the actions and assets in the PDDL language, and using off-the-shelf AI tools to generate attack plans. This approach however is limited. 
In particular, the planning is classical (the actions are deterministic) and thus not able to handle the uncertainty involved in this form of attack planning. 

In this chapter we contribute a planning model that does capture the uncertainty about the results of the actions, which is modeled as a probability of success of each action. We present efficient planning algorithms, specifically designed for this problem, that achieve industrial-scale runtime performance
(able to solve scenarios with several hundred hosts and exploits). 
These algorithms take into account the probability of success of the actions and their expected cost (for example in terms of execution time, or network traffic generated).
We thus show that probabilistic attack planning can be solved efficiently
for the scenarios that arise when assessing the security of large networks
(under certain simplifying assumptions that will be discussed during the chapter).

Two ``primitives'' are presented, which are used as building blocks
in a framework separating the overall problem into two levels of abstraction.
We also present the experimental results obtained with our implementation,
and conclude with some ideas for further work.

\section{Introduction}

Penetration testing is one of the most trusted ways of assessing the
security of networks large and small. The result of a penetration test is a
repeatable set of steps that result in the compromise of particular assets in the network.
As we have discussed in Chapter~\ref{chap:penetration_testing},
penetration testing frameworks have been developed to
facilitate the work of penetration testers 
and make the assessment of network security more accessible to non-expert users.

In Section \ref{sec:need-for-automation} we discussed how 
the evolution of pentesting tools -- covering new
attack vectors, and shipping increasing numbers of exploits and information gathering techniques --
created the need to automate the control of the pentest framework.
This is what we call the {\em attack planning} problem.
This problem was introduced to the AI planning community by Boddy \emph{et al.} 
as the ``Cyber Security'' domain \cite{BodGohHaiHar05}.
In the pentesting industry, Lucangeli \emph{et al.} proposed a solution based 
on modeling the actions and assets in the PDDL language,\footnote{PDDL stands for 
Planning Domain Definition Language. 
Refer to \cite{FoxLon03} for a specification of PDDL 2.1.}
and using off-the-shelf 
planners to generate attack plans \cite{LucSarRic10}. 
Using PDDL to work with attack graphs has also been explored in \cite{GhoGho09}.
Herein we are concerned with the specific context of regular automated pentesting,
as in ``Core Insight Enterprise'' tool, and use the term ``attack planning'' in that sense.

Recently, a model based on partially observable Markov decision processes (POMDP)
was proposed \cite{SarBufHof11} (this is the subject of Chapter~\ref{chap:pomdp}).
This grounded the attack planning problem in a well-researched formalism,
and provided a precise representation of the attacker's uncertainty
with respect to the target network. In particular, the information gathering phase
was modeled as an integral part of the planning problem.
However, as the authors show, this solution does not scale to medium or large real-life networks.

In this chapter, we take a different direction: 
the uncertainty about the results of the actions is modeled 
as a {\em probability of success} of each action, 
whereas in \cite{SarBufHof11} the uncertainty is modeled
as a distribution of probabilities over the states.
This allows us to produce an efficient planning algorithm,
specifically designed for this problem,
that achieves industrial-scale runtime performance.

Of course planning in the probabilistic setting is far more difficult 
than in the deterministic one.
We do not propose a general algorithm, but a solution suited
for the scenarios that need to be solved in a real world penetration test.
In particular, we make simplifying assumptions about the actions and
the machines, namely supposing that the actions and machines are independent
from each other. In the trade-off between realism of the model
and scalability of the resulting planning problem, we chose here to prioritize scalability.
The computational complexity of our planning solution is $\calO ( n \log n) $,
where $n$ is the total number of actions in the case of an attack tree
(with fixed source and target hosts),
and $\calO ( M^{2} \cdot n \log n) $ where $M$ is the number of machines
in the case of a network scenario.
With our implementation, we were able to solve planning
in scenarios with up to 1000 hosts distributed in different networks.

We start with a brief review of the attack model in Section \ref{sec:model},
then continue with a presentation of two ``primitives'' in 
Sections \ref{sec:choose} and \ref{sec:combine}.
These primitives are applied in more general settings in 
Sections \ref{sec:dynamic} and \ref{sec:distinguished-assets}.
Section \ref{sec:experiments} shows experimental results from the implementation
of these algorithms.
We conclude with some ideas for future work.

\section{The Attack Model}\label{sec:model}

We summarize below the basic background from the 
conceptual model of computer attacks presented in Chapter~\ref{chap:model}
(for more details refer to \cite{ArcRic03,FutNotRic03,Richarte03,RusTis07}),
and further develop some aspects of the model like the actions' costs. 
This model is based on the concepts of assets, goals, agents and actions.
In this description, an attack involves a set of agents, 
executing sequences of actions, 
obtaining assets (which can be information
or actual modifications of the real network and systems) in order to reach a set of goals.

An \textit{asset} can represent anything that an attacker may need to
obtain during the course of an attack, including the actual goal. 
Examples of assets:
information about the Operating System (OS) of a host $H$;
TCP connectivity with host $H$ on port $P$;
an Agent installed on a given host $H$.
To \emph{install an agent} means to break into a host,
take control of its resources, 
and eventually use it as pivoting stone to continue the attack
by launching new actions based from that host.

The \textit{actions} are the basic steps which form an attack.
Actions have requirements (also called preconditions)
and a result: the asset that will be obtained if the action
is successful.
For example, consider the exploit 
\Name{IBM Tivoli Storage Manager Client Remote Buffer 
Overflow}\footnote{The particular implementations that we have studied
are the exploit modules for Core Impact and Core Insight Enterprise, although the same model
can be applied to other implementations, such as Metasploit.}
for the vulnerabilities in \emph{dsmagent} described by CVE-2008-4828.
The result of this action is to install an agent,
and it requires that the OS of the target host is 
Windows 2000, Windows XP, Solaris 10, Windows 2003, or AIX 5.3.
In this model, all the exploits (local, remote, client-side, webapps) are represented
as actions. Other examples of actions are:
TCP Network Discovery, UDP Port Scan, DCERPC OS Detection,
TCP Connectivity Probe.

The major differences between the attack model used in this work
and the {\em attack graphs} used in 
\cite{AmmWijKau02,JajNoeBer05,JhaSheWin02,PhiSwi98,RitAmm00,SheHaiJha02}
are twofold: 
to improve the realism of the model, we consider that
the actions can produce numerical effects (for example, the expected running time of each action);
and that the actions have a probability of success (which models the uncertainty about the
results of the action).

\subsection{Deterministic Actions with Numerical Effects}

In the deterministic case, the actions and assets that compose a specific planning
problem can be successfully represented in the PDDL language.
This idea was proposed in \cite{SarWei08} and further analyzed in \cite{LucSarRic10},
and treated in Chapter~\ref{chap:deterministic} of this thesis.
The assets are represented as PDDL predicates, and the actions
are translated as PDDL operators.
The authors show how this PDDL representation allowed
them to integrate a penetration testing tool with an external planner,
and to generate attack plans in realistic scenarios.
The planners used -- Metric-FF \cite{Hoffmann02} and SGPlan \cite{CheWahHsu06} --
are state-of-the-art planners able to handle numerical effects.

Fig.~\ref{fig:pddl-action} shows an example of a PDDL action: an exploit for the 
IBM Tivoli vulnerability, that will attempt
to install an agent on target host $t$ from an agent previously installed
on the source host $s$. To be successful, this exploit requires that
the target runs a specific OS, has the service {\ttfamily mil-2045-47001} running and
listening on port 1581.

\begin{figure}[ht]
{
\footnotesize

\begin{verbatim} 
(:action IBM_Tivoli_Storage_Manager_Client_Exploit
:parameters (?s - host ?t - host)
:precondition (and
  (compromised ?s)
  (and (has_OS ?t Windows)
    (has_OS_edition ?t Professional)
    (has_OS_servicepack ?t Sp2)
    (has_OS_version ?t WinXp)
    (has_architecture ?t I386))
  (has_service ?t mil-2045-47001)
  (TCP_connectivity ?s ?t port1581)
)
:effect(and 
  (installed_agent ?t high_privileges)
  (increase (time) 4)
))
\end{verbatim} 
}
\vmenos
\caption{Exploit represented as PDDL action.}
\label{fig:pddl-action}
\end{figure}

The average running times of the exploits are measured by 
executing all the exploits of the penetration testing tool in a testing lab.
More specifically, in Core's testing lab there are more than 
748 virtual machines with different OS and installed applications,
where all the exploits of Core Impact are executed every night \cite{Pico06}.

\subsection{Actions' Costs} \label{sec:actions-costs}

The execution of an action has a multi-dimensional cost. We detail below
some values that can be measured (and optimized in an attack):

\begin{description}
	\item[Execution time:] Average running time of the action.

	\item[Network traffic:] The amount of traffic sent over the network
	increases the level of noise produced.

	\item[IDS detection:] Logs lines generated and alerts triggered by the execution of the 
	action increase the noise produced.

	\item[Host resources:] The execution of actions will consume resources of 
	both the local and remote host, in terms of CPU, RAM, hard disc usage, etc.

  \item[Traceability of the attack:] Depends on the number of intermediate hops and topological factors.

	\item[Zero-day exploits:] Exploits for vulnerabilities that are not
	publicly known are a valuable resource, that should be used only when 
	other exploits have failed (the attacker usually wants to minimize the use 
	of ``0-days'').
\end{description}

In our experiments, we have chosen to optimize the expected execution time.
In the context of regular penetration tests, 
minimizing the expectation of total execution time is a way of
maximizing the amount of exploits successfully launched in a fixed time frame
(pentests are normally executed in a bounded time period).

However, the same techniques can be applied to any other scalar cost,
for example to minimize the noise produced by the actions
(and the probability of being detected).
The issue of measuring the costs of the exploits, and more generally of 
evaluating \emph{Exploit Quality Metrics}, was more recently discussed
in \cite{Sarraute-8dot8}.

\subsection{Probabilistic Actions}

Another way to add realism to the attack model is to consider that the actions
are nondeterministic. This can be modeled by associating probabilities 
to the outcomes of the actions. In the case of an exploit,
the execution of the exploit can be successful (in that case the attacker
takes control of the target machine) or a failure. This is represented
by associating a \emph{probability of success} to each exploit.

The probability of success is conditional: it depends on the environment conditions.
For example, the IBM Tivoli exploit for CVE-2008-4828 is
more reliable (has a higher probability of success) 
if the OS is Solaris since it has no heap protection,
the stack is not randomized and is executable by default.
Alternatively, the exploit is less reliable (has a lower probability of success)
if the OS is Windows XP SP2 or Windows 2003 SP1, 
with Data Execution Prevention (DEP) enabled.
On Windows Vista, the addition of 
Address Space Layout Randomization (ASLR)
makes the development of an exploit even more difficult, and diminishes its 
probability of success.
In practice, the probability of success of each exploit is measured
by exhaustively executing the exploit against a series of targets,
covering a wide range of OS and application versions.

Although it improves the realism of the model, considering probabilistic
actions also makes the planning problem more difficult.
Using general purpose probabilistic planners did not work
as in the deterministic case;
for instance, we experimented with Probabilistic-FF \cite{DomHof07}
with poor results, since it was able to find plans in only very small cases.

In the rest of this chapter, we will study algorithms
to find optimal attack paths in scenarios of increasing difficulty.
We first describe two primitives, and then apply them 
in the context of regular automated pentesting.
In these scenarios we make an additional hypothesis:
the independence of the actions. 
Relaxing this hypothesis is a subject for future work.

\section{The \emph{Choose} Primitive}\label{sec:choose}

We begin with the following basic problem.
Suppose that the attacker (i.e. pentester) wants to gain access to the credit cards 
stored in a database server $H$ by installing a system agent.
The attacker has a set of $n$ remote exploits that he can launch against that server.
These exploits result in the installation of a system agent when successful 
(see Fig.~\ref{fig:grafo1}).

\Figure{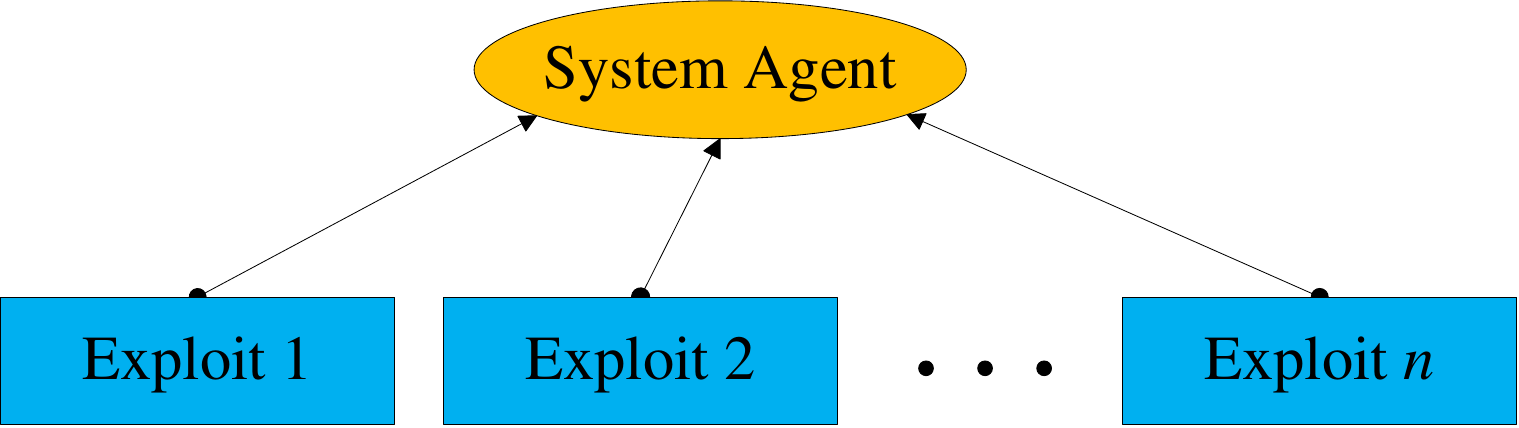}{!tbh}{ 0.8 \linewidth}{grafo1}
{Multiple exploits may install a System agent (on the target host).}

In this scenario, the attacker has already performed information gathering
about the server $H$, collecting a list of open/closed ports,
and running an OS detection action such as Nmap.
The pentesting tool used provides statistics on
the probability of success and expected running time for each exploit 
in the given conditions.\footnote{In our experiments
we used the database of tests of
Core Impact and Core Insight Enterprise.}
The attacker wants to minimize the expected execution time of the whole attack.
A more general formulation follows:

\begin{problem}\label{action_order}
Let $\frakg$ be a fixed goal, and let $\{ A_1, \ldots, A_n  \}$ be a set 
of $n$ independent actions whose result is $\frakg$.
Each action $A_k$ has a probability of success $p_k$ and expected
cost $t_k$. 
Actions are executed until an action is successful and provides the goal $\frakg$
(or all the actions fail).

\noindent
Task: Find the order in which the actions must be executed
in order to minimize the expected total cost.

\end{problem}

As already stated, we make the simplifying assumption that the probability of success of each action 
is independent from the others.
If the actions are executed in the order $A_1, \ldots, A_n$,
using the notation $ \overline{p_i} = 1 - p_i $, the expected cost
can be written as
\begin{align}\label{exp_time_or}
T_{\{ 1 \ldots n \} } = t_1 + \overline{p_1} \, t_2 + \ldots 
+ \overline{p_1} \, \overline{p_2} \ldots \overline{p_{n-1}} \, t_n .
\end{align}
The probability of success is given by
$$
P_{ \{ 1 \ldots n \} } = p_1 + \overline{p_1} \, p_2 + 
\overline{p_1} \, \overline{p_2} \, p_3 + \ldots +
\overline{p_1} \ldots \overline{p_{n-1}} \, p_n ,
$$
and the complement 
$\overline{ P_{ \{ 1 \ldots n \} }}  
= \overline{p_1} \, \overline{p_2} \ldots \overline{p_{n}} $.
In particular this shows that the total probability of success does not depend on the order of
execution.

Even though this problem is very basic, we didn't find references to its solution. 
This is why we give below some details on the solution that we found.

\begin{lemma} \label{lemma-ordered-actions}
Let $A_1, \ldots, A_n$ be actions such that
$t_1 / p_1 \menor t_2 / p_2 \menor \ldots \menor t_n / p_n $.
Then
$$
\frac{ T_{ \{1 \ldots n-1 \} } } { P_{ \{1 \ldots n-1 \} } } \menor \frac{t_n}{p_n} .
$$
\end{lemma}

\begin{proof}
We prove it by induction. The case with two actions is trivial,
since we know by hypothesis that $t_1 / p_1 \menor t_2 / p_2 $.
For the inductive step, suppose that the proposition holds for $n-1$ actions.
Consider the first three actions $A_1, A_2, A_3$.
The inequality 
\begin{align*}
\frac{T_{\{12\}}}{P_{\{12\}}} \menor \frac{t_3}{p_3}
\end{align*}
holds if and only if
$ t_2 / p_2  \menor t_3 / p_3 $.
So the first two actions can be considered as a single action $A_{12}$ with
expected cost (e.g. running time) $T_{\{12\}}$ and probability of success $P_{\{12\}}$.
We have reduced to the case of $n-1$ actions, and we can use the induction hypothesis
to conclude the proof. 
\end{proof}

\begin{proposition}\label{prop_action_order}
A solution to Problem \ref{action_order} is to sort the actions according to the 
coefficient $t_k / p_k$ (in increasing order), and to execute them in that order.
The complexity of finding an optimal plan is thus $\calO ( n \log n ) $.
\end{proposition}

\begin{proof}
We prove it by induction. We begin with the case of
two actions $A_i$ and $A_j$ such that $t_i / p_i \menor t_j / p_j$.
It follows easily that $- p_i t_j \menor - p_j t_i $ and that
$$
t_i + (1 - p_i) \, t_j \menor t_j + (1 - p_j) \, t_i .
$$
For the inductive step, suppose for the moment
that the actions are numbered so that
$t_1 / p_1 \menor \ldots \menor t_n / p_n $,
and that the proposition holds for all sets of $n-1$ actions.
We have to prove that executing $A_1$ first is better
that executing any other action $A_k$ for all $k \neq 1$.
We want to show that
$$
\begin{array}{c}
t_1 + \displaystyle\sum_{2 \leqslant i \leqslant n} t_i \cdot 
\prod_{1 \leqslant j \leqslant i-1} \overline{p_j}  \\
 \menor \: \:
t_k + \displaystyle\sum_{1 \leqslant i \leqslant n, \, i \neq k} t_i \cdot 
\overline{p_k} \cdot 
\prod_{1 \leqslant j \leqslant i-1, \, j \neq k} \overline{p_j} .
\end{array}
$$
Notice that in the two previous sums, the coefficients of $t_{k+1}, \ldots, t_n$
are equal in both expressions.
They can be simplified,
and using notations previously introduced, the inequality can be rewritten
$$
T_{ \{1 \ldots k-1 \} } + \overline{ P_{ \{1 \ldots k-1 \} } } \; t_k 
 \menor t_k + \overline{p_k} \; T_{ \{1 \ldots k-1 \} }
$$
which holds if and only if
$$
\frac{ T_{ \{1 \ldots k-1 \} } } { P_{ \{1 \ldots k-1 \} } }  \menor  \frac{t_k}{p_k}
$$
which is true by Lemma \ref{lemma-ordered-actions}.
We have reduced the problem to sorting the coefficients $t_k / p_k$.
The complexity is that of making the $n$ divisions $t_k / p_k$
and sorting the coefficients.
Thus it is $\calO ( n + n \log n ) = \calO ( n \log n )$.
\end{proof}

We call this the {\em choose} primitive because it tells you,
given a set of actions, which action to choose first: 
the one that has the smallest $t/p$ value.
In particular, it says that you should execute first the actions
with smaller cost (e.g. runtime) or higher probability of success,
and precisely which is the trade-off between these two dimensions.

The problem of choosing the order of execution within a set of exploits
is very common in practice. 
In spite of that, 
the automation methods currently implemented in penetration testing frameworks
offer an incomplete solution,\footnote{As of July 2011, Immunity Canvas \cite{Aitel04}
doesn't provide automated execution of exploits; Metasploit \cite{Moore10} has
a feature called {\em ``autopwn''} that launches all the exploits available 
for the target ports in arbitrary order; Core Impact Pro launches
first a set of {\em ``fast''} exploits and then {\em ``brute-force''} exploits \cite{SarWei08}, but
arbitrary order is used within each set; Core Insight Enterprise
uses planning techniques based on a PDDL description \cite{LucSarRic10} that takes into account
the execution time but not the probability of success of the exploits.}
over which the one proposed here constitutes an improvement.

\section{The \emph{Combine} Primitive} \label{sec:combine}

\subsection{Predefined Strategies}

We now consider the slightly more general problem where the goal $\frakg$ can be obtained
by predefined strategies.
We call \emph{strategy} a group of actions
that must be executed in a specific order.
The strategies are a way to incorporate the expert knowledge 
of the attacker in the planning system 
(cf. the opening moves in chess).
This idea has been used in the automation of pentesting tools, see \cite{SarWei08}.

\Figure{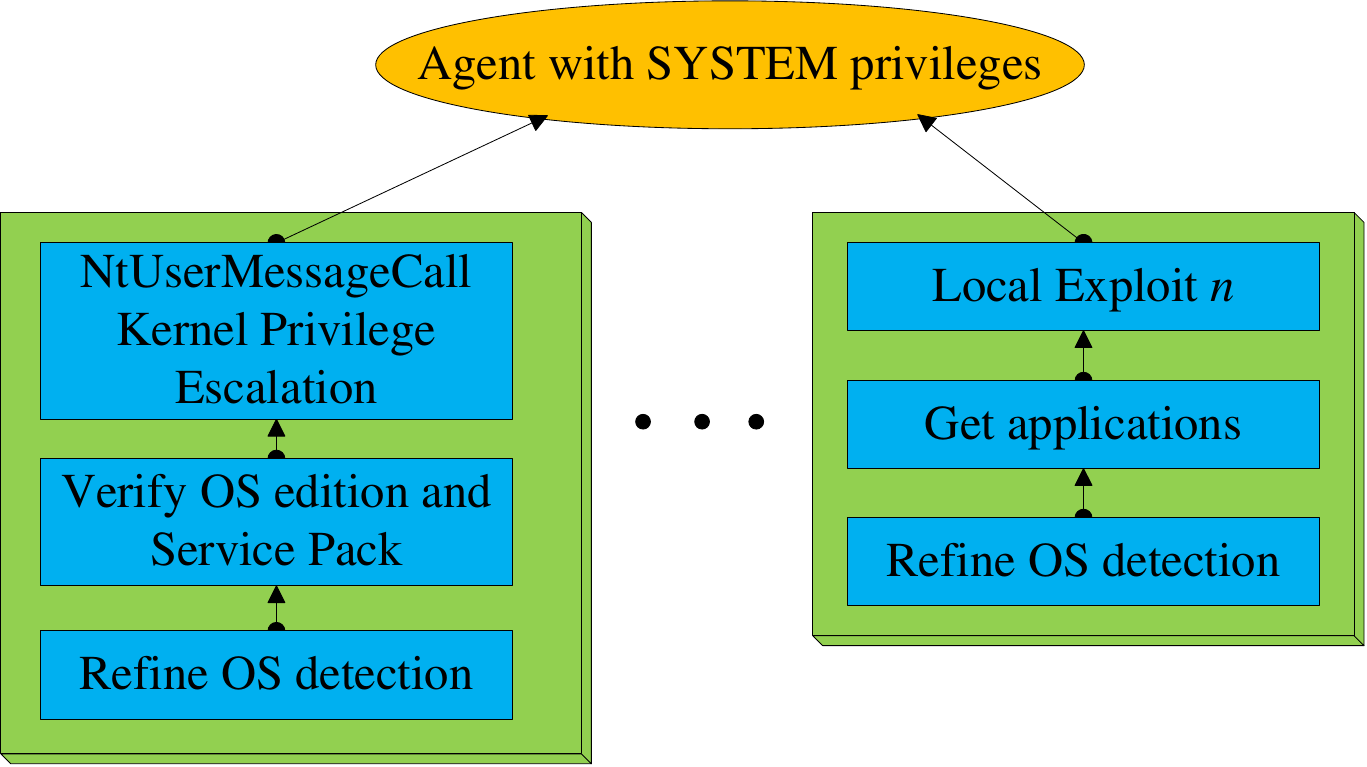}{ht}{ 0.85 \linewidth}{grafo2}
{Multiple strategies for a Local Privilege Escalation.}

For example consider an attacker
who has installed an agent with low privileges on a host $H$ 
running Windows XP, and whose goal
is to obtain \textsc{system} privileges on that host.
The attacker has a set of $n$ predefined strategies to 
perform this privilege escalation (see Fig.~\ref{fig:grafo2}).
An example of a strategy is:
refine knowledge of the OS version;
verify that the edition is Home or Professional, with SP2 installed;
get users and groups;
then launch the local exploit 
\Name{Microsoft NtUserMessageCall Kernel Privilege Escalation}
that (ab)uses the vulnerability CVE-2008-1084.
More generally:

\begin{problem}\label{group_action_order}
Let $\frakg$ be a fixed goal, and $\{ G_1, \ldots, G_n  \}$ a set 
of $n$ strategies,
where each strategy $G_k$ is a group of ordered actions.
For a strategy to be successful, all its actions must be successful.
As in Problem \ref{action_order}, the task is to minimize the expected total cost.

\end{problem}

In this problem, actions are executed sequentially,
choosing at each step one action from one group, until the goal $\frakg$ is obtained.
Considering only one strategy $G$, we can calculate
its expected cost and probability of success.
Suppose the actions of $G$ are $\{ A_1, \ldots, A_n \}$
and are executed in that order. Then the expected cost (e.g. expected runtime)
of the group $G$ is
$$
T_G = t_1 + p_1 \, t_2 + p_1 \, p_2 \, t_3 + \ldots
+ p_1 \, p_2 \ldots p_{n-1} \, t_n
$$
and, since all the actions must be successful,
the probability of success of the group is simply
$ P_G = p_1 \, p_2 \ldots p_n $ (again, we suppose that the actions are independent).

\begin{proposition}\label{prop_group_action_order}
A solution to this problem is to sort the 
strategies according to the 
coefficient $T_G / P_G$ (smallest value first), and execute them in that order.
For each strategy group, execute the actions until
an action fails or all the actions are successful.
\end{proposition}

\begin{proof}
In this problem, an attack plan could involve choosing actions
from different groups without completing all the actions of each group.
But it is clear that this cannot happen in an optimal 
plan.

In effect, suppose that there are only two groups $G_A$ and $G_B$,
whose actions are $\{ A_1, \ldots, A_s \}$ and $\{ B_1, \ldots, B_t \}$ respectively.
Suppose that in the optimal plan $A_s$ precedes $B_t$.
Suppose also that the execution of an action $B_j \neq B_t$ precedes the 
execution of $A_s$.
Executing $B_j$ will not result in success (that requires executing $B_t$ as well),
and it will delay the execution of $A_s$ by 
the expected running time of $B_j$.
Thus to minimize the expected total running time, a better solution 
can be obtained by executing $B_j$ after the execution of $A_s$.
This contradiction shows that all the actions of $G_B$ must be executed after $A_s$
in an optimal solution.
This argument can be easily extended to any number of groups.

So an optimal attack plan consists in choosing a group
and executing all the actions of that group.
Since the actions of each group $G$ are executed one after the other,
they can be considered as a single action with probability $P_G$ and expected time $T_G$.
Using the {\em choose} primitive, it follows that groups should be ordered
according to the coefficients $T_G / P_G$.
\end{proof}

\subsection{Multiple Groups of Actions} \label{sec:two-layers-tree}

We extend the previous problem to consider groups of actions
bounded by an AND relation (all the actions of the group must be
successful in order to obtain the result $\frakg$),
but where the order of the actions
is not specified.
The difference with Problem \ref{group_action_order} is that now
we must determine the order of execution within each group.

\Figure{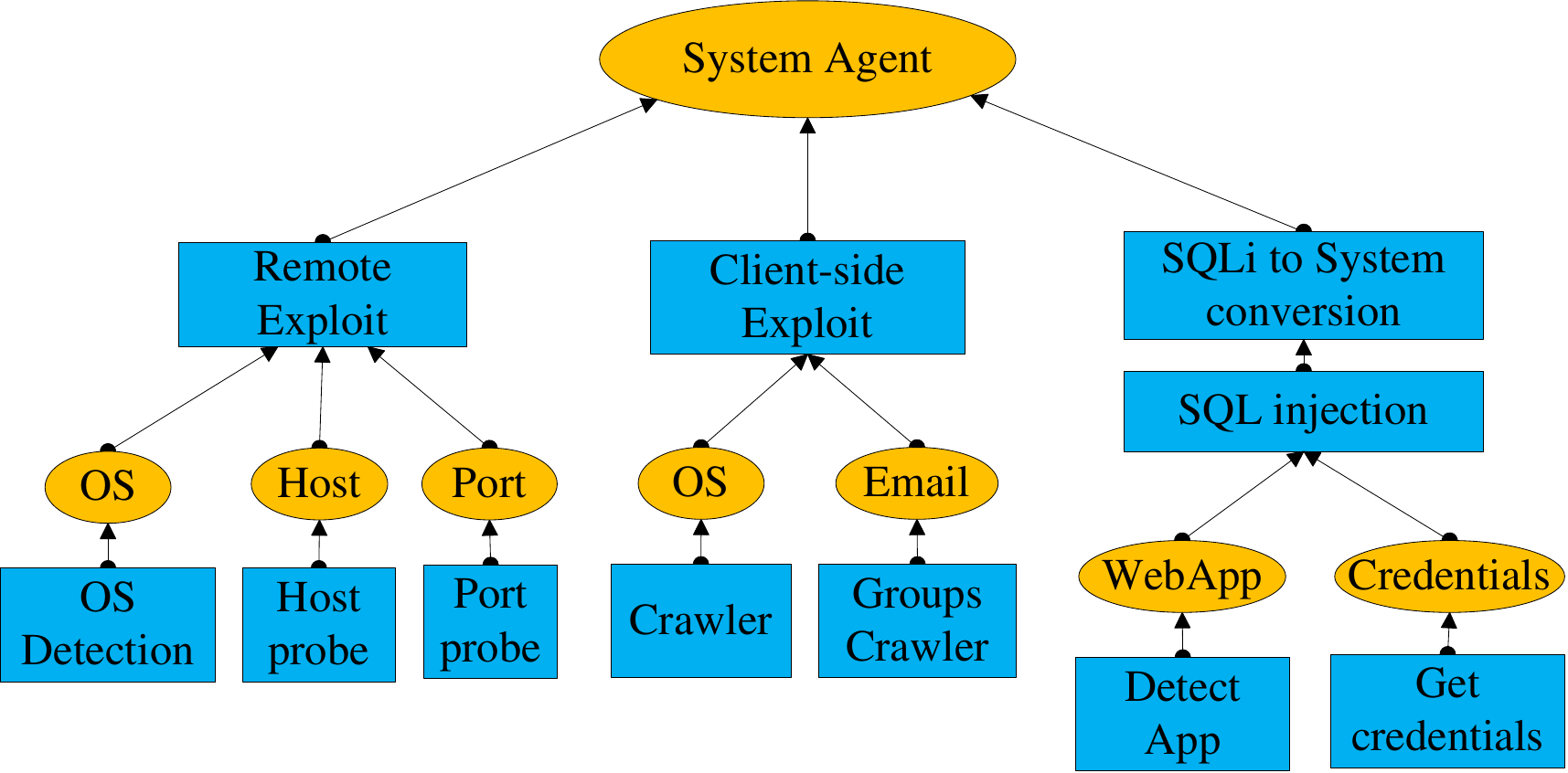}{ht}{\linewidth}{grafo3}
{Probabilistic attack tree (with two layers).}

Fig.~\ref{fig:grafo3} shows an example of this situation. A System Agent
can be installed by using a Remote exploit, a Client-side exploit or 
a SQL injection in a web application.
Each of these actions has requirements represented as assets, which can be
fulfilled by the actions represented on the second layer.
For example, before executing the Remote exploit, the attacker
must run a Host probe (to verify connectivity with the target host),
Port probe (to verify that the target port of the exploit is open),
and an OS Detection module (to verify the OS of the target host).

\begin{problem}\label{free_group_action_order}
Same as Problem \ref{group_action_order}, except that we have
$n$ groups $\{ G_1, \ldots, G_n  \}$ of unordered actions.
If all the actions in a group are successful,
the group provides the result $\frakg$.
\end{problem}

\begin{proposition}\label{order_and_group}
Let $G = \{ A_1, \ldots, A_n \}$ be a group of actions
bounded by an AND relation.
To minimize the expected total cost, the actions must be ordered
according to the coefficient $t_k / (1 - p_k)$.
\end{proposition}

\begin{proof}
If the actions are executed in the order $A_1, \ldots, A_n$, then 
the expected cost is
\begin{align}\label{exp_time_and}
T_G = t_1 + p_1 \, t_2 +  
\ldots + p_1 \, p_2 \ldots p_{n-1} \, t_n
\end{align}
This expression is very similar
to equation \eqref{exp_time_or}.
The only difference is that costs are multiplied by $p_k$ instead of $\overline{p_k}$.
So in this case, the optimal solution is to order the actions according 
to the coefficient $t_k / \overline{p_k} = t_k / (1 - p_k)$. 
\end{proof}

Intuitively the actions that have higher probability of failure 
have higher priority, since a failure ends the execution of the group.
The coefficient $t_k / (1 - p_k)$ represents a trade-off between
cost (time) and probability of failure.

Wrapping up the previous results, to solve Problem \ref{free_group_action_order},
first order the actions in each group according to the coefficient $t / (1-p)$ in increasing order.
Then calculate for each group $G$ the values $T_G$ and $P_G$.
Order the groups according to the 
coefficient $T_G / P_G$, and select them in that order.
For each group, execute the actions until
an action fails or all the actions are successful.

We call it the {\em combine} primitive, because it tells you
how to combine a group of actions and consider them
(for planning purposes) as a single action with probability
of success $P_G$ and expected running time $T_G$.

\section{Using the Primitives in an Attack Tree}\label{sec:dynamic}

We apply below the {\em choose} and the {\em combine} primitives
to a probabilistic attack tree, where the nodes are bounded 
by AND relations and OR relations.
The tree is composed of two types of nodes, distributed 
in alternating layers of asset nodes and action nodes (see Fig.~\ref{fig:grafo4}).

\Figure{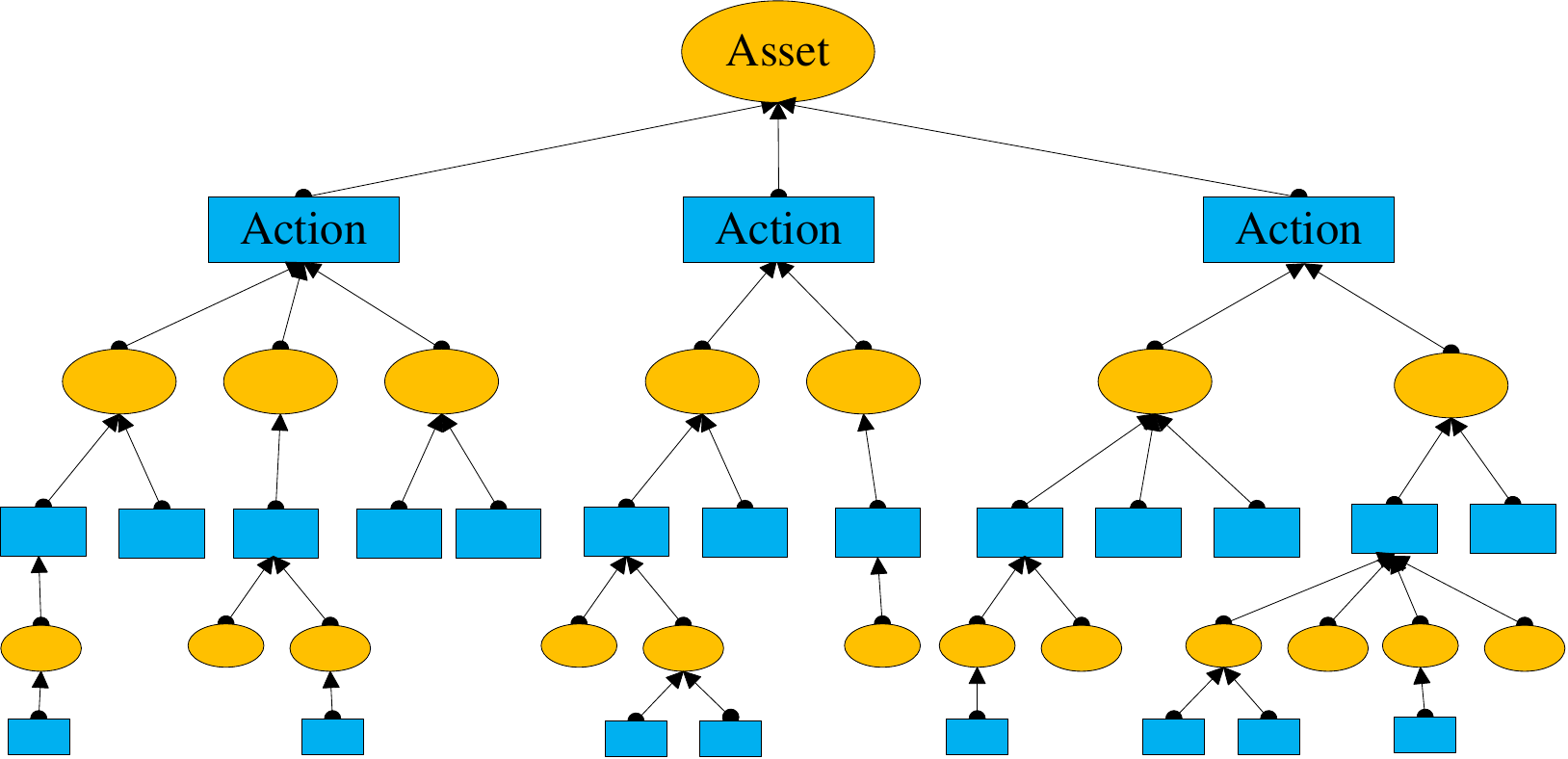}{ht}{\linewidth}{grafo4}
{Attack tree with alternating layers of Assets and Actions.}

An \emph{asset node} is connected by an OR relation to all the actions
that provide this asset: for example, an Agent asset
is connected to the Exploit actions that may install an agent on the target host.

An \emph{action node} is connected by an AND relation to its requirements:
for example, the local exploit 
\Name{Microsoft NtUserMessageCall Kernel Privilege Escalation}
requires an agent asset (with low level privileges) on the target host $H$,
and a Windows XP OS asset for $H$.

The proposed solution is obtained by composing the primitives from previous sections.
In the AND-OR tree, 
the leaves that are bounded by an AND relation 
can be considered as a single node. 
In effect, using the {\em combine} primitive, 
that group $G$ can be considered
as a single action with compound probability of success $P_G$ and execution time $T_G$.

The leaves that are bounded by an OR relation can also
be (temporarily) considered as a single node.
In effect, in an optimal solution, the node that minimizes the $t/p$ coefficient
will be executed first (using the {\em choose} primitive), 
and be considered as the cost of the group in a single step plan.

By iteratively reducing groups of nodes, we build a single path of execution
that minimizes the expected cost.
After executing a step of the plan, the costs may be modified and the shape of the graph may vary.
Since the planning algorithm is very efficient, we can replan after each execution
and build a new path of execution.
We are assured that before each execution, the proposed attack plan is optimal
given the current environment knowledge and within the horizon of a single next step.
It is still only an approximation to the global optimal plan, whose computation
requires a much powerful (and computationally expensive) framework such as 
the POMDP framework of Chapter~\ref{chap:pomdp}.

\subsection{Constructing the Tree} \label{sec:tree-construction}

We briefly describe how to construct a tree beginning with an agent asset
(e.g. the objective is to install an agent on a fixed machine). 
Taking this goal as root of the tree, we recursively add the actions that 
can complete the assets that appear in the tree,
and we add the assets required by each action.

To ensure that the result is a tree and not a DAG, we make an additional 
independence assumption: the assets required by each action are considered
as independent (i.e. if an asset is required by two different actions,
it will appear twice in the tree).

That way we obtain an AND-OR tree with alternating layers of asset nodes 
and action nodes (as the one in Fig.~\ref{fig:grafo4}).
The only actions added are Exploits, TCP/UDP Connectivity
checks, and OS Detection modules. These actions don't have as requirements
assets that have already appeared in the tree, in particular 
the tree only has one agent asset (the root node of the tree).
So, by construction, we are assured that no loops will appear,
and that the depth of the tree is very limited.

We construct the tree in this top-down fashion,
and as we previously saw, we can solve it bottom-up to obtain
as output the compound probability of success and the expected running time
of obtaining the goal agent.

\section{The Graph of Distinguished Assets} \label{sec:distinguished-assets}

In this section we use the previous primitives to build an algorithm
for \emph{attack planning} in arbitrary networks,
by making an additional assumption of independence between machines.
First we distinguish a class of assets, namely the assets related with agents.
We refer to them as \emph {distinguished assets}.
At the PDDL level, the predicates associated with the agents are considered
as a separate class.

Planning is done in two different abstraction levels: in the \textbf{first level}, we evaluate the cost of 
compromising one \emph{target} distinguished asset from one fixed \emph{source} distinguished asset.
More concretely, we compute 
the cost and probability of obtaining a target agent given a source agent. At this level, the attack plan 
must not involve a third agent.
The algorithm at the first level is thus to construct the attack tree
and compute an attack plan as described in Section \ref{sec:dynamic}.

At the \textbf{second level}, we build a directed graph $\cG = (\cV, \cE)$
where the nodes are distinguished assets
(in our scenario, the hosts in the target network where we may install agents),
and the edges are labeled with the compound probability and expected time
obtained at the first level.
Given this graph, an initial asset $s \in \cV$ (the local agent of the attacker)
and a final asset $g \in \cV$ (the goal of the attack), 
we now describe two algorithms to find a path that 
approximates the minimal expected time of obtaining the goal $g$.

The first algorithm is a modification of Floyd-Warshall's algorithm to find
shortest paths in a weighted graph.
Let $M = | \cV | $ be the number of machines in the target network.
By executing $M^2$ times the first level procedure, we obtain two functions:
the first is $Prob(i,j)$ which returns the compound probability
of obtaining node $j$ from node $i$ (without intermediary hops), 
or $0$ if that is not possible in the target network;
the second is $Time(i,j)$ which returns the expected time
of obtaining node $j$ directly from node $i$, or $+\infty$ if that is not possible.
The procedure is described in Algorithm~\ref{floyd-warshall}.

\begin{algorithm}[ht]
\caption{Modified Floyd-Warshall}
\label{floyd-warshall}
\algsetup{indent=1.5em}
\fontsize{12}{14}\selectfont
\begin{algorithmic}
\STATE $P[i,j] \gets Prob(i,j) \quad \forall \; 1 \leq i,j \leq M $
\STATE $T[i,j] \gets Time(i,j) \quad \forall \; 1 \leq i,j \leq M $

\FOR{$k = 1$ to $M$}
  \FOR{$i = 1$ to $M$}
    \FOR{$j = 1$ to $M$}
      \STATE $ T' \gets T[i,k]  +  P[i,k] \times T[k,j] $
      \STATE $ P' \gets P[i,k] \times P[k,j] $
      \IF{${T'}/{P'} \, < \, T[i,j] / P[i,j]$}
        \STATE $T[i,j] \gets T' $
        \STATE $P[i,j] \gets P' $
      \ENDIF
    \ENDFOR
  \ENDFOR
\ENDFOR
\RETURN $ \langle T, P \rangle $
\end{algorithmic}
\end{algorithm}

When the execution of this algorithm finishes, for each $i,j$ 
the matrices contain the compound probability $P[i,j]$ and
the expected time $T[i,j]$ of obtaining the node $j$
starting from the node $i$. This holds in particular when $i = s$ (the source
of the attack) and $j = g$ (the goal of the attack).
The attack path is reconstructed just as in the classical Floyd-Warshall algorithm.

In a similar fashion, Dijkstra's shortest path algorithm can be modified to use
the {\em choose} and {\em combine} primitives. See the description of Algorithm \ref{dijkstra}.

{
\begin{algorithm}[ht]
\caption{Modified Dijkstra's algorithm}
\label{dijkstra}
\algsetup{indent=1.5em}
\fontsize{12}{14}\selectfont
\begin{algorithmic}

\STATE $ T[s] = 0,\; P[s] = 1 $
\STATE $ T[v] = +\infty,\; P[v] = 0 \quad \forall v \in \cV, v \neq s $
\STATE $ S \gets \emptyset $
\STATE $ Q \gets \cV $ (where $Q$ is a priority queue)
\WHILE{$ Q \neq \emptyset $}
  \STATE $ u \gets \arg \min_{x \in Q} T[x]/P[x]$
  \STATE $ Q \gets Q \backslash \{ u \}, \; S \gets S \cup \{ u \} $
  \FORALL{ $v \in \cV \backslash S $ adjacent to $u$ }
    \STATE $ T' = T[u] + P[u] \times Time(u,v) $
    \STATE $ P' = P[u] \times Prob(u,v) $
    \IF{${T'}/{P'} < T[v] / P[v] $}
      \STATE $ T[v] \gets T' $
      \STATE $ P[v] \gets P' $
    \ENDIF
  \ENDFOR
\ENDWHILE
\RETURN $ \langle T, P \rangle $
\end{algorithmic}
\end{algorithm}
}

When execution finishes, the matrices contain the compound probability $P[v]$ and
the expected time $T[v]$ of obtaining the node $v$
starting from the node $s$.
Using the modified Dijkstra's algorithm has the advantage that its complexity
is $\calO( M ^ 2)$ instead of $\calO( M ^3 ) $ for Floyd-Warshall.
Let $n$ be the number of actions that appear in the attach trees,
this gives us that the complexity of the complete planning solution is
$\calO( M^2 \cdot n \log n + M^2 ) = \calO( M^2 \cdot n \log n )$.

\section{Our implementation} \label{sec:experiments}

We have developed a proof-of-concept implementation of these ideas
in the Python language. This planner takes as input a description of the scenario
in the PPDDL language, an extension of PDDL for expressing probabilistic effects \cite{YouLit04}.

Our main objective was to build a probabilistic planner
able to solve scenarios with 500 machines, which was the limit reached with classical
(deterministic) planning solutions in \cite{LucSarRic10}.
Additionally we wanted to tame memory complexity, which was the limiting factor.
The planner was integrated with the pentesting framework Core Impact, 
using the procedures previously developed for the work \cite{LucSarRic10}.
The architecture of this solution is described in Fig.~\ref{fig:prob-architecture}.

\Figure{img_deterministic/architecture.pdf}{ht}{ 0.9 \linewidth}{prob-architecture}
{Architecture of our solution.}

This planner solves the planning problem by breaking it into two levels
as described in Section~\ref{sec:distinguished-assets}.
On the higher level, a graph representation of \emph{goal} objects is built. 
More concretely, there is a distinguished node for each host.
The directed edges in this graph are obtained by carrying out the tree procedure 
described in Section \ref{sec:tree-construction},
obtaining a value for the probability and the cost of obtaining the predicate represented 
by the target node, when the predicate represented by the source node is true. 

The final plan can then be determined by using the modified versions
of Dijkstra and Floyd-Warshall algorithms.
The figures that follow show the planner running time using
the modified Dijkstra's algorithm.

\subsection{Testing and Performance}

The experiments were run on a machine with an Intel Core2 Duo CPU at 2.4 GHz and 8 GB of RAM.
We focused our performance evaluation on the number of machines $M$ in the attacked network.
We generated a network consisting of five subnets with varying number of machines,
all joined to one main network to which the attacker initially has access.

\Figure{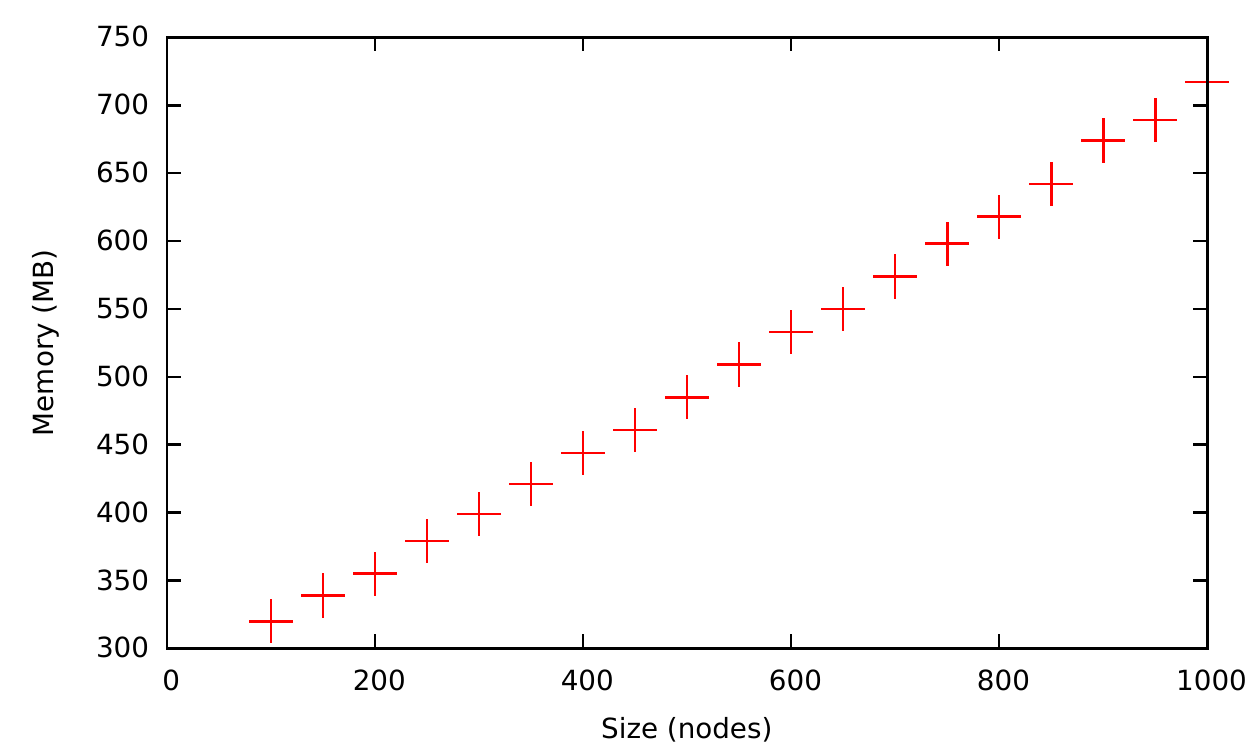}{ht}{ 0.9 \linewidth}{memabs}
{Memory consumption vs number of machines.}

\Figure{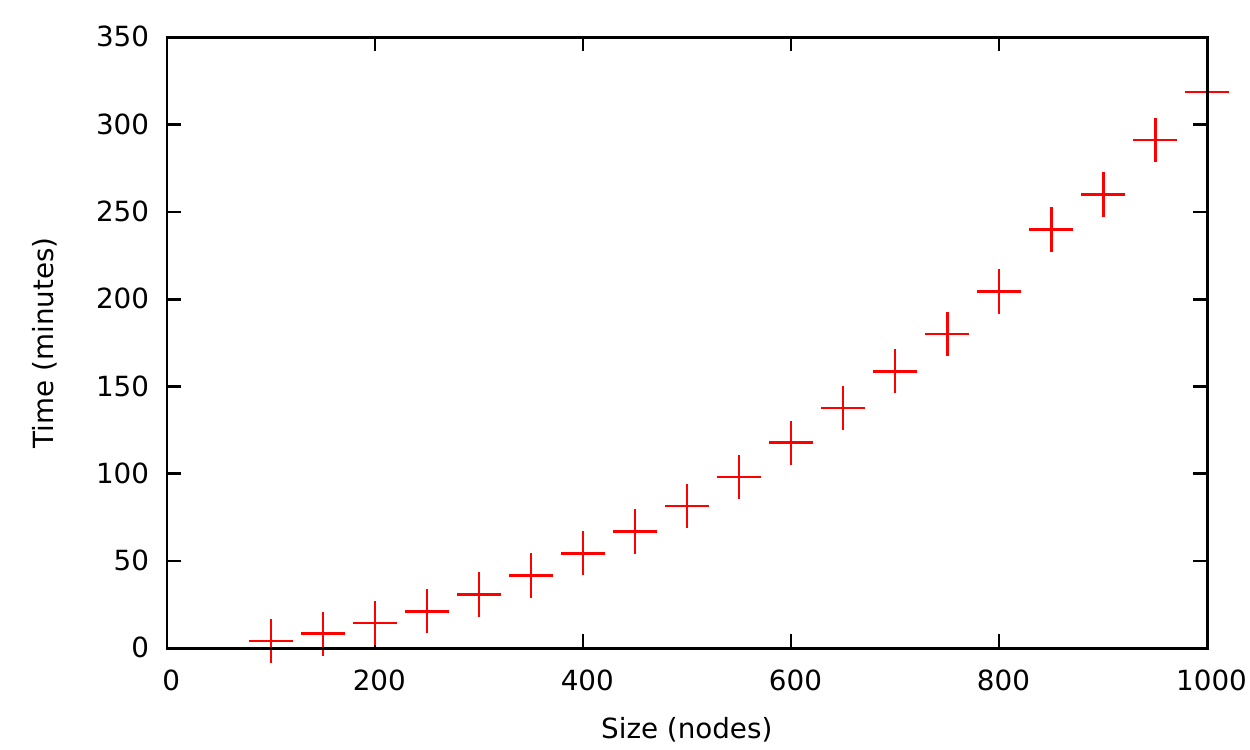}{ht}{ 0.9 \linewidth}{timeabs}
{Solver runtime vs number of machines.}

\Figure{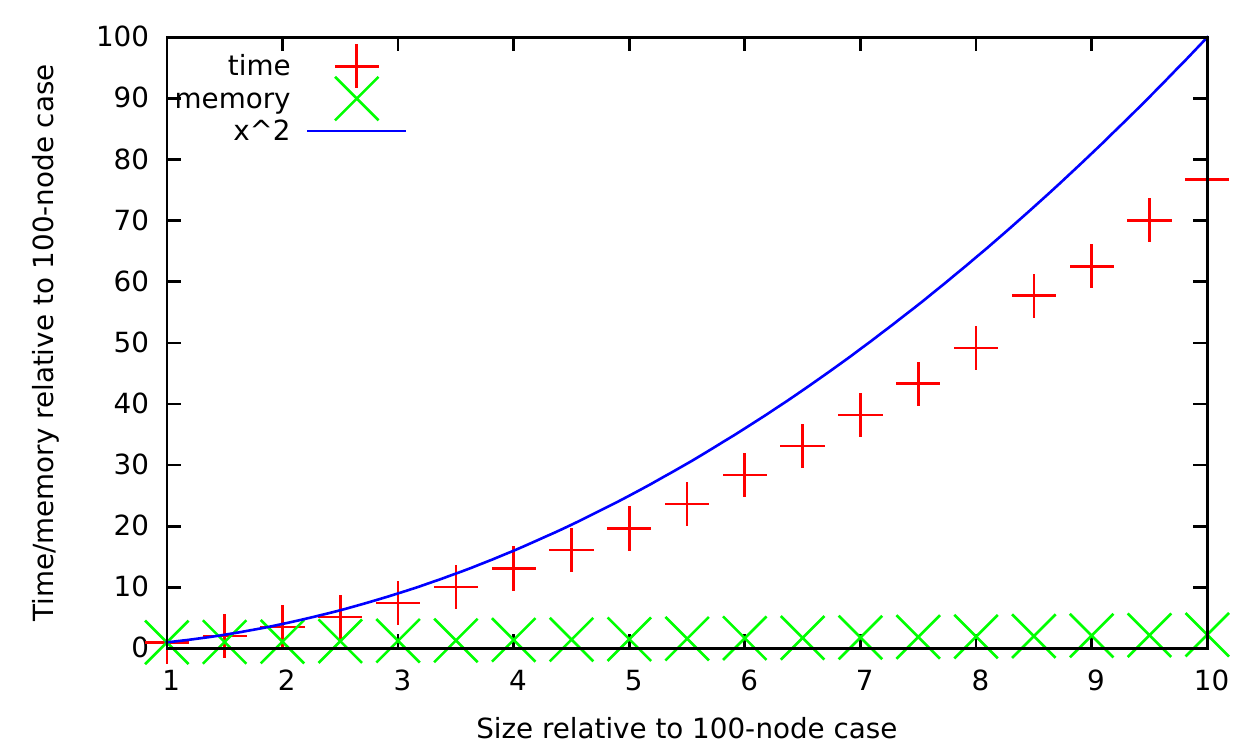}{ht}{ 0.9 \linewidth}{plotrel}
{Time and memory relatives to the 100 machines case.}

Fig.~\ref{fig:memabs} shows the memory consumption of this planning solution, 
which clearly grows linearly with $M$.
Our current implementation manages to push the network size limit up to 1000 machines, and brings memory 
consumption under control.\footnote{By contrast, in \cite{LucSarRic10} the hard limit was memory:
in scenarios with 500 machines we ran out of memory in a computer with 8 GB of RAM.
The memory consumption growth was clearly exponential, for instance
400 machines used 4 GB of RAM. This was difficult to scale up. }
For $M = 1000$, we are using less than 1 GB of RAM,
with a planner completely written in Python (not optimized in terms of memory consumption).

Fig.~\ref{fig:timeabs} shows the growth of solver running time, which seems clearly quadratic,
whereas in \cite{LucSarRic10} the growth was exponential.
It should be noted however that, comparing only up to 500 machines, 
running times are slightly worse than those of the solution based 
on deterministic planners.
This can be improved: since our planner is written in Python, a reasonable implementation in C 
of the more CPU intensive loops should allow us to lower significantly the running time. 

And of course we added a notion of probability of success that wasn't present before.
As a comparison, in another approach that accounts for the uncertainty about
the attacker's actions \cite{SarBufHof11},
the authors use off-the-shelf solvers, managing to solve scenarios with up to 7 machines -- 
and are thus still far from the network sizes reached here.

Both curves are compared in Fig.~\ref{fig:plotrel} showing the quadratic growth of solver runtime.
In the testing scenarios, the nodes are fully connected,
so we have to solve a quadratic number of attack trees.
This figure also confirms in practice the computed complexity.

An interesting characteristic of the solution proposed is that
it is inherently parallelizable. 
The main workload are the $M^2$ executions of the \emph{first level} procedure 
of Section~\ref{sec:distinguished-assets}.
This could be easily distributed between CPUs or GPUs to obtain a faster planner.
Another possible improvement is to run the planner ``in the cloud'' 
with the possibility of adding processors on demand.

\section{Related Work}

Early work on attack graph solving relied on model checking techniques \cite{JhaSheWin02,SheHaiJha02},
with their inherent scalability restrictions; or on monotonicity assumptions \cite{AmmWijKau02,NoeEldJaj09,NoeJaj05} 
that are not able to express situations in which compromised resources are lost due to crashes, 
detection or other unforeseen circumstances.

The first application of planning techniques and PDDL solving for the security realm was \cite{BodGohHaiHar05},
however this application was not focused on finding actual attack paths or driving penetration testing tools. 
In \cite{GhoGho09} attack paths are generated from PDDL description of networks, hosts and exploits, 
although the scenarios studied do not cover realistic scales. 
Previous work by the authors \cite{LucSarRic10} addresses this limitation by solving scenarios with up 500 machines, and feeding the generated attack plans to guide a penetration testing tool. However, this work does not include probabilistic considerations. Recent work \cite{ElsKohMen11} also manages to provide attack paths to a penetration testing tool, in this case the Metasploit Framework, but again does not include probabilistic considerations.

Previous work by one of the authors \cite{SarBufHof11} takes into account the uncertainty about the result of the attacker's actions. This POMDP-based model also accounts for the uncertainty about the target network, addressing information gathering as an integral part of the attack, and providing a comprehensive notion of attack planning under uncertainty. 
However, as previously stated, this solution does not scale to medium or large real-life networks.

\section{Summary and Future Work}

We have shown in this chapter an extension of
established \emph{attack graphs} models, 
that incorporates probabilistic effects,
and numerical effects (e.g. the expected running time of the actions).
This model is more realistic than the deterministic setting,
but introduces additional difficulties to the planning problem.
We have demonstrated that under certain assumptions,
an efficient algorithm exists that provides optimal attack plans
with computational complexity $\calO ( n \log n) $, where 
$n$ is the number of actions and assets in the case of an attack tree
(between two fixed hosts),
and $\calO ( M^{2} \cdot n \log n) $ where $M$ is the number of machines
in the case of a network scenario.

Over the last years, the difficulties that arose in our research
in \emph{attack planning} were related to the exponential nature of planning
algorithms (especially in the probabilistic setting), and our efforts
were directed toward the aggregation of nodes and simplification of the
graphs, in order to tame the size and complexity of the problem.
Having a very efficient algorithm in our toolbox gives us a new direction of research:
to refine the model, and break down the actions in smaller parts,
without fear of producing an unsolvable problem.

A future step in this research is thus to analyze and divide the exploits
into basic components. 
This separation gives a better probability distribution of the
exploit execution. For example, the
\Name{Debian OpenSSL Predictable Random Number Generation Exploit}
-- which exploits the vulnerability CVE-2008-0166 reported by Luciano Bello --
brute forces the 32,767 possible keys.
Each brute forcing iteration can be considered as a basic action, 
and be inserted independently in the attack plan.
Since the keys depend on the Process ID (PID), some keys are more probable than 
others.\footnote{The OpenSSL keys generated in vulnerable Debians only depend on the PID. 
Since Secure Shell usually generates the key in a new installation,
PIDs between 2,000 and 5,000 are more probable than the others.}
So the planner can launch the \Name{Debian OpenSSL PRNG} exploit, 
execute brute forcing iterations for the more probable keys, 
switch to others exploits and come to back to the Debian PRNG exploit if the others failed.
This finer level of control over the exploit execution should produce
significant gains in the total execution time of the attack.

Other research directions in which we are currently working are
to consider actions with multidimensional numeric effects
(e.g. to minimize the expected running time and generated network traffic 
\emph{simultaneously});
and to extend the algorithm to solve probabilistic attack planning in 
Directed Acyclic Graphs (DAG) instead of trees.
In this setting, an asset may influence the execution of several actions.
This relaxes the independence assumption of Sections \ref{sec:two-layers-tree} 
and \ref{sec:dynamic}.
Although finding a general algorithm that scales to the network sizes that we consider 
here seems a difficult task, we believe that efficient 
algorithms specifically designed for network attacks scenarios can be found.

\part{The Search for a Better Model}


\chapter{The POMDP Model} \label{chap:pomdp}

In Chapter \ref{chap:deterministic} we presented an approach 
to the \emph{attack planning} problem that uses classical planning and hence ignores all
the incomplete knowledge that characterizes hacking. 
The more recent approach of Chapter \ref{chap:probabilistic}
makes strong independence assumptions for the sake
of scaling, and lacks a clear formal concept of what the attack
planning problem actually {\em is}. In this chapter, we model that problem in
terms of partially observable Markov decision processes (POMDP). This
grounds penetration testing in a well-researched formalism,
highlighting important aspects of this problem's nature. 
POMDPs allow to model information gathering as an integral part of the
problem, thus providing for the first time a means to intelligently
mix scanning actions with actual exploits.

\section{Introduction}

The problem of automatically generating attacks to assess network security
is known in the AI Planning community as the ``Cyber
Security'' domain \cite{BodGohHaiHar05}. Independently (though
considerably later), the approach was put forward also by the
pentesting industry \cite{LucSarRic10}. The two domains essentially
differ only in the industrial context 
addressed.\footnote{Boddy et al. \cite{BodGohHaiHar05} provide a tool for system administrators to
  experiment with attacker models including behavioral attributes, and
  model also actions in the physical world, i.e., office
  environments.}
Herein, we are concerned exclusively with the specific context of
regular automatic pentesting, as in Core Security's ``Core Insight
Enterprise'' tool. We will use the term ``attack planning'' in that
sense.

Lucangeli et al.\ \cite{LucSarRic10} encoded attack planning into
PDDL, and used off-the-shelf planners. %
This already is useful,\footnote{In fact, this technology is currently
  employed in Core Security's commercial product, using a variant of
  Metric-FF.} however it is still quite limited. In particular, the
planning is classical---complete initial states and deterministic
actions---and thus not able to handle the uncertainty involved in this
form of attack planning. We herein contribute a planning model that
does capture this uncertainty, and allows to generate plans taking it
into account. To understand the added value of this technology, it is
necessary to examine the relevant context in some detail.

The pentesting tool has access to the details of the client
network. So why is there any uncertainty? The answer is simple:
pentesting is not Orwell's ``Big Brother''. Do {\em your} IT guys know
everything that goes on inside {\em your} computer?

It is safe to assume that the pentesting tool will be kept up-to-date
about the structure of the network, i.e., the set of machines and
their connections---these changes are infrequent and can easily be
registered. It is, however, impossible to be up-to-date regarding all
the details of the configuration of each machine, in the typical
setting where that configuration is ultimately in the hands of the
individual users. Thus, since the last series of attacks was
scheduled, the configurations may have changed, and the pentesting
tool does not know how exactly. Its task is to figure out whether any
of the changes open new dangerous vulnerabilities.

One might argue that the pentesting tool should first determine what
has changed, via {\em scanning} methods, and then address what is now
a classical planning problem involving only {\em exploits}, i.e.,
hacking actions modifying the system state. There are two flaws in
this reasoning: (a) scanning doesn't yield perfect knowledge so a
residual uncertainty remains; (b) scanning generates significant costs
in terms of running time and network traffic. So what we want is a
technique that (like a real hacker) can deal with uncertainty by
intelligently inserting scanning actions where they are useful for
scheduling the best exploits. To our knowledge, ours is the first work
that indeed offers such a method.

There is hardly any related work tackling uncertainty measures
(probabilities) in network security. The few works that exist (e.g.,
\cite{Bil03thesis,DawHal04}) are concerned with the defender's
viewpoint, and tackle a very different kind of uncertainty attempting
to model what an attacker would be likely to do. The above
mentioned work on classical planning is embedded into a pentesting
tool running a large set of scans as a pre-process, and afterwards
ignoring the residual uncertainty. This incurs both drawbacks (a) and
(b) above. The single work addressing (a) was performed in part by one
of the authors 
\cite{SarRicLuc11}. On the positive side, the proposed attack planner
demonstrates industrial-scale runtime performance, and in fact its
worst-case runtime is low-order polynomial. On the negative side, the
planner does not offer a solution to (b)---it still reasons only
about exploits, not scanning---and of course its efficiency is bought
at the cost of strong simplifying assumptions. Also, the work provides
no clear notion of what attack planning under uncertainty actually {\em is}.

Herein, we take the opposite extreme of the trade-off between accuracy
and performance. We tackle the problem in full, in particular
addressing information gathering as an integral part of the attack. We
achieve this by modeling the problem in terms of partially observable
Markov decision processes (POMDP). As a side effect, this modeling
activity serves to clarify some important aspects of this problem's
nature. A basic insight is that, whereas Sarraute et
al.\ \cite{SarRicLuc11} model the uncertainty as
non-deterministic actions---success probabilities of exploits---this
uncertainty is more naturally modeled as an uncertainty about {\em
  states}. The exploits as such are deterministic in that their
outcome is fully determined by the system
configuration.\footnote{Sometimes, non-deterministic effects are an
  adequate abstraction of state uncertainty, as in ``crossing the
  street''. The situation in pentesting is different because repeated
  executions will yield identical outcomes.}  Once this basic modeling
choice is made, all the rest falls into place naturally.

Our experiments are based on a problem generator that is not
industrial-scale realistic, but that allows to create reasonable test
instances by scaling the number of machines, the number of possible
exploits, and the time elapsed since the last activity of the
pentesting tool. Unsurprisingly, we find that POMDP solvers do not
scale to large networks. However, scaling is reasonable for individual
pairs of machines. As argued by Sarraute et
al.\ \cite{SarRicLuc11}, such pairwise strategies can serve as
the basic building blocks in a framework decomposing the overall
problem into two abstraction levels.

We next provide some additional background on pentesting and
POMDPs. We then detail our POMDP model of attack planning, and our
experimental findings. We close the chapter with a brief discussion of
future work.

\section{Comment on Penetration Testing}

\begin{figure}[tbp]
  \centerline{
    \resizebox{ 0.8 \linewidth}{!}{
      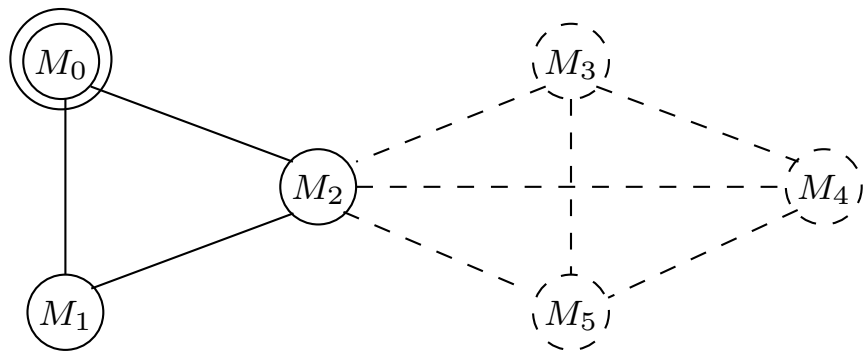
    }
  }
  \caption[An example network made of two sub-networks (cliques).]
  {An example network made of two sub-networks (cliques):
    $(M_0, M_1, M_2)$ (e.g., $M_0$ an outside computer, $M_1$ a web
    server and $M_2$ a firewall), and $(M_2, M_3, M_4, M_5)$. 1
    computer is under control ($M_0$), 2 are reachable ($M_1$ and
    $M_2$), and 3 are unreachable ($M_3$, $M_4$, $M_5$).}
  \label{fig:network}
\end{figure}

The objective of a typical penetration testing task is to gain control
over as many computers in a network as possible, with a preference for
some machines (e.g., because of their critical content). It starts
with one controlled computer: either outside the targeted network (so
that its first targets are machines accessible from the internet), or
inside this network (e.g., using a Trojan horse). As illustrated in
Figure~\ref{fig:network}, at any point in time one can distinguish
between 3 types of computers: those under control (on which an agent
has been installed, allowing to perform actions); those which are
reachable from a controlled computer because they share a sub-network
with one of them: and those which are unreachable from any controlled
computer.

Given currently controlled machines, one can perform two types of
actions targeting a reachable machine: tests---to identify its
configuration (OS, running applications, \dots)---, and exploits---to
install an agent by exploiting a vulnerability. A successful exploit
turns a reachable computer into a controlled one, and all its
previously unreachable neighbors into reachable computers.

A ``classic'' pentest methodology consists of a series of fixed steps 
(refer also to Section \ref{sec:main_steps}),
for example:
\begin{itemize*}
\item perform a network discovery (obtain a list of all the reachable
  machines),
\item port scan all the reachable machines (given a fixed list of
  common ports, probe if they are open/closed/filtered),
\item given the previous information, perform OS detection module(s)
  on reachable machines (e.g., run nmap tests),
\item once the information gathering phase is completed, the following
  phase is to launch exploits against the (potentially vulnerable)
  machines.
\end{itemize*}
This could be improved---a long-term objective of this
work---as POMDP planning allows for more efficiency by mixing actions
from the different steps.

More details on pentesting will be given later when we describe how to
model it using the POMDP formalism.

\section{Background on POMDPs}

Before describing how penetration testing can be modeled in terms of POMDPs
in Section~\ref{sec:pomdp-modeling}, we provide the necessary background and
terminology.
Planning based on Markov Decision Processes (MDPs) is designed
to deal with nondeterminism, probabilities, partial observability,
and extended goals. It is based on the following conventions:
\begin{itemize}
	\item {A planning domain is modeled as a \emph{stochastic system},
	that is a nondeterministic state-transition system that assigns
	probabilities to state transitions.}
	\item{Goals are represented by means of \emph{utility functions},
	numeric funtions that give preferences to states to be traversed
	and/or actions to be performed. Utility functions can express
	preferences on the entire execution path of a plan, rather than just
	desired final states.}
	\item{Plans are represented as \emph{policies} that specify the action
	to execute in each belief state.}
	\item{The planning problem is seen as an \emph{optimization problem},
	in which planning algorithms search for a plan that maximizes the utility function.}
	\item{Partial observability is modeled by observations that return a probability
	distribution over the state space, called \emph{belief states}.}
\end{itemize}

\subsection{Basic Definitions of MDPs}

\begin{definition}
A Markov Decision Process (MDP), also called {\em stochastic system} \cite{GhaNauTra04},
is a nondeterministic state-transition system with a probability distribution on each
state transition. It is defined by a tuple 
$\Sigma = \langle \cS, \cA, T \rangle$ 
where:
\begin{itemize}
	\item {The {\em state space} $\cS$ is a finite set of states.}
	\item {The {\em action space} $\cA$ is a finite set of actions.}
	\item {$T : \cS \times \cA \rightarrow \Pi(\cS)$ is the {\em state-transition function},
	giving for each world state $s$ and agent action $a$, a probability distribution 
	over world states.
	We write $T(s,a,s')$ for the probability of ending in state $s'$ given that the agent
	starts in state $s$ and executes action $a$. 
	}
  \item {$r : \cS \times \cA \rightarrow \R $ is the {\em reward function},
  giving the expected immediate reward gained by the agent for taking action $a$ in state $s$.
  }
\end{itemize}
\end{definition}

\begin{definition}
A plan specifies the actions that a controller should execute in a given state,
and can thus be represented as a {\em policy} $\pi$, a function mapping states into actions:
\begin{equation}
 \pi : \cS \rightarrow \cA 
 \end{equation}

\end{definition}

\begin{definition}
Policy executions correspond to infinite sequence of states, called {\em histories},
which are Markov Chains.
Given a policy, we can compute the probability of a history.
Let $\pi$ be a policy and $h = \langle s_0, s_1, s_2, \ldots \rangle$ be a history.
The probability of $h$ induced by $\pi$ is the product of all transition
probabilities induced by the policy:
\begin{equation}
 Pr(h | \pi) = \prod_{i \geq 0} T (s_i, \pi(s_i), s_{i+1} )
\end{equation}
\end{definition}

\begin{definition}
A common way to ensure a bounded measure of utilities for infinite histories
is to introduce a {\em discount factor} $\gamma$, with $ 0 < \gamma < 1 $,
that makes rewards accumulated at later stages count less than those accumulated
at early stages.
\end{definition}

\begin{definition}
Let $h$ be a history $\langle s_0, s_1, s_2, \ldots \rangle$.
We define the utility of $h$ induced by a policy $\pi$ as follows:
\begin{equation}
V(h | \pi) = \sum_{i \geq 0} \gamma^{i} \; r(s_i, \pi(s_i))
\end{equation}
\end{definition}

\begin{definition}
Let $\Sigma$ be a stochastic system, $\cH$ be the set of all the possible histories of $\Sigma$,
and $\pi$ be a policy for $\Sigma$. Then the expected utility of $\pi$ is
\begin{equation}
E(\pi) = \sum_{h \in \cH} Pr(h | \pi) \, V(h|\pi)
\end{equation}
\end{definition}

\begin{definition}
A policy $\pi^{*}$ is an {\em optimal policy} for a stochastic system $\Sigma$
if $ E(\pi^*) \geq E(\pi) $, for any policy $\pi$ for $\Sigma$.
We can now define a planning problem as an optimization problem:
given a stochastic system $\Sigma$ and a utility function, a {\em solution}
to a planning problem is an {\em optimal policy}.
\end{definition}

\begin{definition}
Given an optimal policy $\pi^*$, $E(\pi^*)$ is called the {\em optimal reward}.
\end{definition}

\begin{definition}\label{def:expected-reward}
Let $E(s)$ be the expected reward in a state $s$.
We define $Q(s,a)$, the expected reward in a state $s$ when we execute action $a$:
\begin{equation}
Q(s,a) = r(s,a) + \gamma \sum_{s'\in \cS} T(s,a,s') \, E(s')
\end{equation}
\end{definition}

It can be shown that the optimal reward $E(\pi^*)$ satisfies the fixed-point equation
\begin{equation}\label{eq:bellman-equation}
E(s) = \max_{a \in \cA} Q(s,a)  \quad \text{for all } s \in \cS
\end{equation}
Formula \eqref{eq:bellman-equation} is called the {\em Bellman Equation}~\cite{Bellman54dynprog}.
We call $E_{\pi^*} (s)$ the optimal reward in state $s$.
From the Bellman Equation we have that:
\begin{equation}
E_{\pi^*} (s) = \max_{a \in \cA} \left\{ r(s,a) + \gamma 
 \sum_{s'\in \cS} T(s,a, s') \, E_{\pi^*}(s') \right\}
\end{equation}

\subsection{Basic Definitions of POMDPs}

\begin{definition}
POMDPs are usually defined \cite{Monahan82,Cassandra98thesis,GhaNauTra04} by a tuple 
$\Sigma = \langle \cS, \cA, \cO, T, O, r, b_0 \rangle$ 
where:
\begin{itemize}
	\item {$\langle \cS, \cA, T, r \rangle$ is a Markov decision process.}
	\item {The {\em observation space} $\cO$ is a finite set of observations.}
	\item { $b_0$ is the initial probability distribution over states.}
	\item {At any time step,
the system being in some state $s \in \cS$,
the agent performs an action $a\in \cA$ that
results in 
\begin{enumerate}
	\item {a transition to a state $s'$ according to 
	the {\em transition function}:
    \begin{align*}
    T : \cS \times \cA \times \cS &\rightarrow [0, 1] \\
    T(s,a,s') &= Pr(s'| s,a) 
    \end{align*}
    }
  \item {an observation $o\in\cO$ according to 
  the {\em observation function}:
  \begin{align*}
  O : \cS \times \cA \times \cO &\rightarrow [0, 1] \\
   O(s',a,o) &= Pr(o| s',a) 
   \end{align*}
   }
  \item {a scalar reward $r(s,a)$ according to the {\em reward function}:
  \begin{align*}
		r : \cS \times \cA \rightarrow \R  
   \end{align*}
  }
\end{enumerate}
}
\end{itemize}
\end{definition}

\begin{definition} \label{def:belief-state}
In POMDPs, the controller can observe a probability distribution
over states of the system, rather than exactly the state of the system.
Probability distributions over $\cS$ are called {\em belief states}.
Let $b$ be a belief state and $\cB = \Pi(\cS) $ the set of belief states.
Let $b(s)$ denote the probability assigned to state $s$ by the belief state $b$.
Because $b(s)$ is a probability, we require $0 \leq b(s) \leq 1$ for all $s \in \cS$
and $\sum_{s \in \cS} b(s) = 1$.
\end{definition}

\begin{definition}
In POMDPs, a {\em policy} is a function that maps belief states into actions.
Let $\cB \subseteq [0,1] ^{|\cS|}$ be the set of belief states. A policy is a function 
\begin{equation}
 \pi : \cB \rightarrow \cA 
 \end{equation}
 
\end{definition}

In this setting, the problem is for the agent to find a decision policy $\pi$ 
choosing, at each time step, the best action based on
its past observations and actions so as to maximize its future gain
(which can be measured for example through the total accumulated
reward). Compared to classical
deterministic planning, the agent has to face the difficulty in
accounting for a system not only with uncertain dynamics but also whose
current state is imperfectly known.

Given a belief state $b$, the execution of an action $a$ results in a temporary
belief state $b_a$ (before the observation is made). 
For each $s'' \in \cS$, the probability $b_a(s'')$ can be
computed as:
\begin{align}
 b_a(s'') = Pr(s''|a,b) &= \sum_{s \in \cS} Pr(s'' | a,s) \, b(s) \nonumber \\
 &= \sum_{s \in \cS} T(s,a,s'') \, b(s)
\end{align}
The probability of observing $o \in \cO$ after executing action $a \in \cA$
in belief state $b$ is:
\begin{align}
 Pr(o | a,b) = \sum_{s''\in \cS} O(s'',a,o) \, b_a(s'')
\end{align}
The agent typically reasons about the hidden state of the system 
using the following Bayesian update formula,
which gives the probability $b_a^o(s')$ that the state is $s'$ 
after performing action $a$ in belief state $b$ and observing $o$:
\begin{align}
  b_a^{o}(s') = Pr(s'|o,a,b) &= \frac{Pr(o | s',a,b)}{Pr(o | a,b)} Pr(s' | a,b) \nonumber \\
  &= \frac{O(s',a,o)}{Pr(o| a,b)} \sum_{s\in \cS} T(s,a,s') b(s) .
\end{align}

\subsection{Reformulation as an MDP over $\cB$}

Using belief states, a POMDP can be rewritten as an MDP over
the belief space, or {\em belief MDP}, $\Sigma'= \langle \cB, \cA, \cT, \rho
\rangle$, where the new transition and reward functions are both
defined over $\cB\times\cA\times\cB$. With this reformulation, a
number of theoretical results about MDPs can be extended, such as the
existence of a deterministic policy that is optimal. An issue is that this
belief MDP is defined over a continuous---and thus infinite---belief
space.

Let us first compute the transition probabilities for belief states.
Given a belief state $b \in \cB$ and an action $a \in \cA$,
each observation can yield a different succeeding belief state.
The belief state transition $\tau$ can be written as:
\begin{align}
\tau(b,a,b') = Pr(b'| a,b) 
 = \sum_{ o \in \cO }  Pr(o | a,b) \; \delta (b_a^o, b') 
\end{align}
where
\begin{equation}
\delta (x,y) = 
\begin{cases}
 1 &\text{if } x = y  \\
 0 &\text{if } x \neq y .
\end{cases}
\end{equation}
In other words, the probability of a belief state is the sum of the probabilities
of all the observation that would lead to that belief state.

Since $\cA$ and $\cO$ are finite, there are only a finite number of possible
successor belief states. We define the set of possible successor states as:
\begin{align}
\cB'(b,a) = \{ b_a^o \; | \; o \in \cO \}
\end{align}

The reward $\rho$ for the belief MDP has to be defined for each belief state-action pair.
For the belief state $b$, it's simply the expectation over all states:
\begin{align}
\rho (b,a) = \sum_{s \in \cS} r(s,a) \, b(s) 
\end{align}

As in Definition \ref{def:expected-reward}, 
we define $Q(b,a)$, the expected reward in a state $b$ when we execute action $a$:
\begin{equation}
Q(b,a) = \rho(b,a) + \gamma \sum_{b'\in \cB} \tau(b,a,b') \, E(b')
\end{equation}

The Bellman Equation \eqref{eq:bellman-equation} is now written as:
\begin{equation}
E_{\pi^*} (b) = \max_{a \in \cA} \left\{ \rho(b,a) + \gamma 
 \sum_{b'\in \cB} \tau(b,a, b') \, E_{\pi^*}(b') \right\}
\end{equation}

For a finite horizon\footnote{In practice we consider an infinite
  horizon.} $T>0$ the objective is to find a policy verifying $\pi^* =
\arg\max_{\pi \in \cA^{\cB}} J^\pi(b_0)$ with
\begin{align*}
  J^{\pi}(b_0) &= E\left[ \sum_{t=0}^{T-1} \gamma^t r_t \middle| b_0, \pi \right],
\end{align*}
where $b_0$ is the initial belief state, $r_t$ the reward obtained at
time step $t$, and $\gamma \in (0,1)$ a discount factor. 

Bellman's principle of
optimality~\cite{Bellman54dynprog} lets us compute this function
recursively through the {\em value function}
\begin{align*}
  V_n(b) & = \max_{a\in\cA} \left[ \rho(b,a) + \beta \sum_{b' \in
      \cB} \varphi(b,a,b')V_{n-1}(b') \right],
\end{align*}
where, for all $b\in \cB$, $V_0(b)=0$, and $J^{\pi}(b)=V_{n=T}(b)$.

\subsection{POMDP Solving Algorithms}

A number of algorithms exploit the fact that $V_n(b)$ is a piecewise
linear and convex function (PWLC). This allows for direct computations
or approximations where the target function is the upper envelope of a
set of hyperplanes. This led to algorithms performing exact updates
like {\em Batch Enumeration} \cite{Monahan82}, {\em Witness} or {\em
  Incremental Pruning} \cite{Cassandra98thesis}, but also approximate
ones as in {\em Point-Based Value Iteration} (PBVI)
\cite{PinGorThr06}, {\em Heuristic Search Value Iteration} (HSVI)
\cite{SmiSim04}, PERSEUS \cite{SpaVla05} or SARSOP \cite{KurHsuLee08}.

If not relying on PWLC functions, one can also solve a POMDP as an MDP
on a continuous state space, for example with tree search
algorithms---which allow for online planning
\cite{RosPinPaqCha08jair}---or with RTDP-bel, a variant of dynamic
programming that continuously focuses on relevant parts of the state
space \cite{BonGef09}.

Let us also mention that some algorithms---such as Symbolic HSVI
\cite{SimKKCK08}---can exploit the structure of a factored POMDP,
i.e., a POMDP in which the state and/or the observation is described
through multiple variables.

\subsubsection{SARSOP Solver}

For our experiments we use SARSOP \cite{KurHsuLee08}, a state of the
art point-based algorithm, i.e., an algorithm approximating the value
function as the upper envelope of a set of hyperplanes, these
hyperplanes corresponding to a selection of particular belief points.

SARSOP is being developed at the National University of Singapore,
with the goal of creating practical POMDP algorithms and software
for common robotic tasks -- such as
coastal navigation, grasping, mobile robot exploration and
target tracking, all modeled as POMDPs with a large number of states.

The key idea of point-based POMDP algorithms is to \textit{sample}
a set of points from $\cB$ and use it as an approximate representation of $\cB$,
instead of representing $\cB$ exactly.

\begin{definition}
The reachable space $\cR(b_0)$ is the subset of belief points reachable from a given initial point $b_0 \in \cB$
under arbitrary sequences of actions.
\end{definition}

\begin{definition}
The optimally reachable space $\cR^{*}(b_0)$ is the subset of belief points reachable 
from a given initial point $b_0 \in \cB$
under {\em optimal} sequences of actions.
\end{definition}

Note that $\cR^{*}(b_0)$ is usually much smaller than $\cR(b_0)$, which is in turn
usually much smaller than $\cB$. In particular:
\[
\cR^{*}(b_0) \subseteq \cR(b_0) \subseteq \cB
\]

Of course, $\cR^{*}(b_0)$ is not known in advance, since it requires to know
which are the {\em optimal} sequences of actions. SARSOP proceeds
by computing successive approximations of $\cR^{*}(b_0)$, hence the acronym
(Successive Approximations of the Reachable Space under Optimal Policies).

\section{Modeling Penetration Testing with POMDPs} \label{sec:pomdp-modeling}

As penetration testing is about acting under partial
observability, POMDPs are a natural candidate to model this particular
problem. They allow to model the problem of knowledge acquisition and
to account for probabilistic information, e.g., the fact that certain
configurations or vulnerabilities are more frequent than others. In
comparison, classical planning approaches \cite{LucSarRic10} assume
that the whole network configuration is known, so that no exploration
is required.
The present section discusses how to formalize penetration testing
using POMDPs. As we shall see, the uncertainty is located essentially
in the initial belief state.
This is different from modeling the uncertainty in pentesting using
probabilistic action outcomes as in \cite{SarRicLuc11}, which does not
account for the real dynamics of the system. Also, as indicated
previously, unlike our POMDPs, the approach of Sarraute et
al.\ \cite{SarRicLuc11} only chooses exploits, assuming
a naive {\em a priori} knowledge acquisition and thus ignoring the
interaction between these two.

\subsection{States}

First, any sensible penetration test will have a finite
execution. There is nothing to be gained here by infinitely executing
a looping behavior. Every pentest terminates either when some event
(e.g., an attack detection) stops it, or when the additional access
rights that could yet be gained (from the finite number of access
rights) do not outweigh the associated costs. %
This implies that there exists an absorbing {\em terminal} state and
that we are solving a Stochastic Shortest Path problem (SSP).

Then, in the context of pentesting, we do not need the full state of the
system to describe the current situation. We will thus focus on
aspects that are relevant for the task at hand.
This state for example does not need to comprise the network topology
as it is assumed here to be static and known. But it will have to
account for the configuration and status of each computer on the
network.

A computer's {\em configuration} needs to describe the applications
present on the computer and that may (i) be vulnerable or (ii) reveal
information about potentially vulnerable applications. This comprises
its operating system (OS) as well as server applications for the web,
databases, email, ... The description of an application does not need
to give precise version numbers, but should give enough details to
know which (known) vulnerabilities are present, or what information
can be obtained about the system. For example, the open ports on a
given computer are aspects of the OS that may reveal not only the OS
but also which applications it is running.

The computers' configurations (and the network topology) give a {\em
  static} picture of the system independently of the progress of the
pentest. To account for the current situation one needs to specify,
for each computer, whether a given agent has been installed on it,
whether some applications have crashed (e.g., due to the failure of an
exploit), and which computers are accessible. Which computers are
accessible depends only on the network topology and on where agents
have been installed, so that there is no need to explicitly add this
information in the state.
Table~\ref{tab:States-M0} gives a {\tt states} section from an actual
POMDP file (using the file format of Cassandra's toolbox) in a setting
with a single machine {\tt M0}, which is always accessible (not
mentioning the computer from which the pentest is started).

\begin{table}[tbp]
  {
\small
    \begin{minipage}{0.48\linewidth}
      \begin{verbatimtab}
states :
terminal
M0-win2000
M0-win2000-p445
M0-win2000-p445-SMB
M0-win2000-p445-SMB-vuln
M0-win2000-p445-SMB-agent
M0-win2003
M0-win2003-p445
M0-win2003-p445-SMB
\end{verbatimtab}
\end{minipage}
\hfill
\begin{minipage}{0.48\linewidth}
\begin{verbatimtab}
M0-win2003-p445-SMB-vuln
M0-win2003-p445-SMB-agent
M0-winXPsp2
M0-winXPsp2-p445
M0-winXPsp2-p445-SMB
M0-winXPsp2-p445-SMB-vuln
M0-winXPsp2-p445-SMB-agent
M0-winXPsp3
M0-winXPsp3-p445
M0-winXPsp3-p445-SMB
      \end{verbatimtab}
    \end{minipage}
  }
  \caption[A list of states in a setting with a single computer (M0).]
  {A list of states in a setting with a single computer (M0)
    which can be a Windows 2000, 2003,
    XPsp2 or XPsp3, may have port 445 open and, if so, may be running
    a SAMBA server which may be vulnerable (except for XPsp3) and
    whose vulnerability may have been exploited.}
  \label{tab:States-M0}
\end{table}

Note that a computer's configuration should also provide information
on whether having access to it is valuable in itself, e.g., if there
is valuable data on its hard drive. This will be used when defining
the reward function.

\subsection{Actions \& Observations}

First, we need a {\tt Terminate} action that can be used to reach the
{\tt terminal} state voluntarily. Note that specific outcomes of
certain actions could also lead to that state.

Because we assume that the network topology is known {\em a priori},
there is no need for actions to discover reachable machines. We are
thus left with two types of actions: {\em tests}, which allow to
acquire information about a computer's configuration, and {\em
  exploits}, which attempt to install an agent on a computer by
exploiting a vulnerability. Table~\ref{tab:Actions-M0} lists actions in
our running example started in Table~\ref{tab:States-M0}.

\begin{table}[tbp]
  {
    \small
    \begin{minipage}{0.48\linewidth}
      \begin{verbatimtab}
actions :
Terminate
Probe-M0-p445
OSDetect-M0
\end{verbatimtab}
\end{minipage}
\hfill
\begin{minipage}{0.48\linewidth}
\begin{verbatimtab}

Exploit-M0-win2000-SMB
Exploit-M0-win2003-SMB
Exploit-M0-winXPsp2-SMB
      \end{verbatimtab}
    \end{minipage}
  }
  \caption{A list of actions in the same setting as
    Table~\ref{tab:States-M0}, with 1 {\tt Terminate} action, 2 tests,
    and 3 possible exploits.}
  \label{tab:Actions-M0}
\end{table}

\subsubsection{Tests}

Tests are typically performed using programs such as {\em nmap}
\cite{Fyodor98}, which scans a specific computer for open ports and,
by analyzing the response behavior of ports, allows to make guesses
about which OS and services are running. Note that such observation
actions have a cost either in terms of time spent performing analyses,
or because of the probability of being detected due to the generated
network activity. This is the reason why one has to decide which
tests to perform rather than perform them all.

In our setting, we only consider two types of tests:
\begin{description*}

\item[OS detection:] A typical OS detection will return a list of
  possible OSes, the ones likely to explain the observations of the
  analysis tool. As a result, one can prune from the belief state
  (=set to zero probability) all the states corresponding with
  non-matching OSes, and then re-normalize the remaining non-zero
  probabilities.

  Keeping with the same running example,
  Table~\ref{tab:OSDetect-M0} presents the transition and observation
  models associated with action {\tt OSDetect-M0}, which can
  distinguish winXP configurations from win2000/2003; and following is
  an example of the evolution of the belief state:
 
  \vspace{0.3cm}
  \centerline{
    \begin{tabular}{c@{ $($}c@{,}c@{,}c@{,}c@{,}c@{,}c@{,}c@{,}c@{,}c@{,}c@{,}c@{,}c@{,}c@{,}c@{,}c@{,}c@{,}c@{,}c@{,}c@{$)$}}
      initial
      & 0& 0& 0& 0& 0& 0& $\frac{1}{8}$& $\frac{1}{8}$& $\frac{1}{8}$& $\frac{1}{8}$& 0& $\frac{1}{8}$& $\frac{1}{8}$& $\frac{1}{8}$& $\frac{1}{8}$& 0& 0& 0& 0 \\[0.15cm] 
  \hline 
      {\tt winXP}
      & 0& 0& 0& 0& 0& 0& 0& 0& 0& 0& 0& $\frac{1}{4}$& $\frac{1}{4}$& $\frac{1}{4}$& $\frac{1}{4}$& 0& 0& 0& 0 \\[0.15cm]
      {\tt win2000/2003}
      & 0& 0& 0& 0& 0& 0& $\frac{1}{4}$& $\frac{1}{4}$& $\frac{1}{4}$& $\frac{1}{4}$& 0& 0& 0& 0& 0& 0& 0& 0& 0
    \end{tabular}
  }
  \vspace{0.3cm}
  
  \begin{table}[tbp]
    {
      \footnotesize
      \begin{verbatimtab}
T: OSDetect-M0 identity
O: OSDetect-M0: * : * 0
O: OSDetect-M0: * : undetected 1
O: OSDetect-M0: M0-win2000               : win 1
O: OSDetect-M0: M0-win2000-p445          : win 1
O: OSDetect-M0: M0-win2000-p445-SMB      : win 1
O: OSDetect-M0: M0-win2000-p445-SMB-vuln : win 1
O: OSDetect-M0: M0-win2000-p445-SMB-agent: win 1
O: OSDetect-M0: M0-win2003               : win 1
O: OSDetect-M0: M0-win2003-p445          : win 1
O: OSDetect-M0: M0-win2003-p445-SMB      : win 1
O: OSDetect-M0: M0-win2003-p445-SMB-vuln : win 1
O: OSDetect-M0: M0-win2003-p445-SMB-agent: win 1
O: OSDetect-M0: M0-winXPsp2               : winxp 1
O: OSDetect-M0: M0-winXPsp2-p445          : winxp 1
O: OSDetect-M0: M0-winXPsp2-p445-SMB      : winxp 1
O: OSDetect-M0: M0-winXPsp2-p445-SMB-vuln : winxp 1
O: OSDetect-M0: M0-winXPsp2-p445-SMB-agent: winxp 1
O: OSDetect-M0: M0-winXPsp3               : winxp 1
O: OSDetect-M0: M0-winXPsp3-p445          : winxp 1
O: OSDetect-M0: M0-winXPsp3-p445-SMB      : winxp 1
\end{verbatimtab}
    }
\vspace{-0.4cm}                
    \caption[Transition and observation models for action {\tt OSDetect-M0}.]
    {Transition and observation models for action {\tt
        OSDetect-M0}. The first line specifies that this action's
      transition matrix is the identity matrix. The remaining lines
      describe this action's observation function by giving the
      probability (here 0 or 1) of each possible state-observation
      pair, defaulting to the {\tt undetected} observation for all
      states.}
    \label{tab:OSDetect-M0}
  \end{table}

\item[Port scan:] Scanning port $X$ simply tells if it is
  open or closed; by pruning from the belief state the states that
  match the open/closed state of port $X$, one implicitely
  refines which OS and applications may be running.

  Action {\tt Probe-M0-p445}, for example, is modeled as depicted on
  Table~\ref{tab:Probe-M0-p445} and could give the following evolution:

  \vspace{0.3cm}
  \centerline{
    \begin{tabular}{c@{ $($}c@{,}c@{,}c@{,}c@{,}c@{,}c@{,}c@{,}c@{,}c@{,}c@{,}c@{,}c@{,}c@{,}c@{,}c@{,}c@{,}c@{,}c@{,}c@{$)$}}
      initial
      & 0& 0& 0& 0& 0& 0& $\frac{1}{8}$& $\frac{1}{8}$& $\frac{1}{8}$& $\frac{1}{8}$& 0& $\frac{1}{8}$& $\frac{1}{8}$& $\frac{1}{8}$& $\frac{1}{8}$& 0& 0& 0& 0 \\[0.15cm]
      \hline
      {\tt open-port}
      & 0& 0& 0& 0& 0& 0& 0& $\frac{1}{6}$& $\frac{1}{6}$& $\frac{1}{6}$& 0& 0& $\frac{1}{6}$& $\frac{1}{6}$& $\frac{1}{6}$& 0& 0& 0& 0 \\[0.15cm]
      {\tt closed-port}
      & 0& 0& 0& 0& 0& 0& $\frac{1}{2}$& 0& 0& 0& 0& $\frac{1}{2}$& 0& 0& 0& 0& 0& 0& 0
    \end{tabular}
  }
  \vspace{0.3cm}

  \begin{table}[tbp]
    {
      \footnotesize
      \begin{verbatimtab}
T: Probe-M0-p445 identity
O: Probe-M0-p445: * : * 0
O: Probe-M0-p445: * : closed-port 1
O: Probe-M0-p445: M0-win2000-p445           : open-port 1
O: Probe-M0-p445: M0-win2000-p445-SMB       : open-port 1
O: Probe-M0-p445: M0-win2000-p445-SMB-vuln  : open-port 1
O: Probe-M0-p445: M0-win2000-p445-SMB-agent : open-port 1
O: Probe-M0-p445: M0-win2003-p445           : open-port 1
O: Probe-M0-p445: M0-win2003-p445-SMB       : open-port 1
O: Probe-M0-p445: M0-win2003-p445-SMB-vuln  : open-port 1
O: Probe-M0-p445: M0-win2003-p445-SMB-agent : open-port 1
O: Probe-M0-p445: M0-winXPsp2-p445          : open-port 1
O: Probe-M0-p445: M0-winXPsp2-p445-SMB      : open-port 1
O: Probe-M0-p445: M0-winXPsp2-p445-SMB-vuln : open-port 1
O: Probe-M0-p445: M0-winXPsp2-p445-SMB-agent: open-port 1
O: Probe-M0-p445: M0-winXPsp3-p445          : open-port 1
O: Probe-M0-p445: M0-winXPsp3-p445-SMB      : open-port 1
      \end{verbatimtab}
    }
\vspace{-0.7cm}                
    \caption{Transition and observation models for action {\tt
        Probe-M0-p445}. The transition is again the identity, and the
      observation is {\tt closed-port} by default, and {\tt open-port}
      for all states in which port 445 is open.}
    \label{tab:Probe-M0-p445}
  \end{table}
\end{description*}

Note that a test has no state outcome (the state remains the same),
and that its observation outcome is considered as deterministic: given
the---real, but hidden---configuration of a computer, a given test
always returns the same observation.
Another interesting point is that (i) tests provide information about
computer configurations and (ii) computer configurations are static,
so that there is no use repeating a test as it cannot provide or
update any information.

\subsubsection{Exploits}

Exploits make use of an application's vulnerability to gain (i) some
control over a computer from another computer (remote exploit), or
(ii) more control over a computer (local exploit / privilege
escalation). Local exploits do not differ significantly from remote
exploits since it amounts to considering each privilege level as
a different (virtual) computer in a sub-network. As a consequence, for
the sake of clarity, we only consider one privilege level per
computer.

More precisely, we consider that any successful exploit will provide
the same control over the target computer, whatever the exploit and
whatever its configuration. This allows (i) to assume that the same
set of actions is available on any controlled computer, and (ii) to
avoid giving details about which type of agent is installed on a
computer.

The success of a given exploit action $E$ depends deterministically on
the configuration of the target computer, so that: (i) there is no use
in attempting an exploit $E$ if none of the probable configurations is
compatible with this exploit, and (ii) the outcome of $E$---either
success or failure---provides information about the configuration of
the target. In the present chapter, we even assume that a computer's
configuration is completely observed once it is under control.

{\tt Exploit-M0-win2003-SMB} is modeled in
Table~\ref{tab:Exploit-M0-win2003-SMB}, and an example evolution of the
belief under this action is:

\vspace{0.3cm}
\centerline{\small
    \begin{tabular}{c@{ $($}c@{,}c@{,}c@{,}c@{,}c@{,}c@{,}c@{,}c@{,}c@{,}c@{,}c@{,}c@{,}c@{,}c@{,}c@{,}c@{,}c@{,}c@{,}c@{$)$}}
      initial
      & 0& 0& 0& 0& 0& 0& $\frac{1}{8}$& $\frac{1}{8}$& $\frac{1}{8}$& $\frac{1}{8}$& 0& $\frac{1}{8}$& $\frac{1}{8}$& $\frac{1}{8}$& $\frac{1}{8}$& 0& 0& 0& 0 \\[0.15cm] 
      \hline
      {\tt success}
      & 0& 0& 0& 0& 0& 0& 0& 0& 0& 0& 0& 0& 0& 0& 0& 1& 0& 0& 0 \\[0.15cm]
      {\tt failure}
      & 0& 0& 0& 0& 0& 0& $\frac{1}{7}$& $\frac{1}{7}$& $\frac{1}{7}$& $\frac{1}{7}$& 0& $\frac{1}{7}$& $\frac{1}{7}$& $\frac{1}{7}$& 0& 0& 0& 0& 0
    \end{tabular}
  }
\vspace{0.3cm}

  \begin{table}[tbp]
    {
      \small
      \begin{verbatimtab}
T: Exploit-M0-win2003-SMB identity
T: Exploit-M0-win2003-SMB: M0-win2003-p445-SMB-vuln
                             : * 0
T: Exploit-M0-win2003-SMB: M0-win2003-p445-SMB-vuln
                             : M0-win2003-p445-SMB-agent 1
O: Exploit-M0-win2003-SMB: * : * 0
O: Exploit-M0-win2003-SMB: * : no-agent 1
O: Exploit-M0-win2003-SMB: M0-win2003-p445-SMB-agent
                             : agent-installed 1
      \end{verbatimtab}
    }
\vspace{-0.7cm}                
    \caption{Transition and observation models for action {\tt
        Exploit-M0-win2003-SMB}. The transition is the identity except
      if M0 is vulnerable, where an agent gets installed. The
      observation is {\tt no-agent} by default and {\tt
        agent-installed} if the exploit is successful.}
    \label{tab:Exploit-M0-win2003-SMB}
  \end{table}

\subsection{Rewards}

First, no reward is received when the {\tt Terminate} action is used,
or once the terminal state is reached.
Otherwise, the reward function has to account for various things:
\begin{description*}
\item[Value of a computer ($r_c$):] The objective of a pentest is to
  gain access to a number of computers. Here we thus propose to assign
  a fixed reward for each successful exploit (on a previously
  uncontrolled machine). In a more realistic setting, one could reward
  accessing for the first time a given valuable data, whatever
  computer hosts these data.
\item[Time is money ($r_t$):] Each action---may it be a test or an
  exploit---has a duration, so that the expected duration of the pentest
  may be minimized by assigning each transition a cost (negative
  reward) proportional to its duration. One could also
  consider a maximum time for the pentest rather than minimizing it.
\item[Risk of detection ($r_d$):] We do not explicitely model the
  event of being detected (that would lead to the terminal state with
  an important cost), but simply consider transition costs that depend
  on the probability of being detected.
\end{description*}
As a result, a transition $s,a,s'$ comes with a reward that is the sum
of these three components: $r=r_c+r_t+r_d$. Although some rewards are
positive, we are still solving an SSP
since such positive rewards cannot be received multiple times and thus
cyclic behavior is not sensible.

\subsection{POMDP Model Generation} \label{sec:pomdp-model-generation}

Generating a POMDP model for pentesting requires knowledge about
possible states, actions, and observations, plus the reward function
and the initial belief state.  Note first that the POMDP model may
evolve from one pentest to the next due to new applications, exploits
or tests.

Action and observation models for the various possible tests and
exploits can be derived from the documentation of testing tools (see,
e.g., nmap's manpage) and databases such as CVE (Common
Vulnerabilities and
Exposures)\footnote{http://cve.mitre.org/}. Information could
presumably be automatically extracted from such databases, which are
already very structured. In our experiments, we start from a
proprietary database of Core Security Technologies. %
The two remaining components of the model---the reward function and
the initial belief state---involve quantitative information which is
more difficult to acquire. In our experiments, this information is
estimated based on expert knowledge. 

Regarding rewards, statistical models can be used to estimate, for any
particular action, the probability of being detected, and the
probabilistic model of its duration. But a human decision is required
to assign a value for the cost of a detection, for gaining control
over one target computer or the other, and for spending a certain
amount of time.

The definition of the initial belief state is linked to the fact that
penetration testing is a task repeated regularly, and has access to
previous pentesting reports on the same network. The pentester thus
has knowledge about the previous configuration of the network
(topology and machines), and which weaknesses have been reported. This
information, plus knowledge of typical update behaviors (applying
patches or not, downloading service packs...), allows an informed
guess on the current configuration of the network.

\begin{figure}[tbp]
  \centerline{
    \resizebox{0.88\linewidth}{!}{
      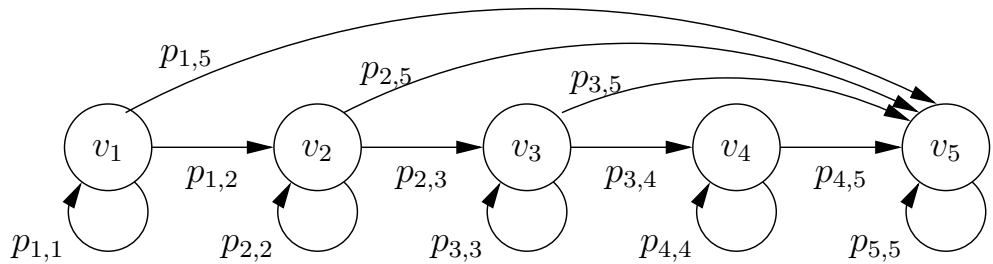
    }
  }
  \caption{Markov Chain modeling an application's updating
    process. Vertices are marked with version numbers, and edges with
    transition probabilities per time step.}
  \label{fig:updating-MC}
\end{figure}

We propose to mimick this reasoning to compute the initial belief
state. To keep things simple, we only consider a basic software update
behavior (assuming that softwares are independent from each other):
each day, an application may probabilistically stay unchanged, or be
upgraded to the next version or to the latest version. The updating
process of a given application can then be viewed as a Markov chain as
illustrated in Fig.~\ref{fig:updating-MC}. Assuming that (i) the
belief about a given application version was, at the end of the last
pentest, some vector $\vv_0$, and (ii) $T$ days (the time unit in the
Markov chain) have passed, then this belief will have to be updated as
$\vv_T = U^T \vv_0$, where $U$ is the matrix representation of the
chain. For Fig.~\ref{fig:updating-MC}, this matrix reads:
\begin{align*}
  U &=
  \left(
    \begin{array}{ccccc}
      p_{1,1} & 0 & 0 & 0 & 0 \\
      p_{1,2} & p_{2,2} & 0 & 0 & 0 \\
      0 & p_{2,3} & p_{3,3} & 0 & 0 \\
      0 & 0 & p_{3,4} & p_{4,4} & 0 \\
      p_{1,5} & p_{2,5} & p_{3,5} & p_{4,5} & p_{5,5} \\
    \end{array}
  \right).
\end{align*}

\noindent
This provides a factored approach to compute initial belief states. Of
course, in this form the approach is very simplistic. A realistic
method would involve elaborating a realistic model of system
development. This is a research direction in its own right. We come
back to this at the end of the chapter.

\section{Solving Penetration Testing with POMDPs}

We now describe our experiments. We first fill in some details on the
setup, then discuss different scaling scenarios, before having a
closer look at some example policies generated by the POMDP solver.

\subsection{Setup of Experiments}

The experiments are run on a machine with an Intel Core2 Duo CPU at
2.2 GHz and 3 GB of RAM. We use the APPL (Approximate POMDP Planning)
toolkit\footnote{APPL 0.93 at
  http://bigbird.comp.nus.edu.sg/pmwiki/farm/appl/}.  This C++
implementation of the SARSOP algorithm is easy to compile and use, and
has reasonable performance. The solver is run without time horizon
limit, until a target precision $\epsilon = 0.001$ is reached. Since
we are solving a stochastic shortest path problem, a discount factor
is not required, however we use $\gamma = 0.95$ to improve
performance. We will briefly discuss below the effect of changing
$\epsilon$ and $\gamma$.

Our problem generator is implemented in Python. It has 3 parameters:
\begin{itemize*}
\item number of machines $M$ in the target network,
\item number of exploits $E$ in the pentesting tool, that are
  applicable in the target network,
\item time delay $T$ since the last pentest, measured in days.
\end{itemize*}
For simplicity we assume that, at time $T = 0$, the information about
the network is perfect, i.e., there is no uncertainty. As $T$ grows,
uncertainty increases as described in the previous section, where the
parameters of the underlying model, cf.\ Fig.~\ref{fig:updating-MC},
are estimated by hand. The network topology consists of $1$ outside
machine and $M-1$ other machines in a fully connected network. The
configuration details are scaled along with $E$, i.e., details are
added as relevant for the exploits (note that irrelevant configuration
details would not serve any purpose in this application). As
indicated, the exploits are taken from a Core Security database which
contains the supported systems for each exploit (specific OS and
application versions that are vulnerable). The $E$ exploits are
distributed evenly over the $M$ machines. We require that $E \geq M$
so that each machine gets at least one exploit (otherwise the machine
could be removed from the encoding).

\subsection{Combined Scaling}

\Figure{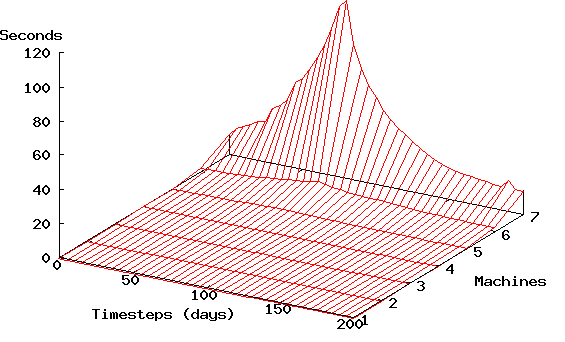}{ht}{ 0.75 \linewidth }{pomdp-machine-time}
{POMDP solver runtime ($z$ axis) when scaling time delay vs. the number of machines}

\Figure{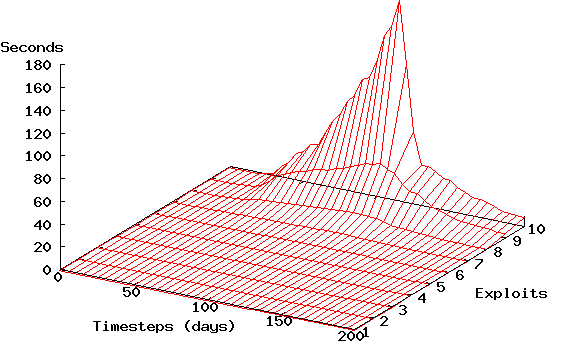}{ht}{ 0.75 \linewidth }{pomdp-exploits-time}
{POMDP solver runtime ($z$ axis) when scaling time delay vs. the number of exploits}

\Figure{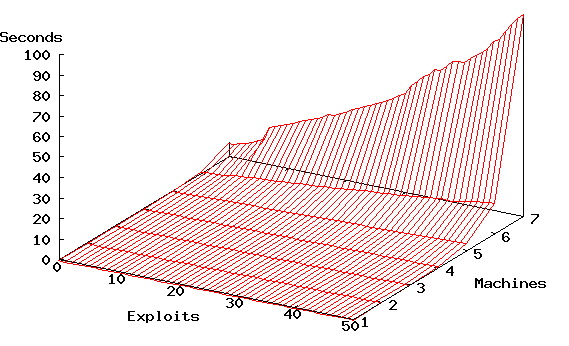}{ht}{ 0.75 \linewidth }{pomdp-exploits-machines-T10}
{POMDP solver runtime ($z$ axis) when scaling machines
    vs. exploits with time delay 10}

\Figure{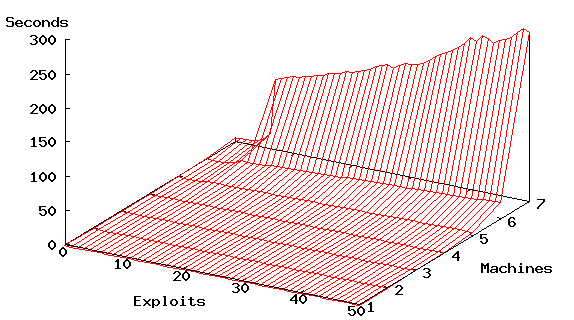}{ht}{ 0.75 \linewidth }{pomdp-exploits-machines-T80}
{POMDP solver runtime ($z$ axis) when scaling machines
    vs. exploits with time delay 80}

We discuss performance---solver runtime---as a function of $M$, $E$,
and $T$. To make data presentation feasible, at any one time we scale
only 2 of the parameters.

Consider first Figure~\ref{fig:pomdp-machine-time}, which scales $M$
and $T$. $E$ is fixed to the minimum value, i.e., each machine has a
fixed OS version and one target application. In this setting, there
are $3^M$ states. For $M = 8$, the generated POMDP file has $6562$
states and occupies $71$ MB on disk; the APPL solver runs out of
memory when attempting to parse it. Thus, in this and all experiments
to follow, $M \leq 7$.

Naturally, runtime grows exponentially with $M$---after all, even the
solver input does. As for $T$, interestingly this exhibits a very
pronounced easy-hard-easy pattern. Investigating the reasons for this,
we found that it is due to a low-high-low pattern of the ``amount of
uncertainty'' as a function of $T$. Intuitively, as $T$ increases, the
probability distribution in the initial belief state first becomes
``broader'' because more application updates are possible. Then, after
a certain point, the probability mass accumulates more and more ``at
the end'', i.e., at the latest application versions, and the
uncertainty decreases again. Formally, this can be captured in terms
of the entropy of $b_0$, which exhibits a low-high-low pattern
reflecting that of Figure~\ref{fig:pomdp-machine-time}.

In Figure~\ref{fig:pomdp-exploits-time}, scaling the number $E$ of
exploits as well as $T$, the number of machines is fixed to 2 (the
localhost of the pentester, and one target machine). We observe the
same easy-hard-easy pattern over $T$. As with $M$, runtime grows
exponentially with $E$ (and must do so since the solver input
does). However, with small or large $T$, the exponential behavior does
not kick in until the maximum number of exploits, 10, that we consider
here. This is important for practice since small values of $T$ (up to
$T=50$) are rather realistic in regular pentesting. In the next
sub-section, we will examine this in more detail to see how far we can
scale $E$, in the 2-machines case, with small $T$.

Figures~\ref{fig:pomdp-exploits-machines-T10} and \ref{fig:pomdp-exploits-machines-T80} show the combined
scaling over machines and exploits, for a favorable value of $T$
($T=10$, Fig.~\ref{fig:pomdp-exploits-machines-T10}) 
and an unfavorable one ($T=80$, Fig.~\ref{fig:pomdp-exploits-machines-T80}). 
Here the behavior is rather regular. By all appearances, it grows exponentially in both
parameters. An interesting observation is that, in Fig.~\ref{fig:pomdp-exploits-machines-T10}, the growth in
$M$ kicks in earlier, and rather more steeply, for $M$. This is not as
much the case for the larger $T$ value in Fig.~\ref{fig:pomdp-exploits-machines-T80}. Note though that,
there, the curve over $E$ flattens around $T=10$. It is not clear to
us what is causing this behavior; cf.\ the next sub-section.

To give an impression on the effect of the discount factor on solver
performance, with $M=2, E=11, T=40$, solver
runtime goes from 17.77 s (with $\gamma = 0.95$) to 279.65 s (with
$\gamma = 0.99$). APPL explicitly checks that $\gamma < 1$, so $\gamma
= 1$ could not be tried.  With our choice $\gamma = 0.95$ we still get
good policies (cf. further below).

\subsection{The 2-Machines Case}

\begin{figure}[htb]
\centerline{\includegraphics[width=0.8\textwidth]{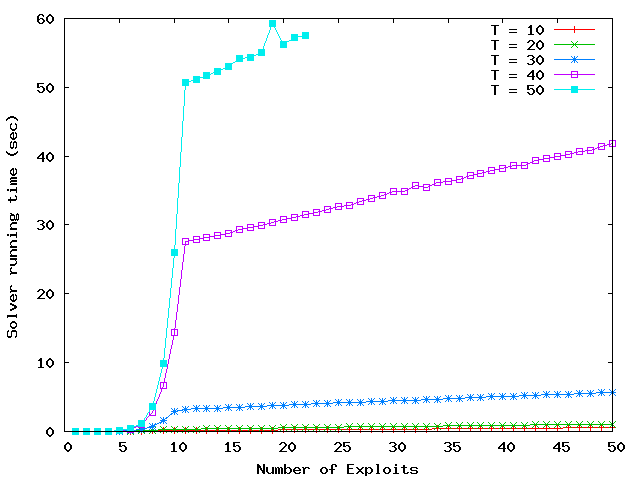}}
  \caption{\label{fig:2-machines} POMDP solver runtime when scaling
    the number of exploits, for different realistically small settings
    of the time delay, in the 2-machines case.}
\end{figure}

As hinted, the 2-machines case is relevant because it may serve as the
``atomic building block'' in an industrial-scale solution, cf.\ also
the discussion in the outlook below. The question then is whether or
not we can scale the number of exploits into a realistic region. We
have seen above already that this is not possible for unfavorable
values of $T$. However, are these values to be expected in practice?
As far as Core Security's ``Core Insight Enterprise'' tool goes, the
answer is ``no''. In security aware environments, pentesting should be
performed at regular intervals of at most 1 month. Consequently,
Figure~\ref{fig:2-machines} shows data for $T \leq 50$.

For the larger values of $T$, the data shows a very steep incline
between $E=5$ and $E=10$, followed by what appears to be linear
growth. This behavior is caused by an unwanted bias in our current
generator.\footnote{%
The exploits to be added are ordered in a way so that their likelihood
of succeeding decreases monotonically with $|E|$. After a certain
point, they are too unlikely to affect the policy quality by more than
the target precision $\epsilon$. The POMDP solver appears to determine
this effectively. } Ignoring this phenomenon, what matters to us here
is that, for the most realistic values of $T$ ($T=10,20$), scaling is
very good indeed, showing no sign of hitting a barrier even at $E=50$. %
Of course, this result must be qualified against the realism of the
current generator. %
It remains an open question whether similar scaling will be achieved
for more realistic simulations of network development.

\subsection{POMDPs make Better Hackers}

As an illustration of the policies found by the POMDP solver, consider
a simple example wherein the pentester has 4 exploits: an SSH exploit
(on OpenBSD, port 22), a wu-ftpd exploit (on Linux, port 21), an IIS
exploit (on Windows, port 80), and an Apache exploit (on Linux, port
80). The probability of the target machine being Windows is higher than
the probability of the other OSes.

Previous automated pentesting methods, e.g.\ Lucangeli {\em et
  al.}\ \cite{LucSarRic10}, proceed by first performing a port
scan on common ports, then executing OS detection module(s), and
finally launching exploits for potentially vulnerable services.

With our POMDP model, the policy obtained is to first test whether
port 80 is open, because the expected reward is greater for the two
exploits which target port 80, than for each of the exploits for port
21 or 22.  If port 80 is open, the next action is to launch the IIS
exploit for port 80, skipping the OS detection because Windows is more
probable than Linux, and the additional information that OS Detect can
provide doesn't justify its cost (additional running time). %
If the exploit is successful, terminate. Otherwise, continue with the
Apache exploit (not probing port 80 since that was already done), and
if that fails then probe port 21, etc.

In summary, the policy orders exploits by promise, and executes port
probe and OS detection actions on demand where they are
cost-effective. %
This improves on Sarraute et al.\ \cite{SarRicLuc11}, whose
technique is capable only of ordering exploits by promise. %
What's more, practical cases typically involve exploits whose outcome
delivers information about the success probability of other exploits,
due to common reasons for failure---exploitation prevention
techniques. Then the best ordering of exploits depends on previous
exploits' outcome. POMDP policies handle this naturally, however it is
well beyond the capabilities of Sarraute et al.'s approach. We omit
the details for space reasons.

\section{Discussion}

POMDPs can model pentesting more naturally and accurately than
previously proposed planning-based models
\cite{LucSarRic10,SarRicLuc11}. While, in general, scaling is limited,
we have seen that it appears reasonable in the 2-machines case where
we are considering only how to get from one machine to another. An
idea to use POMDP reasoning in practice is thus to perform it for all
connected pairs of machines in the network, and thereafter use these
solutions as the input for a high-level planning procedure. That
procedure would consider the pairwise solutions to be atomic, i.e., no
backtracking over these decisions would be made. Indeed, this is one
of the abstractions made---successfully, as far as runtime performance
is concerned---by Sarraute et al.\ \cite{SarRicLuc11}. Our
immediate future work will be to explore whether a POMDP-based
solution of this type is useful, the question being how large the
overhead for planning all pairs is, and how much of the solution
quality gets retained at the global level.

A line of basic research highlighted by our work is the exploitation
of special structures in POMDPs. First, in our model, all actions are
deterministic. Second, some of the uncertain parts of the state (e.g.\
the operating systems) are static, for the purpose of pentesting, in
the sense that none of the actions affect them. Third, unless one
models possible detrimental side-effects of exploits (cf.\ directly
below), pentesting is ``monotonic'': accessibility, and thus the set
of actions applicable, can only grow. Fourth, any optimal policy will
apply each action at most once. Finally, some aspects of the
state---in particular, which computers are controlled and
reachable---are directly visible and could be separately modeled as
being such. To our knowledge, this last property alone has been
exploited in POMDP solvers (e.g.,\ \cite{AraThoBufCha-ICTAI10}), and the only
other property mentioned in the literature appears to be the first one
(e.g.,\ \cite{Bon09}).

While accurate, our current model is of course not ``the final word''
on modeling pentesting with POMDPs. As already mentioned, we currently
do not explicitly model the detrimental side-effects exploits may
have, i.e., the cases where they are detected (spawning a reaction of
the network defense) or where they crash a
machine/application. Another important aspect that could be modeled in
the POMDP framework is that machines are not independent. Knowing the
configuration of some computers in the network provides information
about the configuration of other computers in the same network. This
can be modeled in terms of the probability distribution given in the
initial belief. An interesting question for future research then is
how to generate these dependencies---and thus the initial belief---in
a realistic way. Answering this question could go hand in hand with
more realistically simulating the effect of the ``time delay'' in
pentesting. Both could potentially be adressed by learning appropriate
graphical models \cite{KolFri09}, based on up-to-date real-world
statistics.

To close the chapter, it must be admitted that, in general, ``pentesting
$\neq$ POMDP solving''.
Computer security is always evolving, so that the
probability of meeting certain computer configurations changes with
time. An ideal attacker should continuously learn the probability
distributions describing the network and computer configurations it
can encounter. This kind of learning can be done outside the POMDP
model, but there may be better solutions doing it more
natively. Furthermore, if the administrator of a target network reacts
to an attack, running specific counter-attacks, then the problem turns
into an adversarial game.


\chapter{Conclusions and Future Work} \label{chap:conclusion}

\section*{Part I}

In this first part, we presented the 
basic model and a complete implementation that integrates a planner system
with a penetration testing framework. 
We described an algorithm for transforming the information
present in the pentesting tool to the planning domain, 
and we showed how the scalability issues of attack graphs can be solved using current
planners.

The planner retained in our implementation is Metric-FF, developed by 
J\"org Hoffmann \cite{Hoffmann02}.
Our research prototype continued its way outside the research lab:
it was adopted by the engineering team in charge of developing 
the product Core Insight Enterprise,
integrated into the product and shipped to customers.
To meet the engineering requirements, further work was performed on 
the prototype. In particular, the testing team of Core Insight Enterprise
developed a suite of testing scenarios, more realistic and closer to the customer's 
networks than the ones used during our research.
Having access to those testing scenarios was a valuable resource 
in subsequent steps of the research, in particular the testing and evaluation
of the prototypes of Parts II and III.

\section*{Part II}

In the second part, we introduced a custom probabilistic planner, 
specifically designed for this problem.
We contributed a planning model that captures the uncertainty about the results of the actions, 
which is modeled as a probability of success of each action;
and we presented efficient planning algorithms, that achieve industrial-scale runtime performance.
The computational complexity of this planning solution is 
$ \calO ( M^{2} \cdot n \log n) $
where $n$ is the total number of actions,
and  $M$ is the number of machines in the target network.

After this work was published in \cite{Sarraute09,Sar09proba,SarRicLuc11},
the following step was to further evaluate this research prototype
as a potential engineering solution (i.e. to perform attack planning
for the company's product).
As part of this evaluation, we tested the probabilistic planner prototype
with the scenarios developed by the testing team. This allowed us 
to improve both the planner prototype and the testing scenarios.
In particular, by making clever use of the results of 
intermediate computations, we were able to bring down the complexity
of the solution to $\calO ( M \cdot n \log n) $,
and obtain remarkable improvements in planner runtime.
A natural next step for this prototype is thus to graduate from the
research lab, and enter the engineering circuit.

\section*{Part III}

The work described in the third part is the result of a collaboration
with J\"org Hoffman (Saarland University) and Olivier Buffet (INRIA, Nancy).
In this collaboration we took a different direction: 
instead of trying to improve the efficiency of the solutions developed,
we focused on improving the model of the attacker, and formulating the
problems that we think will have to be solved in the longer term.

We modeled the attack planning problem in terms of partially observable
Markov decision processes (POMDP).
This grounded penetration testing in a well-researched formalism,
and highlighted important aspects of the problem's nature. 
In particular POMDPs allowed us to model information gathering as an integral part of the
attack, thus providing for the first time a means to intelligently
mix scanning actions with actual exploits.

As a continuation of this modeling activity,
we devised a method that relies on POMDPs to find good attacks on
individual machines, which are then composed into an attack on the
network as a whole. This decomposition exploits the network structure to
the extent possible, making targeted approximations (only) where
needed. This new approach will be presented 
at the JFPDA conference in Nancy, France, during May 2012 \cite{SarBufHof-jfpda12};
and at the AAAI conference,
that will take place in Toronto, Canada, in July 2012 \cite{SarBufHof-aaai12}.
An insight of this approach was given at the Hackito Ergo Sum conference this year \cite{Sarraute12}.

We give below a brief summary of the ideas of this decomposition.
The approach is called \emph{4AL} since it
addresses network attack at 4 different levels of abstraction. 4AL is
a POMDP solver specialized in attack planning as addressed here. Its
inputs are the logical network $LN$ and POMDP models encoding attacks
on individual machines. Its output is a policy (an attack) for the
global POMDP encoding $LN$, as well as an approximation of the value
of the global value function.

\begin{figure}[t]
    \begin{minipage}{0.45\linewidth}
      \centering
      \scalebox{0.7}{
        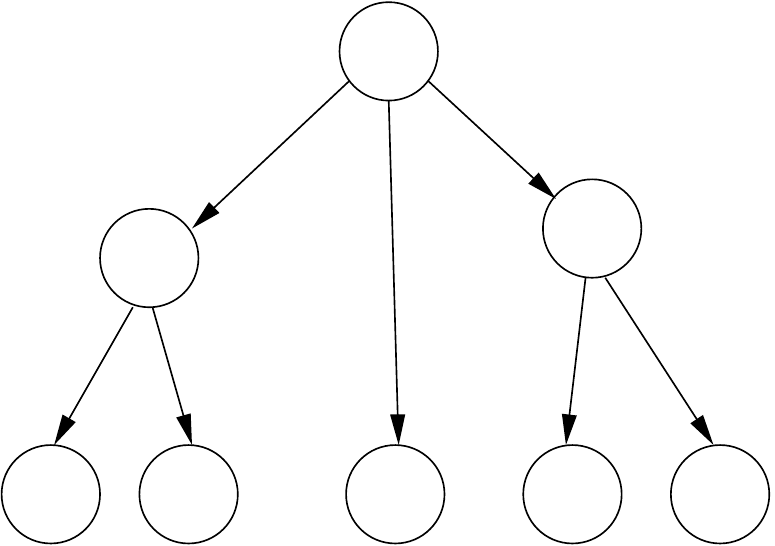
      }
      \\
      (a) $LN$ as tree of components $C$.
    \end{minipage}
    \hfill  
    \begin{minipage}{0.45\linewidth}
      \centering
      \scalebox{0.7}{
        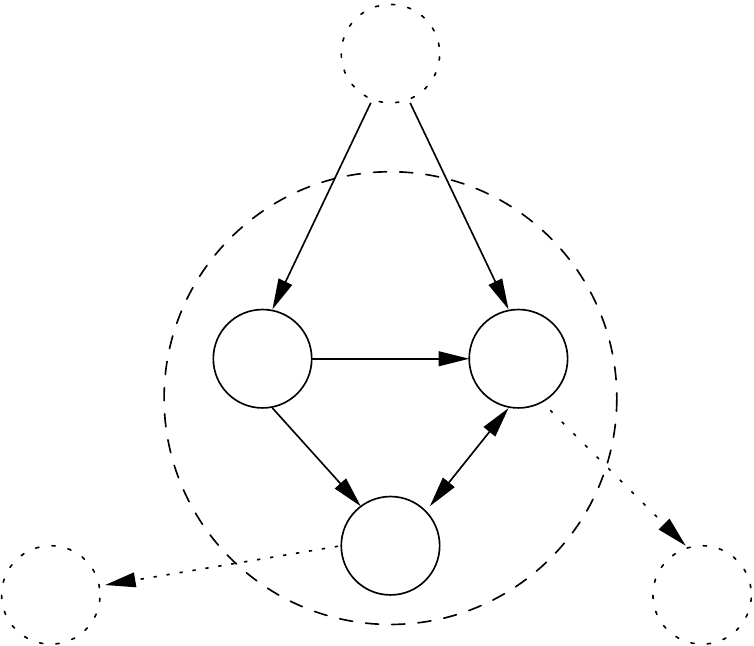
      }
      \\
      (b) Paths for attacking $C_1$.
    \end{minipage}

    \medskip
    
    \begin{minipage}{1\linewidth}
      \centering
      \scalebox{0.7}{
        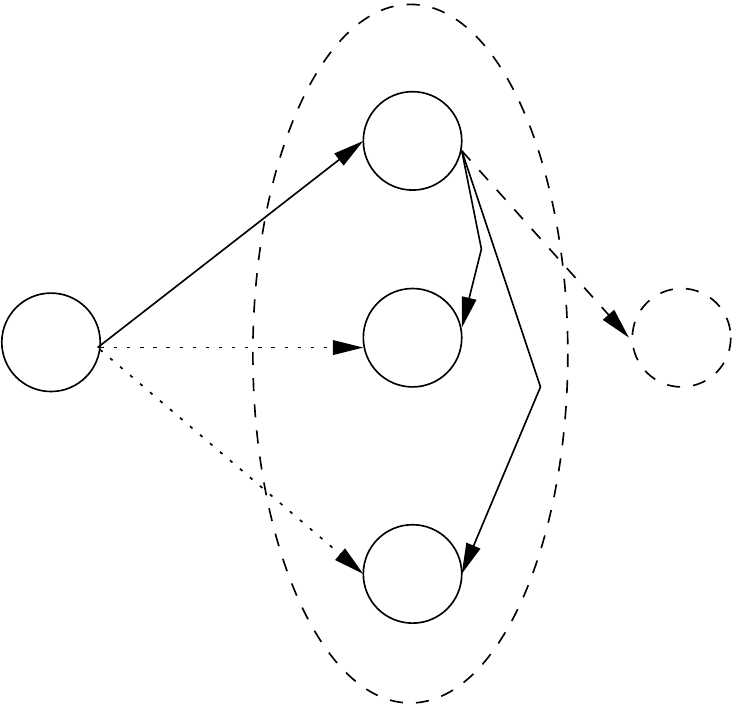
      }
      \\
      (c) Attacking $N_3$ from $N_1$, using $m$ first.   
    \end{minipage}
  \caption{Illustration of Levels 1, 2, and 3 (from left to right) of
    the 4AL algorithm.}
  \label{fig:4ALlevels123}
\end{figure}

The four levels are: (1) \emph{Decomposing the Network}, (2)
\emph{Attacking Components}, (3) \emph{Attacking Subnetworks}, and (4)
\emph{Attacking Individual Machines}. We outline these levels in
turn. Figure~\ref{fig:4ALlevels123} provides illustrations.

\vspace{-0.05cm}
\begin{itemize}
\item \textbf{Level~1:} Decompose the logical network $LN$ into a tree
  of biconnected components, rooted at \start. In reverse topological
  order, call the Level~2 procedure on each component; propagate the outcomes
  upwards in the tree.

\item \textbf{Level~2:} Given component $\componentC$, consider, for
  each rewarded subnetwork $N \in \componentC$, all paths $P$ in
  $\componentC$ that reach $N$. Backwards along each $P$, call the Level~3 procedure
  on each subnetwork and associated firewall. Choose the best path for
  each $N$. Aggregate these path values over all $N$, by summing up
  but disregarding rewards that were already accounted for by a
  previous path in the sum.

\item \textbf{Level~3:} Given a subnetwork $N$ and a firewall $F$
  through which to attack $N$, for each machine $m \in N$ approximate
  the reward obtained when attacking $m$ first. For this, modify $m$'s
  reward to take into account that, after breaking $m$, we are behind
  $F$: call Level~4 to obtain the values of all $m' \neq m$ with an
  empty firewall; then add these values, plus any pivoting reward, to
  the reward of $m$ and call Level~4 on this modified $m$ with
  firewall $F$. Maximize the resulting value over all $m \in N$.

\item \textbf{Level~4:} Given a machine $m$ and a firewall $F$, model
  the single-machine attack planning problem as a POMDP, and run an
  off-the-shelf POMDP solver. Cache known results to avoid duplicated
  effort.
\end{itemize}

In conclusion, the POMDP model of penetration testing that we devised allows us to
naturally represent many of the features of this application, in
particular incomplete knowledge about the network configuration, as
well as dependencies between different attack possibilities, and
firewalls. Unlike any previous methods, the approach is able to
intelligently mix scans with exploits. While this accurate solution
does not scale, large networks can be tackled by a decomposition
algorithm. Our present empirical results suggest that this can be
accomplished with a small loss in quality relative to a global POMDP
solution.

An important open question is to what extent our POMDP + decomposition
approach is more cost-effective than the classical planning solution
currently employed by Core Security. Our next step will be to answer
this question experimentally, comparing the attack quality of 4AL
against that of the policy that runs extensive scans and then attaches
FF's plan for the most probable configuration.

Other important directions for future work are to devise more accurate models
of software updates (hence obtaining more realistic designs of the
initial belief); to tailor POMDP solvers to this particular kind of
problem, which has certain special features, in particular the absence
of non-deterministic actions and that some of the uncertain parts of
the state (e.g.\ the operating systems) are static; and to drive the
industrial application of this technology. We hope that these will
inspire other researchers as well.


\bibliographystyle{alpha}
\bibliography{../biblio/planning,../biblio/simulations}


\end{document}